\documentclass{article}
\usepackage{url}
\usepackage{graphicx} 
\usepackage{amsthm}
\usepackage{amsmath}
\usepackage{amssymb}
\usepackage{appendix}
\usepackage{xcolor}
\usepackage{hyperref}
\usepackage{subcaption}
\usepackage{multirow}
\usepackage{booktabs}
\theoremstyle{definition}

\newtheorem{proposition}{Proposition}[section]
\newtheorem{assumption}{Assumption}[section]
\newtheorem{theorem}{Theorem}[section]
\newtheorem{remark}{Remark}[section]

\usepackage[accepted]{icml2024}

\date{}
\begin{document}
\allowdisplaybreaks
\twocolumn[
\icmltitle{Superiority of Multi-Head Attention in In-Context Linear Regression}



\icmlsetsymbol{equal}{*}

\begin{icmlauthorlist}
\icmlauthor{Yingqian Cui}{msu}
\icmlauthor{Jie Ren}{msu}
\icmlauthor{Pengfei He}{msu}
\icmlauthor{Jiliang Tang}{msu}
\icmlauthor{Yue Xing}{msustat}
\end{icmlauthorlist}

\icmlaffiliation{msu}{Department of Computer Science and Engineering, Michigan State University, US.}
\icmlaffiliation{msustat}{Department of Statistics and Probability, Michigan State Uniersity, US.}
\icmlcorrespondingauthor{Yue Xing}{xingyue1@msu.edu}

\icmlkeywords{Machine Learning, ICML} 

\vskip 0.3in
]
\printAffiliationsAndNotice{}
\begin{abstract}
    We present a theoretical analysis of the performance of transformer with softmax attention in in-context learning with linear regression tasks.
    While the existing literature predominantly focuses on the convergence of transformers with single-/multi-head attention, our research centers on comparing their performance. We conduct an exact theoretical analysis to demonstrate that multi-head attention with a substantial embedding dimension performs better than single-head attention.
    When the number of in-context examples $D$ increases, the prediction loss using single-/multi-head attention is in $O(1/D)$, and the one for multi-head attention has a smaller multiplicative constant. In addition to the simplest data distribution setting, we consider more scenarios, e.g., noisy labels, local examples, correlated features, and prior knowledge. We observe that, in general, multi-head attention is preferred over single-head attention. Our results verify the effectiveness of the design of multi-head attention in the transformer architecture.
\end{abstract}
\section{Introduction}

In-context learning (ICL) is a concept developed in natural language processing (NLP). With the rise of transformer architecture, NLP models become increasingly powerful and show their ability to learn new knowledge even without tuning the model parameters. 
Given prompts with several examples, these models can generate improved responses, showcasing their ability to adapt and `learn' from the provided context \cite{dong2022survey}.

The mechanism of transformers has been widely studied in the theoretical literature, with a main focus on linear attention~\cite{katharopoulos2020transformers,choromanski2020rethinking, schlag2021linear, liu2023linrec,ahn2023linear}, and emerging interest in the effectiveness and superiority of softmax attention function~\cite{deng2023superiority,deng2023attention, trauger2023sequence,hahn2020theoretical,chiang2022overcoming}. In recent literature, people have started to work on the theoretical understanding of ICL, e.g., \citet{zhang2023trained,oymak2023role,li2023theoretical, huang2023context, mahankali2023one,wu2023many}. Besides, \citet{von2023transformers,ahn2023transformers,akyurek2022learning,zhang2023trained} explain how ICL learns gradient descent and linear regression models. \citet{bai2023transformers} studies ICL in generalized linear models, ridge regression, and LASSO. \citet{cheng2023transformers} investigate the ability of transformers to conduct ICL on non-linear functions. Based on \citet{von2023transformers,dai2023can}, ICL can be connected with the gradient descent method.

Besides, some other studies work on multi-head attention. 
For example, \citet{mahdavi2023memorization} explored the memorization capacities of multi-head attention, and \citet{an2020repulsive} indicates a trade-off between the approximation accuracy and number of heads. Another work \cite{li2023transformers} studies the effectiveness of ReLU-activated transformers and shows the existence of multi-layer large transformers that can conduct various regression tasks. In addition, the work of ~\citet{deora2023optimization} investigates the convergence and generalization performance of multi-head attention in classification tasks.

However, we notice that existing theoretical literature focuses on either single-head or multi-head attention, and there is limited theoretical understanding of their difference.
This work bridges this gap by considering transformer with \textbf{single/multi-head softmax attention} to study its ICL performance in linear regression tasks. We provide a clear comparison to quantify the superiority of multi-head attention over single-head attention. Different from \citet{zhang2023trained}, we do not consider linear multi-attention because linear-activated single-layer single-head attention is sufficient to learn linear regression tasks.

Our contributions are summarized as follows:

First, we study the transformer architecture and show the effectiveness of single-head attention with softmax activation in ICL. We derive the exact prediction risk under the considered data generation model. (Section \ref{sec:single_head})

Second, we show that multi-head attention is better than single-head attention by figuring out the exact prediction risk of multi-head attention. With a high input embedding dimension, multi-head attention improves the flexibility of the transformer and can obtain a better kernel for the linear regression task. (Section \ref{sec:multi_head})

Finally, we also investigate the scenarios where the training data include prior knowledge, noisy responses, correlated features, or local examples. While our analysis shows that in most scenarios, multi-head attention is preferred over single-head attention, we also reveal some interesting behaviors of ICL when the data consists of local examples or have prior knowledge.
Specifically, we observe that (1) when there is a ``strong" prior knowledge, predicting using this prior knowledge leads to good performance; (2) whether local examples help or not depends on their distance to the query. (Section \ref{sec:other})

Our results provide a comprehensive understanding about the impact of single-/multi-head attention on the performance of ICL. In addition, it also offers practical guidance for selecting the efficient attention mechanism in real-world applications. In particular, multi-head attention is preferred than single-head attention, and the total number of embedding dimensions should be much larger than the number of heads.

\section{Other Related Works}
In addition to the aforementioned theoretical studies, we review some empirical studies below:


The initial work utilized by \citet{zhang2023trained} is done by \citet{garg2022can}.
They empirically show the effectiveness of the transformer in performing ICL, with performance matching the optimal least squares estimator. Furthermore, \citet{akyurek2022learning} demonstrate that the ICL done by transformers implicitly applies standard learning algorithms to conduct the in-context tasks. 

Following these works, \citet{ahuja2023context} extend the setting of \citet{garg2022can} by considering a mixture of in-context tasks in the pre-training and demonstrating the ability of the transformer to resemble the effect of Bayesian predictor under the multi-task setting. \citet{raventos2023pretraining} empirically investigates how the diversity of the tasks in the pre-training dataset influences the performance of the transformer to do in-context tasks that are unseen in the pre-training stage. Some other related works can also be found in \citet{fu2023transformers,von2023uncovering,shi2022language,saparov2022language,lu2021fantastically,liu2021makes,workrethinking,min2021noisy,zhang2022active,chen2022improving,min2021metaicl}.

\section{Notations}\label{sec:notation}


To mathematically define ICL, 
instead of merely passing a query $x_{q} \in \mathbb{R}^d$ (or a test sample) to the transformer to make a prediction, ICL passes a prompt, i.e., a few examples with their labels $\{(x_i,y_i)\}_{i=1,\ldots,D}$ together with the query $x_{q}$, to the transformer. Using the prompt in the format of
\vspace{-0.1in}
\begin{equation}\label{eqn:input}
  E=\begin{pmatrix}
    x_1 & x_2 & \ldots & x_D & x_{q}\\
    y_1 & y_2 & \ldots & y_D & 0
\end{pmatrix}\in\mathbb{R}^{(d+1) \times (D+1)},  
\end{equation}
the transformer can learn from the examples to infer the prediction for $x_{q}$. Following \citet{zhang2023trained}, we consider the following simplified neural network architecture
\begin{equation}
    f(E)= E+W_{out} H,
\end{equation}
where $H$ denotes the attention node and $W_{out}$ represents a fully-connected layer. Here, 
$H=\text{concat}(H_1,\ldots, H_h)$, with $h$ being the number of heads in the multi-head attention. Each attention head $H_j$ is given by
\begin{equation}\label{eqn:multi_head}
    H_j = W^V_jE\cdot \phi\left(\frac{(W^K_jE)^{\top}W^Q_jE}{\rho_j} \right)
\end{equation}
where $\rho_j$ is a normalization factor, and the activation function $\phi$ is the column-wise softmax function. For each $j$, $W^V_j, W^K_j, W^Q_j\in\mathbb{R}^{m\times d}$, and $m=d/h$. When $h=1$, the attention is \textbf{single-head}. When $h>1$, the structure is called \textbf{multi-head} attention.



To train the model, we fetch the last element of the last row in $f(E)$ as the predicted value of $y_q$ (denote as $\widehat{y}_q$), then minimize
\begin{eqnarray}\label{eqn:loss}
    L(\Omega)=\mathbb{E}_{\{x_i\},x_q,\theta}(\widehat y_q-y_q)^2,
\end{eqnarray}
where $\Omega$ is the set of parameters.
 
\vspace{-0.1in}
\section{Superiority of Multi-Head Attention}\label{sec:benefit}


In this section, we introduce the assumptions, present the optimal solution of single-head attention in ICL, and demonstrate the superiority of multi-head attention.

\subsection{Assumptions}

Before showing the main results, we first introduce the data generation model and configurations of the transformer:

\begin{assumption}[Data Generation Model]\label{assumption:data} In each prompt, the examples $(x_i,y_i)$ and $(x_q,y_q)$ are i.i.d. samples from the following noiseless regression model:
\begin{itemize}
    \item The ``input" $x\sim N(0,I_d)$.
    \item The ``output" $y=\theta^{\top}x$.
    \item The coefficients $\theta$ are the same for the samples in the same prompt and are different across different prompts. In addition, $\theta\sim N(0,I_d/d)$.
\end{itemize}
    
\end{assumption}
\begin{assumption}[Lazy Training]\label{assumption:nn}
We consider a lazy training scheme when deriving the optimal solution of the transformer. We first fix $W_{out}$, $W^V$ and optimize over the other parameters, and then figure out the best solution of $W_{out}, W^V$.
\end{assumption}

Assumption \ref{assumption:data} follows \citet{zhang2023trained} on the data generation model. For simplicity, we use Gaussian distribution to avoid tedious discussions on potential heavy tail issues, and our proofs, in general, can be extended to other data generation models. 

In Assumption \ref{assumption:nn}, we apply lazy training to the attention. As mentioned by \citet{huang2023context}, training all parameters in a transformer is a non-convex problem. Assuming lazy training can simplify the analysis. However, it is important to note that our conclusion, which states that the optimal solution of single-head attention has a worse ICL performance than multi-head attention, is independent of the lazy training assumption.
\vspace{-0.05in}
\subsection{Optimal Solution for Single-Head Attention}\label{sec:single_head}

In this subsection, we figure out the optimal solution of single-head attention and summarize it in Theorem \ref{thm:optimal}.
\begin{theorem}[Optimal Solution of Single-Head Attention]\label{thm:optimal}
    Under Assumption \ref{assumption:data}, \ref{assumption:nn}, assume (1) there is infinite training prompts, (2) $(W_{out}W^V)_{d+1,:}=(0,\ldots,0,v)$, and (3) $(W^K)^{\top}W_Q$ is in a format of 
    \begin{equation*}
        (W^K)^{\top}W^Q = \begin{bmatrix}
        A & 0 \\ b & 0
        \end{bmatrix},
    \end{equation*}
    then when $D\rightarrow\infty$, the loss value is
    \begin{eqnarray*}
        L(A,b,v)=\frac{1}{d}tr((vA-I_d)^2)+v^2\|b\|^2\mathbb{E}\|\theta\|^4+O(1/D),
    \end{eqnarray*}
    and the optimal solution satisfies that $\|vA-I_d\|_F^2=O(1/D)$, and $\|vb\|^2=O(1/D)$. In addition, when taking $A=I_d/v$ and $b=0$, 
    \begin{align}
    L(I_d/v,0,v)=\frac{v^2}{D}\left(\frac{v^2}{v^2-2}\right)^\frac{d}{2} + \frac{1}{D}\left(\frac{v^2}{v^2-2}\right)^{\frac{d}{2}+1} + o(1/D).
    \label{eqn:single_optimal}
    \end{align}
\vspace{-0.25in}

Denoting the optimal solution as $A^*$, $b^*$, for any $v^2>2$,
\begin{eqnarray*}
    L(A^*,b^*,v) - L(I_d/v,0,v)=o(1/D).
\end{eqnarray*}

\end{theorem}
\vspace{-0.1in}

Theorem \ref{thm:optimal} shows the optimal solution of the single-head attention when fixing $(W_{out}W^V)_{d+1,:}$. To prove Theorem \ref{thm:optimal}, we use Taylor expansion to separate the denominator and numerator of the attention scores. Since there are infinitely many training samples, we directly calculate the expectation of the output. In addition, it is also observed that the loss function is a quadratic function of $A$ and $b$. The formal proof can be found in Appendix \ref{sec:proof:optimal}. 





In Theorem \ref{thm:optimal}, we study the optimality of $A$ and $b$ when keeping $v$ fixed. Generally, $v$ affects the prediction loss in two ways. First, as stated in Theorem \ref{thm:optimal}, it is essential that $v^2>2$. When taking $v^2\leq 2$ and $A=I_d/v$,  $\exp(x_q^\top Ax_q)=\exp(\|x_q\|^2/v^2)\geq \exp(\|x_q\|^2/2)$, thus $\exp(x_q^\top Ax_q)$ has no finite expectation, and the attention score of $(x_q,0)$ towards itself becomes predominantly high. Second, when taking Taylor expansion on attention scores, we need the remainder terms to be negligible. 

\begin{remark}
In addition to the optimal solution in Theorem \ref{thm:optimal}, since the prediction loss is approximately a convex function of $(A,b)$, numerical methods such as gradient descent can successfully approximate the optimal solution.
\end{remark}

\begin{figure}[!ht]
    \centering\vspace{-0.05in}
    \includegraphics[scale=0.65]{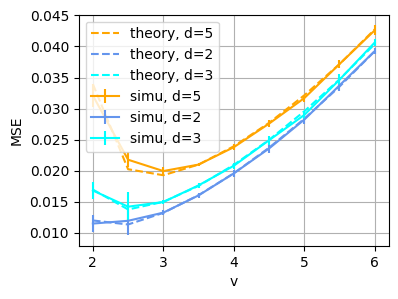}\vspace{-0.2in}
    \caption{ICL performance of single-head attention with $(A,b)=(I_d/v,0)$ and $D=1000$.}
    \label{fig:simu_change_d}
\end{figure}

\begin{figure}[!ht]
    \centering\vspace{-0.1in}
    \includegraphics[scale=0.65]{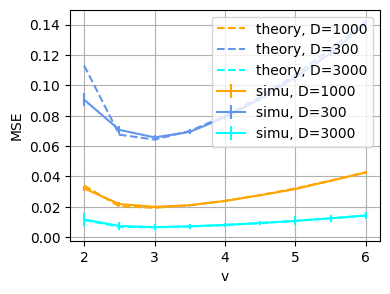}\vspace{-0.2in}
    \caption{ICL performance of single-head attention with $(A,b)=(I_d/v,0)$ and $d=5$.}
    \vspace{-0.in}
 \label{fig:simu_change_D_}
\end{figure}

\textbf{Simulation.} While Theorem \ref{thm:optimal} presents the ICL performance of single-head attention given a fixed $v$, we also conduct some simulation study to investigate the role of $v$. In the simulation, we take different choices of $(d,D,v)$ and set $(A,b)=(I_d/v,0)$ to calculate the corresponding prediction loss (MSE). We run 200k repetitions for each setting to get an average and an error bar. The results are summarized in Figure \ref{fig:simu_change_d} and \ref{fig:simu_change_D_}.

The figures show that the simulation of prediction loss aligns well with theoretical values. 
Besides, there are two main observations. First, with fixed $(d,D)$, the MSE exhibits a U-shaped behavior as a function of $v$. In Figure \ref{fig:simu_change_d}, when $d$ increases, the optimal $v$ increases as well. Second, when fixing $v$, the MSE increases with $d$ (Figure \ref{fig:simu_change_d}) and decreases with $D$ (Figure \ref{fig:simu_change_D_}).










\subsection{Multi-Head Attention is Better}\label{sec:multi_head}
While Section \ref{sec:single_head} shows the effectiveness of single-head attention, in this subsection, through deriving the exact performance, we show that multi-head attention is better than single-head attention.

In the implementation of \citet{garg2022can}, a linear layer is used to transform $E$ into a space with a higher dimension before feeding the input into the transformer. In the last layer of the transformer, another linear layer is added so that the network outputs a single number. This increases the flexibility of the transformer.

We denote the transformation matrix applied before the transformer as $W_{in}\in\mathbb{R}^{p\times (d+1)}$ with $p\geq d+1$. In single-head attention, introducing the linear layer does not change the results. This is because the rank of $W_{in}^{\top}(W^K)^{\top}W^QW_{in}$ is still $d+1$, meaning that the additional layer does not enlarge the representational capacity of single-head attention.
In contrast, multi-head attention benefits from the dimension increase provided by $W_{in}$, which allows each head to learn more features and potentially improve predictions. To explain this, in single-head attention, there is only one attention score matrix, and all the attention scores are non-negative. In contrast, we can combine the attention scores from different heads in multi-head attention so that some weights can negatively contribute to the final prediction.
This flexibility is beneficial in linear regression, as negative weights and positive weights together can provide a better fit for the data.

We consider a two-head attention in the following theorem to illustrate the superiority:


\begin{theorem}[Multi-head Attention is Better]\label{thm:multi_head}
    Consider a two-head attention with
    \begin{eqnarray*}
        (W^K_1)^{\top}W^Q_1 = \begin{bmatrix}
        A_1 & 0 \\ b_1 & 0
        \end{bmatrix},\;(W^K_2)^{\top}W^Q_2 = \begin{bmatrix}
        A_2 & 0 \\ b_2 & 0
        \end{bmatrix}.
    \end{eqnarray*}
    The parameters $W^V_1$, $W^V_2$, $W_{in}$ and $W_{out}$ satisfy
        \begin{eqnarray*}
        f(E)_{d+1,D+1}&=& vm E_{d+1,:}\phi((W_1^KE)^{\top}W_1^QE_{:,D+1})\\
        &&-vn E_{d+1,:}\phi((W_2^KE)^{\top}W_2^QE_{:,D+1}).
    \end{eqnarray*}
    Then the optimal solution satisfies
    that $\|vmA_1-vnA_2\|^2_F=O(1/D)$ and $\|mb_1-nb_2\|^2=O(1/D)$. 
    
   Considering a specific case when $m=2$, $n=1$, $b_1=b_2=0$, and setting $A_1=(c/v)I_d$ for some $0<c<1$, we find that $A_2=((2c-1)/v)I_d$. Consequently, for any $ v^2>\max\{2c^2, 2(2c-1)^2\}$,
    \begin{eqnarray*}
&&L(A_1,A_2,b_1,b_2,v)\\
        &=&\frac{4v^2}{D}\left( \left(\frac{v^2}{v^2-2c^2}\right)^{\frac{d}{2}}- \left(\frac{v^2}{v^2-2c(2c-1)}\right)^{\frac{d}{2}}\right)\\
    &&+\frac{v^2}{D}\left(\frac{v^2}{v^2-2(2c-1)^2}\right)^{\frac{d}{2}} \\ &&+  \frac{(2c-1)^2}{D}\left(\frac{v^2}{v^2-2(2c-1)^2}\right)\left(\frac{v^2}{v^2-2(2c-1)^2}\right)^{\frac{d}{2}} \\
    &&-\frac{(8c-4)c}{D}  \left(\frac{v^2}{v^2-2c(2c-1)}\right)\left(\frac{v^2}{v^2-2c(2c-1)}\right)^{\frac{d}{2}}\\
  &&+  \frac{4c^2}{D}  \left(\frac{v^2}{v^2-2c^2}\right)  \left(\frac{v^2}{v^2-2c^2}\right)^{\frac{d}{2}}+o\left(1/D\right).
    \end{eqnarray*}
    \vspace{-0.2in}
\end{theorem}
In Theorem \ref{thm:multi_head}, the condition $v^2>\max\{2c^2, 2(2c-1)^2\}$ guarantees that $\mathbb{E}(x_q^{\top}Ax_q)$ is finite.
The proof of Theorem~\ref{thm:multi_head} is similar Theorem~\ref{thm:optimal}, and the main difficulty lies in the calculations regarding the cross terms of the two heads.
Details of the proof can be found in Appendix~\ref{sec:proof:multi_head}.

In addition to the formula in Theorem \ref{thm:multi_head}, the following proposition illustrates why the loss of multi-head attention is smaller than the optimal loss of single-head attention:
\begin{proposition}\label{prop:multi}
Following the setting of Theorem~\ref{thm:optimal} and \ref{thm:multi_head}, multi-head attention can be reduced to single-head attention when taking $c=1$, and $c=1$ is not the optimal choice for multi-head attention to achieve the minimal loss.  
\end{proposition}
\vspace{-0.1in}
The proof of Proposition \ref{prop:multi} and some simulations can the found in Appendix~\ref{sec:proof:prop1}. 
\color{black}

\textbf{Simulation.} We also conduct some simulations to compare the prediction loss of single- and multi-head attention. We use the setting in Theorem \ref{thm:multi_head}, i.e., $m=2,n=1$ with $A_1=(c/v)I_d$ and $A_2=(2c-1)I_d/v$. 

\begin{figure}[!ht]
    \centering\vspace{-0.1in}
    \includegraphics[scale=0.65]{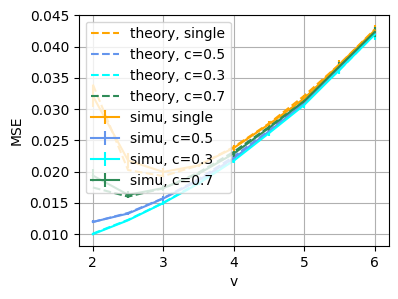}\vspace{-0.2in}
    \caption{ICL performance of multi-head attention with $(m,n)=(2,1)$, $(A_1,A_2,b_1,b_2)=\left((c/v)I_d,((2c-1)/v)I_d,0,0\right)$, and $(d,D)=(5,1000)$.}
    \label{fig:simu_change_c}
    \vspace{-0.15in}
\end{figure}

From Figure \ref{fig:simu_change_c}, we can see that the simulation result is close to the theoretical value for every choice of $(c,v)$. In addition, the MSE of multi-head attention is smaller than that of single-head attention.

\vspace{-0.04in}

\section{Other Scenarios}\label{sec:other}
In addition to the simplest scenario in Section \ref{sec:single_head} and Section \ref{sec:multi_head}, in this section, we relax the data generation model in Assumption \ref{assumption:data} and discuss some other scenarios to understand the corresponding optimal solution for single-head attention, and verify that multi-head attention again gives better ICL performance. In particular, we consider $\theta$ with a non-zero mean (prior knowledge, Section \ref{sec:prior}), noisy response (Section \ref{sec:noisy}), correlated features (Section \ref{sec:correlation}), and local examples $x_i$s given $x_q$ (Section \ref{sec:local}).



\vspace{-0.08in}
\subsection{Prior Knowledge}\label{sec:prior}
From the results in Section \ref{sec:benefit}, the trained transformer only learns to compare the similarity of different examples, rather than learning any particular knowledge from the dataset. 
In this subsection, we explore whether the transformer can learn prior knowledge from the training data where $\theta$ is not fully random.



\begin{assumption}\label{assumption:prior}
    For each prompt, assume that $\theta$ follows $\theta_0+N(0,\sigma^2 I_d/d)$ for some $\|\theta_0\|=\Theta(1)$. The value of $\theta_0$ is the same in all prompts.
\end{assumption}

The following theorem presents how the trained transformer learns $\theta_0$:

\begin{theorem}\label{thm:prior}
    Denote $(W_{out}W^V)_{d+1,:}=[u,v]$ for some vector $u$ and value $v$. Assume there are infinite training prompts. Under Assumption \ref{assumption:prior}, for single-head attention, when $\sigma^2=\Theta(1)$, the population loss is minimized when $\|u\|^2=O(1/D)$, $\|b\|^2=O(1/D)$, and $\|vA-I_d\|_F=O(1/D)$. For the optimal solution $(u^*, b^*, A^*)$ at a fixed $v$ such that $v^2>2$, the population loss is given by
    \begin{eqnarray*}
    &&L(u^*,b^*,A^*,v)\\
    &=&\mathbb{E}\left(y_q- (W_{out}W^V_{d+1,:})^{\top}
    E\phi\left(E^{\top}(W^K)^{\top} W^Q \begin{bmatrix}
        x_q\\0
    \end{bmatrix} \right)\right)^2\\
&=&(\|\theta_0\|^2+\sigma^2)L_{no prior}(A^*, b^*,v)+o(1/D),
\end{eqnarray*}
 where $L_{no prior}(A^*,b^*,v)$ denotes the optimal population loss in Theorem~\ref{thm:optimal}.
 When $\sigma^2=O(1/D)$, there exists infinitely many choices of $u$ such that $\|u\|=\Theta(1)$ and $L(u,b,A,v)=O(1/D)$. The specific conditions are in equation (\ref{eqn:appendix:prior:condition}) in Appendix~\ref{sec:proof:prior}.
For multi-head attention, under the same setting as Theorem \ref{thm:multi_head}, we denote the population loss in Theorem \ref{thm:multi_head} as $L_{no prior}$. Then, when taking $u=0$,
\begin{eqnarray*}
        &&L(A_1,A_2,b_1,b_2,v)\\
        &=&(\|\theta_0\|^2+\sigma^2)L_{no prior}(A_1,A_2,b_1,b_2,v)+o(1/D).
\end{eqnarray*}
   
\end{theorem}

The proof of the theorem is done by computing the partial derivatives of the loss with respect to the parameters and identifying the points where the derivatives equal zero. More details are shown in Appendix \ref{sec:proof:prior} together with some simulation results.

There are several implications from Theorem \ref{thm:prior}. First, when the prior knowledge is weak, i.e., $\sigma^2=\Theta(1)$, the best single-head attention does not learn $\theta_0$. Rather, it still makes predictions by comparing the similarity between $x_q$ and $x_i$s. Second, when the prior knowledge is strong, i.e., $\sigma^2=O(1/D)$, we can obtain good prediction performance when $u$ learns from $\theta_0$. Finally, multi-head attention can still be better than single-head attention. 
\vspace{-0.04in}
\subsection{Noisy Response}\label{sec:noisy}
We consider linear regression tasks with noisy responses, i.e., $y_i=x_i^{\top}\theta+\epsilon_i$ with $\epsilon_i\sim N(0,\sigma_\epsilon^2)$. The following theorem demonstrates the effect of the response noise.
\begin{theorem}\label{thm:noise}
Assume infinite training prompts and $y_i=x_i^{\top}\theta+\epsilon_i$ with $\epsilon_i\sim N(0,\sigma_\epsilon^2)$. The optimal solution of single-head attention satisfies $tr((I_d-A/v)^2)=O(1/D)$ and $\|b\|^2=O(1/D)$.

When taking $A=I_d/v$, where $v^2>2$, and $b=0$,
\begin{eqnarray*}
    &&L(I_d/v,0,v)\\
    &=&\sigma_\epsilon^2+\frac{v^2\sigma_\epsilon^2}{D}\left(\frac{v^2}{v^2-2}\right)^{\frac{d}{2}}+\frac{1}{D}\frac{v^4-v^2}{v^2-2}\left(\frac{v^2}{v^2-2}\right)^{\frac{d}{2}}+o(1/D).
\end{eqnarray*}
For multi-head attention, taking the same parameter values as Theorem \ref{thm:multi_head},
\begin{eqnarray*}
&&L(A_1,A_2,b_1,b_2,v)\\
&=& \frac{4v^2(1+\sigma_\epsilon^2)}{D}\left(\left(\frac{v^2}{v^2-2c^2}\right)^{\frac{d}{2}}-\left(\frac{v^2}{v^2-2c(2c-1)}\right)^{\frac{d}{2}}\right)\\ &&+\frac{v^2(1+\sigma_\epsilon^2)}{D}\left(\frac{v^2}{v^2-2(2c-1)^2}\right)^{\frac{d}{2}} +\sigma_\epsilon^2\\ &&+  \frac{(2c-1)^2}{D}\left(\frac{v^2}{v^2-2(2c-1)^2}\right)\left(\frac{v^2}{v^2-2(2c-1)^2}\right)^{\frac{d}{2}} \\
  &&-\frac{(8c-4)c}{D}  \left(\frac{v^2}{v^2-2c(2c-1)}\right) \left(\frac{v^2}{v^2-2c(2c-1)}\right)^{\frac{d}{2}}\\
  &&+  \frac{4c^2}{D}  \left(\frac{v^2}{v^2-2c^2}\right)  \left(\frac{v^2}{v^2-2c^2}\right)^{\frac{d}{2}}+o(1/D).
\end{eqnarray*}
\end{theorem}
The proof of Theorem \ref{thm:noise} is similar to that of Theorem \ref{thm:optimal} and  \ref{thm:multi_head}, which can be found in Appendix~\ref{sec:proof:noisy}.
Theorem \ref{thm:noise} indicates that the existence of the noise $\epsilon_i$ does not significantly change the optimal solution. For both single- and multi-head attention, there are some additional terms in the prediction loss associated with $\sigma^2_\epsilon$. 

Another difference from the noiseless case is the optimal $v$. Specifically, with a larger $\sigma^2_\epsilon$, the optimal $v$ should ensure $v^2$ is smaller. To explain this, denoting $w_i$ as the attention score for each example $i$, and $w_q$ as the attention score for itself, then the predicted value is $\widehat y_q=\sum_i v w_i(x_i^{\top}\theta+\epsilon_i)=\sum_i v w_ix_i^{\top}\theta+\sum_i v w_i\epsilon_i$, and $Var(\sum_i v w_i\epsilon_i)=v^2\sigma^2_\epsilon\sum w_i^2$. Therefore, a smaller $v^2$ is required to achieve a smaller variance of prediction. 

In terms of the difference between single- and multi-head attention, from the theorem it is evident that multi-head attention is still superior to single-head attention.


\vspace{-0.06in}
\subsection{Correlated Features}\label{sec:correlation}


In this subsection, we consider a scenario where $x$ has some correlated features, i.e. $x\sim N(0,\Sigma)$ for some general $\Sigma\in\mathbb{R}^{d\times d}$.
The following theorem presents the ICL performance of the transformer in this situation.

\begin{theorem}\label{prop:correlated}
    Assume $x\sim N(0,\Sigma)$ and the read-in layer is $W_{in}=\Sigma^{-1/2}$. For single-head attention, when $\mathbb{E}(x_q^{\top}Ax_q)<\infty$, the optimal solution satisfies $\mathbb{E}\theta^{\top}(I_d-vA)^2\theta=O(1/D)$ and $\|b\|^2\mathbb{E}\|\theta\|^4=O(1/D)$ where $\theta\sim N(0,\Sigma^{-1/2}/d)$. For multi-head attention, the best ICL performance is not worse than single-head attention.
\end{theorem}

To show Theorem \ref{prop:correlated}, instead of directly deriving the loss starting from correlated features, we show the equivalence of (1) the problem with correlated features and (2) the problem with isotropic features and a new $\theta$ distribution. Detailed discussions can be found in Appendix~\ref{sec:proof:correlated}.

Theorem \ref{prop:correlated} implies some changes in the prediction loss when considering correlated features. In detail, following the setting of Theorem~\ref{thm:optimal}, i.e., $\theta\sim N(0,I_d/d)$ , $\mathbb{E}\theta^{\top}(I_d-vA)^2\theta=tr((I_d-vA)^2)$. But in Theorem \ref{prop:correlated}, the value of $\mathbb{E}\theta^{\top}(I_d-vA)^2\theta$ depends on the exact distribution of $\theta$. 
However, similar to Theorem \ref{thm:optimal}, we still have $A=I_d/v$ and $b=0$ close to the same optimal solution.




\subsection{Local Examples}\label{sec:local}
While ICL can learn from the examples chosen from the whole population, we are also interested in its efficiency when the in-context samples are selected from the neighbors of $x_q$.

The following two theorems indicate the prediction performance when the prompt is constructed with local examples. In Theorem \ref{thm:local_local}, we consider the scenario where $x_i$s are neighbors of $x_q$ in both training stage and inference stage. In Theorem \ref{thm:local_population}, we consider another scenario with distribution shift: $x_i$s are totally random in the training stage, and are neighbors of $x_q$ in the inference stage. We provide the proof of the two theorems in Appendix~\ref{thm:local_local} and ~\ref{thm:local_population}


\begin{theorem}\label{thm:local_local}
    Assuming that for both training and test prompts, the in-context examples in the prompt are generated from $x_i\sim N(x_q,\sigma_x^2 I_d)$, and the response $y_i=x_i^{\top}\theta$ with $\theta\sim N(0,I_d/d)$. Then when $\mathbb{E}(x_q^{\top}Ax_q)<\infty$, the optimal solution of the single-head transformer satisfies
        \begin{eqnarray*}
        v[\sigma_x^2(A+\theta b^{\top})+I_d] \rightarrow I_d,
    \end{eqnarray*}
    and the minimal population risk is
    \begin{eqnarray*}
        L(A^*, b^*,v)=O({\sigma_x^2}/D)+o({1}/ {D}).
    \end{eqnarray*}
    \vspace{-0.3in}
\end{theorem}
Theorem \ref{thm:local_local} indicates that the optimal solution for local examples is different from the one when $x_i$s are fully random. We do not consider multi-head attention because: (1) if $\sigma_x^2\rightarrow 0$, the single-head attention is effective enough with the overall prediction risk in $o(1/D)$; (2) if $\sigma_x^2$ is large enough, the signal $x_q$ is much smaller than the noise size $\sigma_x$, and the problem is similar to the scenario of Theorem \ref{thm:optimal} and \ref{thm:multi_head}.
Another observation is that, when taking different $\sigma_x^2$s in the training and inference stages, as long as $\sigma_x^2=o(1)$ in the two stages, the ICL in the inference stage can still achieve good performance.

While the above result shows that a small distribution shift in $\sigma_x^2$ does not hurt the ICL performance, the following theorem considers a large distribution shift:

\begin{theorem}
\label{thm:local_population}
Assume the training prompts are sampled in the same way as Theorem \ref{thm:optimal}, i.e., $x_i$s are randomly selected from the whole population. Besides, in the inference stage, in each prompt, $x_q\sim N(0,I_d)$, and the other examples $x_i\sim N(x_q,\sigma_x^2 I_d)$ for some $\sigma_x^2>0$. Then for single-head attention, the prediction loss goes to zero only when $\sigma_x^2+v-1=0$.

\end{theorem}

While Theorem \ref{thm:local_local} demonstrates the benefit of local examples, Theorem \ref{thm:local_population} reveals that ICL may not be consistent when facing distribution shifts.  From simulations in Section \ref{sec:experiment}, the actual $v$ obtained in training does not satisfy $\sigma_x^2+v-1=0$. As a result, it is expected that ICL cannot perform well in such a scenario in general.
\vspace{-0.1in}
\section{Experiments}\label{sec:experiment}

While the simulations in previous sections directly calculate the prediction loss of ICL given specific parameter weights, in this section, we conduct experiments starting from training the transformer. Due to the page limit, we postpone the experiments for noisy response and correlated features to Appendix~\ref{apd:experiment}.
\vspace{-0.04in}
\subsection{Experimental Settings}

We modify the implementation of \citet{garg2022can} to conduct the experiments. In particular, we change the input format in \citet{garg2022can} and use the format $E$ defined in (\ref{eqn:input}). In each training iteration, we generate a new batch of 64 prompts to train the transformer. In terms of the loss to be minimized during the training, we use the one defined in (\ref{eqn:loss}), i.e., we optimize the loss between $y_q$ and $\widehat{y}_q$. We train the transformers with 500k iterations and use Adam optimizer with 0.0001 learning rate. 

In the inference stage, we randomly sample 1280 prompts to obtain the average and error bar of the loss. Instead of only using $x_q$ to calculate the loss, for each in-context example $i\in[D]$, we also make the ICL prediction and calculate the corresponding loss. 

\subsection{Single-head vs Multi-head}

\begin{figure}[!ht]
    \centering
    \vspace{-0.05in}
    \includegraphics[width=0.35
  \textwidth]{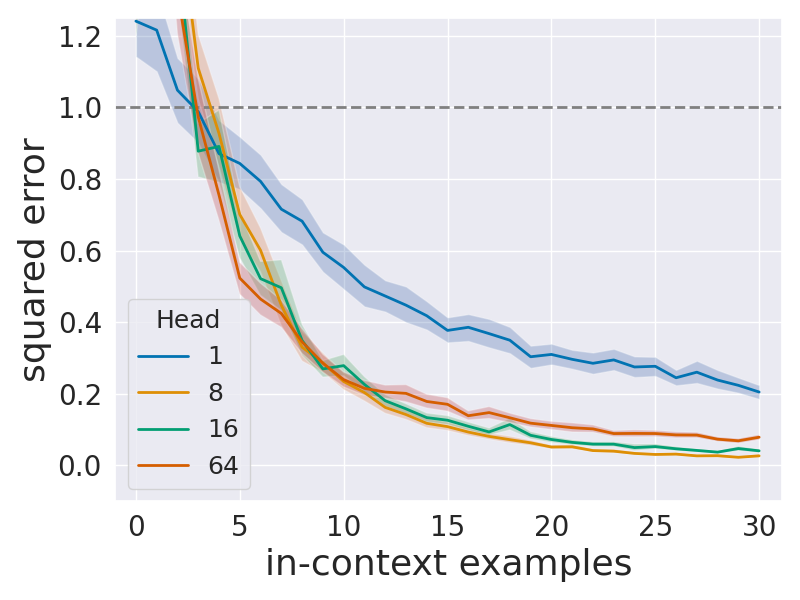}\vspace{-0.15in}
    \caption{A comparison between single-head and multi-head with the input embedding dimension $p=64$.}
    \label{fig:simu_min_c}
\end{figure}

In the experiment, we compare the performance of single-head and multi-head attention. We set the input embedding dimension to $p=64$, the dimension of in-context examples to $d=5$, and vary the number of heads for analysis. The results are summarized in Figure \ref{fig:simu_min_c}.

Figure \ref{fig:simu_min_c} shows that single-head attention has a worse ICL performance than multi-head attention. In addition, although our theorems do not consider such a scenario, for multi-head attention, when $h$ is too large so that $p/h<d$, the ICL performance can be affected. When taking $h=64$, the ICL performance gets worse.

\vspace{-0.05in}

\begin{figure}[!ht]
    \centering
    \vspace{-0.05in}
    \includegraphics[width=0.3
  \textwidth]{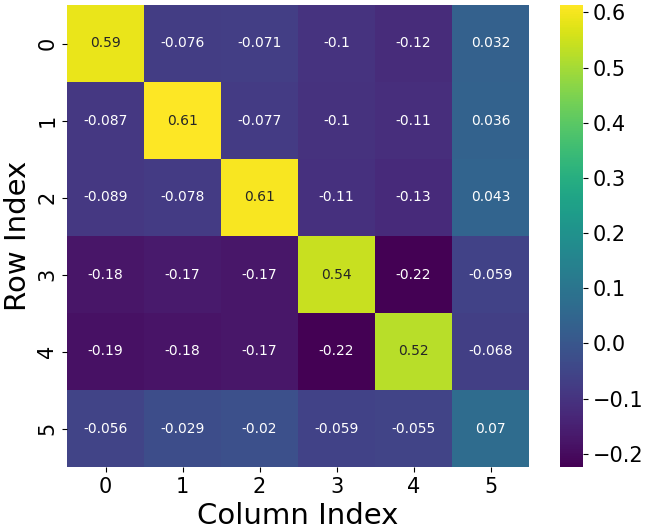}\vspace{-0.15in}
    \caption{An illustration of  the matrix $(W^K)^{\top}W^Q$ for the no read-in case. It is expected to be some kinds of $\alpha I_d$. {4 of 10 trials are like this.}}
    \label{fig:kq_1}
\end{figure}

\begin{figure}[!ht]
    \centering
    \includegraphics[width=0.3
  \textwidth]{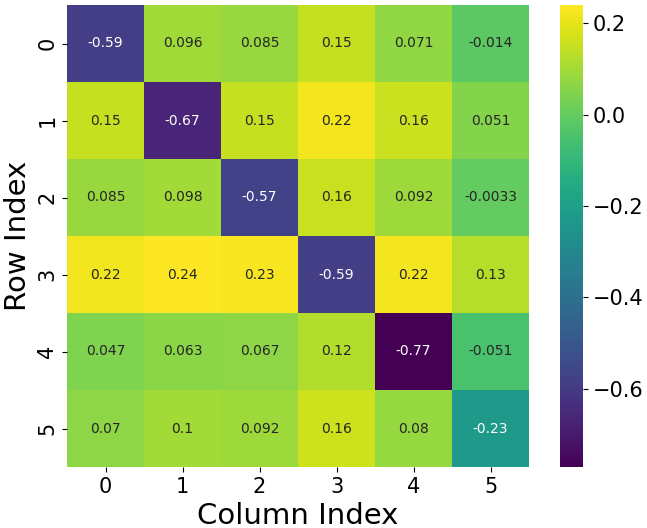}\vspace{-0.08in}
    \caption{An illustration of the matrix $(W^K)^{\top}W^Q$ for the no read-in case. It is expected to be some kinds of $\alpha I_d$. { 5 of 10 trials are like this.}}
    \label{fig:kq_2}
    \vspace{-0.15in}
\end{figure}

In addition to the ICL performance, we also conduct another experiment to examine $(W^K)^{\top}W^Q$. We remove the read-in layer, train the transformer, and print out $(W^K)^{\top}W^Q$. We repeat the experiment 10 times to see the value of $(W^K)^{\top}W^Q$. As in Theorem \ref{thm:optimal}, for single-head attention without the read-in layer, $(W^K)^{\top}W^Q$ is expected to be in the form of $I/v$ when $v^2>2$. In the 10 trials, 9 of them observe such a result, where 5 trials have $v>0$ as in Figure \ref{fig:kq_1} and 4 trials have $v<0$ as in Figure \ref{fig:kq_2}.
We also visualize the attention score corresponding to these two cases in Figure~\ref{fig:atten_score} (See Appendix~\ref{apd:experiment}). These results indicates that the theoretical global minimum is highly likely to be attained in the real practice of transformer training.

\vspace{-0.05in}
\subsection{Input Embedding Dimension}

As mentioned in Section \ref{sec:multi_head}, increasing the input embedding dimension $p$ provides the flexibility of multi-head attention to achieve better ICL performance. In this experiment, we change $p$ to examine the performance.

\begin{figure}[!ht]
    \centering
    \vspace{-0.2in}\includegraphics[width=0.35
  \textwidth]{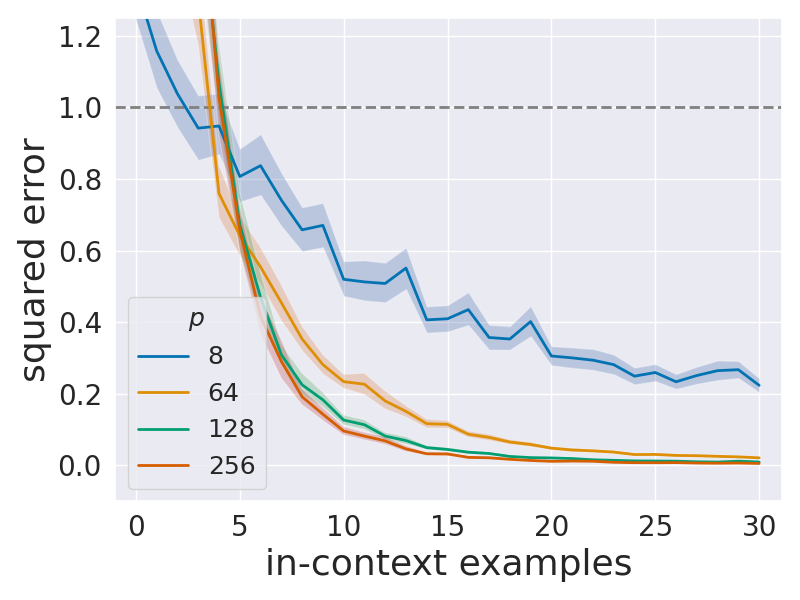}
  \vspace{-0.15in}
    \caption{Results of fixing the dimension allocated in each head ($p/h=8$), and increasing the number of heads.}
    \vspace{-0.15in}
    \label{fig:multi_attention}
\end{figure}

In Figure \ref{fig:multi_attention}, we fix the dimension in each head ($p/h$), and increase $h$. We can observe that when the dimension is sufficient, the increasing $h$ leads to a smaller prediction loss.

In addition, we also run different $p/h$ for different $p$. As shown in Table \ref{tab:multi_p}, we can also see that for all $p=64,128,256$, the following setting gives good ICL performance: (1) $p/h\geq d$ and (2) $h$ is as large as possible. 
 
\begin{table}[!ht]
\vspace{-0.1in}
    \caption{Different choices of head.}
    \centering
  \vspace{0.03in}
    \resizebox{0.48\textwidth}{!}{
    \begin{tabular}{c|c|c|c|c|c|c|c}
    \toprule
    $p$ & $h$ & $p/h$ & ICL & $p$ & $h$ & $p/h$ & ICL\\ 
    \midrule
    \multirow{4}{*}{6} & 1 & 6 & 0.41878 &\multirow{4}{*}{64} & 1 & 64 & 0.18983 \\
    & 2 & 3 & 0.29825 &  & 8 & 8 &0.00769\\
    & 3 & 2 & 0.58036 & & 16 & 4 &0.01724\\
    & 6 & 1 & 0.56292 & & 64 & 1 &0.04899\\ 
    \midrule
    \multirow{6}{*}{128} & 1 & 128 & 0.16619 &\multirow{6}{*}{256} & 1 & 256 & 0.16141 \\
    & 8 & 16 & 0.00577 &  & 8 & 32 & 0.00587\\
    & 16 & 8 & 0.00244 & & 16 & 16 & 0.00144 \\
    & 64 & 2 & 0.00611 & & 64 & 4 & 0.00134 \\
    & 128 & 1 & 0.01254 & & 128 & 2 & 0.00159 \\
    &  &  & & & 256 & 1 & 0.00549 \\
    \bottomrule
    \end{tabular}}
    \label{tab:multi_p}
    \vspace{-0.15in}
\end{table}

\subsection{Prior Knowledge}

\begin{figure}[!ht]
    \centering\vspace{-0.1in}
    \includegraphics[width=0.35
  \textwidth]{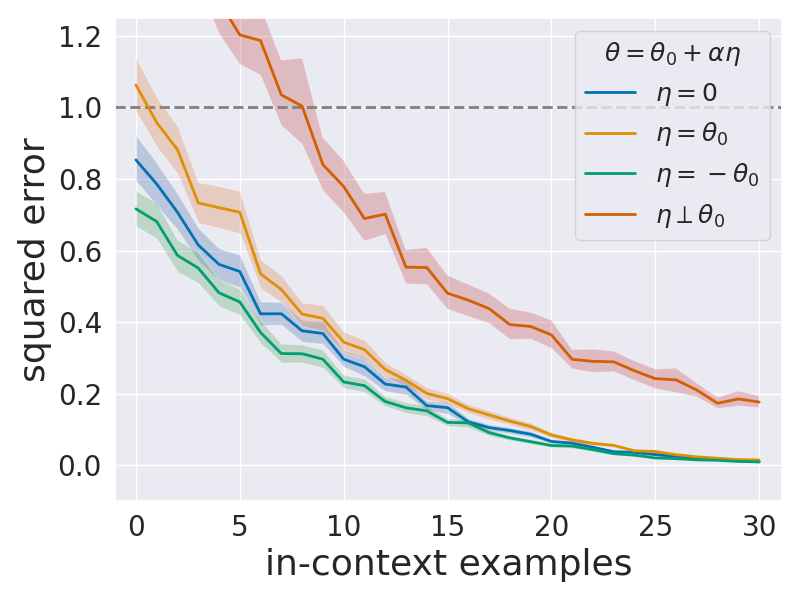}
  \vspace{-0.15in}
    \caption{Head 1 prior knowledge. More results can be found from Figure~\ref{fig:append_prior1} in Appendix~\ref{apd:fig}.}
    \label{fig:head1_prior}
    \vspace{-0.15in}
\end{figure}

\begin{figure}[!ht]
\vspace{-0.06in}
    \centering
    \includegraphics[width=0.35
  \textwidth]{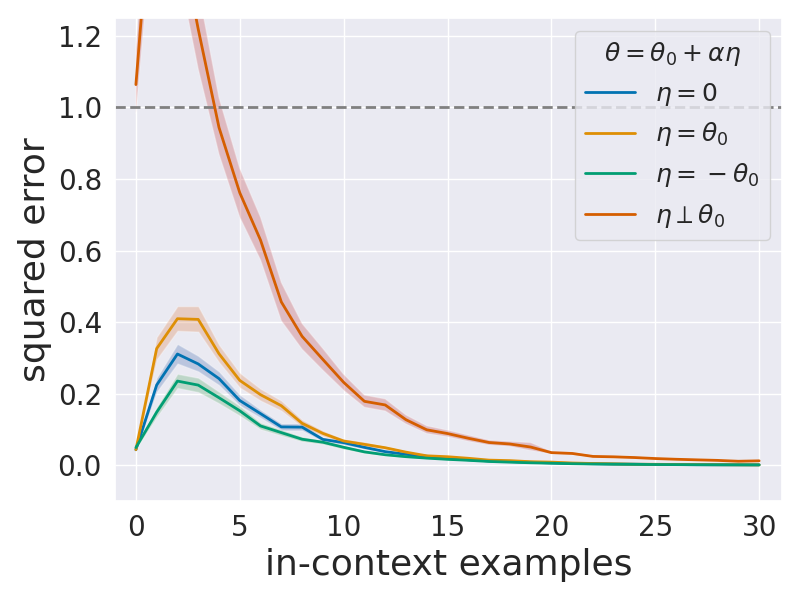}
  \vspace{-0.15in}
    \caption{Head 16 prior knowledge. More results can be found in Appendix~\ref{apd:experiment} Figure~\ref{fig:append_prior2}.}
    \vspace{-0.15in}
    \label{fig:head16_prior}
\end{figure}

In the experiment about prior knowledge, we study the inference-stage performance under different choices of $\theta$. Before training, we randomly generate a $\theta_0\sim N(0,I_d)$. During the training, to generate one training prompt, we generate $\theta=\theta_0+N(0,\sigma^2 I_d/d)$, and then generate the examples $(x_i,y_i)$ based on $\theta$. In the test stage, we generate different prompts following different $\theta$. The prediction results can be found in Figure~\ref{fig:head1_prior} for single-head attention and Figure~\ref{fig:head16_prior} for multi-head attention with $\sigma^2=1$ and $\alpha=0.1$.

We make the following observations. First, comparing Figure~\ref{fig:head1_prior} with Figure~\ref{fig:head16_prior}, we note that multi-head attention gives better ICL performance than single-head attention.
Second, as shown in Figure~\ref{fig:head1_prior}  and Figure~\ref{fig:head16_prior}, when $\theta\parallel\theta_0$, a smaller $\|\theta\|$ implies better ICL performance. To explain this, since the ICL loss is in $O(1/D)$, a smaller $\|\theta\|$ indicates less variation among the response of different examples; thus, the multiplicative constant of the $O(1/D)$ is smaller. 
Finally, comparing $\eta\perp\theta_0$ with $\eta=\theta_0$, although $\|\theta\|$ for $\eta\perp\theta_0$ is smaller, the ICL performance is worse. This observation implies that the transformer learns the prior knowledge $\theta_0$.

\subsection{Local Examples}
As discussed in Theorem \ref{thm:local_local} and \ref{thm:local_population}, when both the training and inference stage use local examples with the same distribution (i.e., same $\sigma_x^2$), ICL leads to consistent predictions. When there is a large distribution shift, the prediction is not consistent. 

In Table \ref{tab:neighbor}, we demonstrate the ICL performance in the inference stage with local examples. As expected, the prediction is more accurate when the training and testing data have the same distribution, with a diminishing $\sigma_x^2$. On the other hand, when training with fully random prompts (i.e., not local examples), the prediction is inconsistent.

\begin{table}[!ht]
\centering\vspace{-0.1in}
    \caption{ICL performance for local examples in the inference stage with/without distribution shift in the training data. ``Fully random": not local examples. More results can be found in Appendix~\ref{fig:append_local}.}
    \vspace{0.04in}
    \resizebox{0.48\textwidth}{!}{
    \begin{tabular}{c|c|c|c}
    \toprule
    \multirow{2}{*}{Training} & \multirow{2}{*}{Testing} & \multicolumn{2}{c}{ICL} \\ \cmidrule{3-4}
 & & 1 head & 16 heads \\
     \midrule
     \multirow{3}{*}{Same as testing} & $\sigma_x^2=1^2$ & 0.01464 & 0.00285 \\
     & $\sigma_x^2=0.1^2$ & 0.00049 & 0.00096 \\
     & $\sigma_x^2=0.01^2$ & 2.50e-05 & 9.79e-06 \\
     \midrule
     \multirow{3}{*}{Fully random} & $\sigma_x^2=1^2$ & 0.29317 & 0.60400 \\
     & $\sigma_x^2=0.1^2$ & 0.39023 & 1.23142 \\
     & $\sigma_x^2=0.01^2$ & 0.41253 & 1.12165 \\
     \bottomrule
    \end{tabular}}
    \label{tab:neighbor}
\end{table}

\vspace{-0.15in}
\section{Conclusion}

This study explicitly calculates the ICL performance in linear regression tasks to show that multi-head attention is preferred over single-head attention. In addition to the simplest case of noiseless regression, we extend the analysis to other scenarios. When the data contain prior knowledge, a transformer that learns the prior knowledge can perform well in ICL prediction. When the examples in the prompt are neighbors of $x_q$, the ICL prediction can be very efficient if there is no distribution shift.

There are several future directions. First, our current study considers the case for large enough $D$. We may consider relaxing this condition and studying the finite-example scenario. Second, although we consider different scenarios of the data, we always consider linear regression tasks. We may extend the analysis to other problems such as non-parametric models. Finally, we assume that the training dataset has almost infinite samples and directly study the population loss. We may extend it to a finite-prompt scenario and investigate the generalization performance.

\bibliographystyle{plainnat}
\section{Impact Statements}
This paper presents work whose goal is to advance the field of Machine Learning via deepening the theoretical understanding of existing neural network architectures. This paper does not introduce new methodology or new datasets. Therefore, there is no extra ethical impact or societal implication which is worth special emphasis here.

\bibliography{ref}

\appendix
\onecolumn

\section{Proofs}

\subsection{Theorem \ref{thm:optimal}}\label{sec:proof:optimal}
\begin{proof}[Proof of Theorem \ref{thm:optimal}]
When taking infinite many training samples (prompts), the loss function becomes
    \begin{eqnarray*}
    &&\mathbb{E}\left(y_q- (W_{out}W^V_{d+1,:})^{\top}
    E\phi\left(E^{\top}(W^K)^{\top} W^Q \begin{bmatrix}
        x_q\\0
    \end{bmatrix} \right)\right)^2\\
    &=&\mathbb{E}\left(y_q- v \begin{bmatrix}
        y_1,y_2,\ldots,y_D,0
    \end{bmatrix}\phi\left(E^{\top}(W^K)^{\top} W^Q \begin{bmatrix}
        x_q\\0
    \end{bmatrix} \right)\right)^2\\
    &=&\mathbb{E}_{(x_q,\theta)}\mathbb{E}_{\{x_i\}_{i\in[D]}}\left(y_q- v\begin{bmatrix}
        y_1,y_2,\ldots,y_D,0
    \end{bmatrix}\phi\left(\begin{bmatrix}
        x_1^{\top}Ax_q+y_1b^{\top}x_q\\\ldots\\x_q^{\top}A x_q+0
    \end{bmatrix} \right)\right)^2\\
    &=&\mathbb{E}_{(x_q,\theta)}\mathbb{E}_{\{x_i\}_{i\in[D]}}
    \left(y_q- \frac{ v\sum_{i=1}^D 
    \theta^{\top} x_i\exp(x_i^{\top}Ax_q+y_ib^{\top}x_q) }{\sum \exp(x_i^{\top}Ax_q+y_ib^{\top}x_q) 
    +\exp(x_q^{\top}Ax_q)}
    \right)^2 \\ &=&\mathbb{E}_{(x_q,\theta)}\mathbb{E}_{\{x_i\}_{i\in[D]}}\left(y_q^2
    \underbrace{-2y_q\left(\frac{v\sum_{i=1}^D 
    \theta^{\top} x_i\exp(x_i^{\top}Ax_q+y_ib^{\top}x_q) }{\sum \exp(x_i^{\top}Ax_q+y_ib^{\top}x_q)
    +\exp(x_q^{\top}Ax_q)}\right)}_{=A_1}+ \underbrace{\left(\frac{ v \sum_{i=1}^D 
    \theta^{\top} x_i\exp(x_i^{\top}Ax_q+y_ib^{\top}x_q) }{\sum \exp(x_i^{\top}Ax_q+y_ib^{\top}x_q)+\exp(x_q^{\top}Ax_q)}\right)^2}_{=A_2}
    \right).
    \end{eqnarray*}
    When $D\rightarrow\infty$, we have
    \begin{eqnarray*}
    &&\mathbb{E}_{\{x_i,y_i\}_{i\in[D]}}A_1\\
    &=& \mathbb{E}_{\{x_i,y_i\}_{i\in[D]}}\frac{(-2v\theta^\top x_q) \sum_{i=1}^D 
    \theta^{\top} x_i\exp(x_i^{\top}Ax_q+y_ib^{\top}x_q) }{\sum \exp(x_i^{\top}Ax_q+y_ib^{\top}x_q) 
    +\exp(x_q^{\top}Ax_q)-D\mathbb{E}\exp({x_1}^{\top}Ax_q+y_1b^{\top}x_q)+D\mathbb{E}\exp({x_1}^{\top}Ax_q+y_1b^{\top}x_q)}\\
    &=&\mathbb{E}_{\{x_i,y_i\}_{i\in[D]}}\frac{(-2v\theta^\top x_q) \sum_{i=1}^D 
    \theta^{\top} x_i\exp(x_i^{\top}Ax_q+y_ib^{\top}x_q) }{\exp(x_q^{\top}Ax_q)+D\mathbb{E}\exp({x_1}^{\top}Ax_q+y_1b^{\top}x_q)}\\
    &&+ \mathbb{E}_{\{x_i,y_i\}_{i\in[D]}}\frac{(2v\theta^\top x_q) \sum_{i=1}^D 
    \theta^{\top} x_i\exp(x_i^{\top}Ax_q+y_ib^{\top}x_q) }{(\exp(x_q^{\top}Ax_q)+D\mathbb{E}\exp({x_1}^{\top}Ax_q+y_1b^{\top}x_q))^2}\left(\sum \exp(x_i^{\top}Ax_q+y_ib^{\top}x_q)-D\mathbb{E}\exp({x_1}^{\top}Ax_q+y_1b^{\top}x_q)\right)\\ &&-\mathbb{E}_{\{x_i,y_i\}_{i\in[D]}}\frac{(2v\theta^\top x_q) \sum_{i=1}^D 
    \theta^{\top} x_i\exp(x_i^{\top}Ax_q+y_ib^{\top}x_q) }{(\exp(x_q^{\top}Ax_q)+D\mathbb{E}\exp({x_1}^{\top}Ax_q+y_1b^{\top}x_q))^3}\left(\sum \exp(x_i^{\top}Ax_q+y_ib^{\top}x_q)-D\mathbb{E}\exp({x_1}^{\top}Ax_q+y_1b^{\top}x_q)\right)^2\\
    &&+o(\frac{1}{D})\\
    &=&A_{11}+A_{12}+A_{13}+o(\frac{1}{D}).
    \end{eqnarray*}
    When taking expectation w.r.t. $\{x_i,y_i\}_{i\in[D]}$, we have
    \begin{eqnarray*}
    \mathbb{E}_{\{x_i,y_i\}_{i\in[D]}}\frac{\sum_{i=1}^D 
    \theta^{\top} x_i\exp(x_i^{\top}Ax_q+y_ib^{\top}x_q) }{\exp(x_q^{\top}Ax_q)+D\mathbb{E}\exp({x_1}^{\top}Ax_q+y_1b^{\top}x_q)}=\mathbb{E}_{\{x_1,y_1\}}\frac{D\mathbb{E}\theta^{\top}x_1\exp(x_1^{\top}Ax_q+y_1b^{\top}x_q)}{\exp(x_q^{\top}Ax_q)+D\mathbb{E}_{x_1}\exp(x_1^{\top}Ax_q+y_1b^{\top}x_q)},
    \end{eqnarray*}
    where
    \begin{eqnarray*}
        \mathbb{E}_{\{x_1,y_1\}}\exp(x_1^{\top}Ax_q+y_1b^{\top}x_q)&=& \exp(x_q^{\top}(A+\theta b^{\top})^{\top}(A+\theta b^{\top}) x_q/2),\\
        \mathbb{E}_{\{x_1,y_1\}}x_1\exp(x^{\top}Ax_q+y_1b^{\top}x_q)&=&\mathbb{E}\frac{\partial }{\partial ((A+\theta b^{\top})x_q)}\exp(x_1^{\top}Ax_q+y_1b^{\top}x_q)\\
        &=&(A+\theta b^{\top})x_q\exp(x_q^{\top}(A+\theta b^{\top})^{\top}(A+\theta b^{\top})x_q/2).
    \end{eqnarray*}
    Therefore, we have
    \begin{eqnarray*}
A_{11}&=&\mathbb{E}_{\{x_1,y_1\}}\left(-\frac{D(2v\theta^\top x_q)\mathbb{E}\theta^{\top}x_1\exp(x_1^{\top}Ax_q+y_1b^{\top}x_q)}{\exp(x_q^{\top}Ax_q)+D\mathbb{E}_{x_1}\exp(x_1^{\top}Ax_q+y_1b^{\top}x_q)}\right)\\
    &=&\mathbb{E}_{\{x_1,y_1\}}\left(-\frac{D(2v\theta^\top x_q)\mathbb{E}\theta^{\top}x_1\exp(x_1^{\top}Ax_q+y_1b^{\top}x_q)}{D\mathbb{E}_{x_1}\exp(x_1^{\top}Ax_q+y_1b^{\top}x_q)}+\frac{D(2v\theta^\top x_q)\mathbb{E}\theta^{\top}x_1\exp(x_1^{\top}Ax_q+y_1b^{\top}x_q)\exp(x_q^{\top}Ax_q)}{(D\mathbb{E}_{x_1}\exp(x_1^{\top}Ax_q+y_1b^{\top}x_q))^2}\right)\\
    &+&\underbrace{\mathbb{E}_{\{x_1,y_1\}}\left(-\frac{D(2v\theta^\top x_q)\mathbb{E}\theta^{\top}x_1\exp(x_1^{\top}Ax_q+y_1b^{\top}x_q)\exp(2x_q^{\top}Ax_q)}{(D\mathbb{E}_{x_1}\exp(x_1^{\top}Ax_q+y_1b^{\top}x_q))^3}\right)}_{=o(1/D)}\\
    &=&-\frac{D(2v\theta^\top x_q)\theta^{\top}(A+\theta b^{\top})x_q\exp(x_q^{\top}(A+\theta b^{\top})^{\top}(A+\theta b^{\top})x_q/2)}{D\exp(x_q^{\top}(A+\theta b^{\top})^{\top}(A+\theta b^{\top})x_q/2)} \\ 
    && +\frac{D(2v\theta^\top x_q)\theta^{\top}(A+\theta b^{\top})x_q\exp(x_q^{\top}(A+\theta b^{\top})^{\top}(A+\theta b^{\top})x_q/2)\exp(x_q^{\top}Ax_q)}{D^2\exp(x_q^{\top}(A+\theta b^{\top})^{\top}(A+\theta b^{\top})x_q)} +o(\frac{1}{D}) \\
    &=&-(2v\theta^\top x_q)\theta^{\top}(A+\theta b^{\top})x_q + \frac{(2v\theta^\top x_q)\theta^{\top}(A+\theta b^{\top})x_q\exp(x_q^{\top}Ax_q)}{D\exp(x_q^{\top}(A+\theta b^{\top})^{\top}(A+\theta b^{\top})x_q/2)} +o(\frac{1}{D}), \\
    \end{eqnarray*}
    
        \begin{eqnarray*}
       A_{12} &=&\mathbb{E}_{\{x_i,y_i\}_{i\in[D]}}\frac{(2v\theta^\top x_q \sum_{i=1}^D 
    \theta^{\top} x_i\exp(x_i^{\top}Ax_q+y_ib^{\top}x_q))(\sum \exp(x_i^{\top}Ax_q+y_ib^{\top}x_q)) }{(\exp(x_q^{\top}Ax_q)+D\mathbb{E}\exp({x_1}^{\top}Ax_q+y_1b^{\top}x_q))^2}\\
    &&-\mathbb{E}_{\{x_i,y_i\}_{i\in[D]}}\frac{(2v\theta^\top x_q\sum_{i=1}^D 
    \theta^{\top} x_i\exp(x_i^{\top}Ax_q+y_ib^{\top}x_q))(D\mathbb{E}\exp({x_1}^{\top}Ax_q+y_1b^{\top}x_q)) }{(\exp(x_q^{\top}Ax_q)+D\mathbb{E}\exp({x_1}^{\top}Ax_q+y_1b^{\top}x_q))^2}\\
     &=&\frac{(2Dv\theta^\top x_q)\mathbb{E}
    \theta^{\top} x_1\exp(2(x_1^{\top}Ax_q+y_1b^{\top}x_q)))}{(\exp(x_q^{\top}Ax_q)+D\mathbb{E}\exp({x_1}^{\top}Ax_q+y_1b^{\top}x_q))^2}+\frac{(2D(D-1)v \theta^\top x_q\mathbb{E}_{x_1,x_2} \theta^{\top}x_1\exp(x_1^{\top}(A+\theta b^{\top})x_q+x_2^{\top}(A+\theta b^{\top})x_q) }{(\exp(x_q^{\top}Ax_q)+D\mathbb{E}\exp({x_1}^{\top}Ax_q+y_1b^{\top}x_q))^2} \\
    &&-\mathbb{E}_{\{x_i,y_i\}_{i\in[D]}}\frac{(2v\theta^\top x_q\sum_{i=1}^D 
    \theta^{\top} x_i\exp(x_i^{\top}Ax_q+y_ib^{\top}x_q))(D\mathbb{E}\exp({x_1}^{\top}Ax_q+y_1b^{\top}x_q)) }{(\exp(x_q^{\top}Ax_q)+D\mathbb{E}\exp({x_1}^{\top}Ax_q+y_1b^{\top}x_q))^2}\\
    &=&A_{121}+A_{122}+A_{123}, 
        \end{eqnarray*}
        where
    \begin{eqnarray*}
     \\
    A_{121}&=&\frac{4Dv(\theta^\top x_q)\theta^{\top}(A+\theta b^{\top})x_q\exp(2x_q^{\top}(A+\theta b^{\top})^{\top}(A+\theta b^{\top})x_q)}{(\exp(x_q^{\top}Ax_q)+D\exp(x_q^{\top}(A+\theta b^{\top})^{\top}(A+\theta b^{\top})x_q/2)^2} \\
        &=& \frac{4Dv(\theta^\top x_q)\theta^{\top}(A+\theta b^{\top})x_q\exp(2x_q^{\top}(A+\theta b^{\top})^{\top}(A+\theta b^{\top})x_q)}{(D\exp(x_q^{\top}(A+\theta b^{\top})^{\top}(A+\theta b^{\top})x_q/2)^2}+o(\frac{1}{D}) \\
        &=& \frac{4v}{D}(\theta^\top x_q )\theta^{\top}(A+\theta b^{\top})x_q\exp(x_q^{\top}(A+\theta b^{\top})^{\top}(A+\theta b^{\top})x_q)+o(\frac{1}{D}),
\end{eqnarray*} 
   \begin{eqnarray*}
  A_{122} &=&
      \frac{2(D(D-1)v(\theta^\top x_q) \theta^{\top}(A+\theta b^{\top})x_q\exp(x_q^{\top}(A+\theta b^{\top})^{\top}(A+\theta b^{\top})x_q)}{(\exp(x_q^{\top}Ax_q)+D\mathbb{E}\exp({x_1}^{\top}Ax_q+y_1b^{\top}x_q))^2} \\
    &=& 2v(\theta^\top x_q)\theta^{\top}(A+\theta b^{\top})x_q
    -\frac{4Dv(\theta^{\top} x_q)\theta^{\top}(A+\theta b^{\top})x_q \exp(x_q^{\top}Ax_q)\exp(x_q^{\top}(A+\theta b^{\top})^{\top}(A+\theta b^{\top})x_q/2)}{(\exp(x_q^{\top}Ax_q)+D\mathbb{E}\exp({x_1}^{\top}Ax_q+y_1b^{\top}x_q))^2} \\
    &&  -\frac{2Dv(\theta^\top x_q)\theta^{\top}(A+\theta b^{\top})x_q \exp(x_q^{\top}(A+\theta b^{\top})^{\top}(A+\theta b^{\top})x_q)}{(\exp(x_q^{\top}Ax_q)+D\mathbb{E}\exp({x_1}^{\top}Ax_q+y_1b^{\top}x_q))^2}-\frac{2v(\theta^{\top}x_q)\theta^{\top}(A+\theta b^{\top})x_q \exp(2x_q^{\top}Ax_q)}{(\exp(x_q^{\top}Ax_q)+D\mathbb{E}\exp({x_1}^{\top}Ax_q+y_1b^{\top}x_q))^2} \\
    &=& 2v(\theta^\top x_q)\theta^{\top}(A+\theta b^{\top})x_q
    - \frac{4v(\theta^\top x_q)\theta^{\top}(A+\theta b^{\top})x_q \exp(x_q^{\top}Ax_q)\exp(x_q^{\top}(A+\theta b^{\top})^{\top}(A+\theta b^{\top})x_q/2)}{(D\mathbb{E}\exp({x_1}^{\top}Ax_q+y_1b^{\top}x_q))^2}   \\
     && -(\frac{ 2Dv(\theta^\top x_q)\theta^{\top}(A+\theta b^{\top})x_q \exp(x_q^{\top}(A+\theta b^{\top})^{\top}(A+\theta b^{\top})x_q)}{(D\mathbb{E}\exp({x_1}^{\top}Ax_q+y_1b^{\top}x_q))^2} +o(\frac{1}{D})\\
     &=&  2v(\theta^\top x_q)\theta^{\top}(A+\theta b^{\top})x_q
    - \frac{ 4v(\theta^\top x_q)\theta^{\top}(A+\theta b^{\top})x_q \exp(x_q^{\top}Ax_q)}{(D\exp(x_q^{\top}(A+\theta b^{\top})^{\top}(A+\theta b^{\top})x_q/2))} -\frac{2v}{D}(\theta^\top x_q)\theta^{\top}(A+\theta b^{\top})x_q +o(\frac{1}{D}),
     \\
\end{eqnarray*} 

 \begin{eqnarray*}
A_{123}&=&-\frac{( 2Dv(\theta^\top x_q) 
    \theta^{\top} (A+\theta b^{\top})x_q\exp(x_q^{\top}(A+\theta b^{\top})^{\top}(A+\theta b^{\top})x_q/2))(D\mathbb{E}\exp({x_1}^{\top}Ax_q+y_1b^{\top}x_q)) }{\exp(x_q^{\top}Ax_q)+D\exp(x_q^{\top}(A+\theta b^{\top})^{\top}(A+\theta b^{\top})x_q/2)^2} \\
&=&-\frac{2D^2 v(\theta^\top x_q)
    \theta^{\top} (A+\theta b^{\top})x_q\exp(x_q^{\top}(A+\theta b^{\top})^{\top}(A+\theta b^{\top})x_q) }{\exp(x_q^{\top}Ax_q)+D\exp(x_q^{\top}(A+\theta b^{\top})^{\top}(A+\theta b^{\top})x_q/2)^2} \\
&=&  -2v(\theta^\top x_q)\theta^{\top}(A+\theta b^{\top})x_q +\frac{(4D v(\theta^\top x_q)
    \theta^{\top} (A+\theta b^{\top})x_q\exp(x_q^{\top}Ax_q) \exp(x_q^{\top}(A+\theta b^{\top})^{\top}(A+\theta b^{\top})x_q)/2) }{(\exp(x_q^{\top}Ax_q)+D\exp(x_q^{\top}(A+\theta b^{\top})^{\top}(A+\theta b^{\top})x_q/2)^2} + o(\frac{1}{D})  \\
&=&  -2v(\theta^\top x_q)\theta^{\top}(A+\theta b^{\top})x_q +\frac{(4D v(\theta^\top x_q)
    \theta^{\top} (A+\theta b^{\top})x_q\exp(x_q^{\top}Ax_q) \exp(x_q^{\top}(A+\theta b^{\top})^{\top}(A+\theta b^{\top})x_q)/2) }{(D\exp(x_q^{\top}(A+\theta b^{\top})^{\top}(A+\theta b^{\top})x_q/2)^2} + o(\frac{1}{D})  \\ 
&=& -2v(\theta^\top x_q)\theta^{\top}(A+\theta b^{\top})x_q +\frac{4v(\theta^\top x_q)
    \theta^{\top} (A+\theta b^{\top})x_q\exp(x_q^{\top}Ax_q) }{D\exp(x_q^{\top}(A+\theta b^{\top})^{\top}(A+\theta b^{\top})x_q/2)} + o(\frac{1}{D}),
\\
\end{eqnarray*} 
and
\begin{eqnarray*}
    A_{13}&=& -\frac{2D^2v(\theta^\top x_q)\sum_{i=1}^D 
    \theta^{\top} x_i\exp(x_i^{\top}Ax_q-y_ib^{\top}x_q)(\mathbb{E}\exp({x_1}^{\top}Ax_q+y_1b^{\top}x_q))^2}{(\exp(x_q^{\top}Ax_q)+D\mathbb{E}\exp({x_1}^{\top}Ax_q+y_1b^{\top}x_q))^3}\\
    && - \frac{2Dv(\theta^\top x_q)\mathbb{E}
_{x_1} \theta^\top x_1\exp(3{x_1}^{\top}Ax_q+3y_1b^{\top}x_q)}{(\exp(x_q^{\top}Ax_q)+D\mathbb{E}\exp({x_1}^{\top}Ax_q+y_1b^{\top}x_q))^3} \\
 &&  -\frac{2D(D-1) (\theta^\top x_q) \mathbb{E}
_{x_1,x_2} \theta^\top x_1\exp({x_1}^{\top}Ax_q+y_1b^{\top}x_q)\exp({2x_2}^{\top}Ax_q+2y_2b^{\top}x_q)}{(\exp(x_q^{\top}Ax_q)+D\mathbb{E}\exp({x_1}^{\top}Ax_q+y_1b^{\top}x_q))^3})  \\
 && +\frac{4D^2 v(\theta^\top x_q)\mathbb{E}
_{x_1} \theta^\top x_1\exp(2{x_1}^{\top}Ax_q+2y_1b^{\top}x_q)\mathbb{E}\exp({x_1}^{\top}Ax_q+y_1b^{\top}x_q)}{(\exp(x_q^{\top}Ax_q)+D\mathbb{E}\exp({x_1}^{\top}Ax_q+y_1b^{\top}x_q))^3}  \\ 
 && +\frac{4D^2(D-1) v(\theta^\top x_q)\mathbb{E}
_{x_1,x_2} \theta^\top x_1\exp({x_1}^{\top}Ax_q+y_1b^{\top}x_q)\exp({x_2}^{\top}Ax_q+y_2b^{\top}x_q)\mathbb{E}\exp({x_1}^{\top}Ax_q+y_1b^{\top}x_q)}{(\exp(x_q^{\top}Ax_q)+D\mathbb{E}\exp({x_1}^{\top}Ax_q+y_1b^{\top}x_q))^3}  \\ 
 && -\frac{4D(D-1) v(\theta^\top x_q)\mathbb{E}
_{x_1} \theta^\top x_1\exp(2{x_1}^{\top}Ax_q+2y_1b^{\top}x_q)\mathbb{E}_{x_2}\exp({x_2}^{\top}Ax_q+y_2b^{\top}x_q)}{(\exp(x_q^{\top}Ax_q)+D\mathbb{E}\exp({x_1}^{\top}Ax_q+y_1b^{\top}x_q))^3}  \\
 && -\frac{2D(D-1)(D-2) v(\theta^\top x_q)\mathbb{E}
_{x_1} \theta^\top x_1\exp({x_1}^{\top}Ax_q+y_1b^{\top}x_q)\mathbb{E}_{x_2}\exp({x_2}^{\top}Ax_q+y_2b^{\top}x_q)\mathbb{E}_{x_3}\exp({x_3}^{\top}Ax_q+y_3b^{\top}x_q)}{(\exp(x_q^{\top}Ax_q)+D\mathbb{E}\exp({x_1}^{\top}Ax_q+y_1b^{\top}x_q))^3}  \\
&=&\frac{2v}{D}(\theta^\top x_q)\theta^\top(A+\theta b^\top)x_q - \frac{2v}{D}(\theta^\top x_q)\theta^\top(A+\theta b^\top)x_q \exp(x_q^{\top}(A+\theta b^{\top})^{\top}(A+\theta b^{\top})x_q). \\
\end{eqnarray*}
To sum up, we have 
\begin{eqnarray*}
     A_1&=& A_{11}+ A_{121}+ A_{122}+ A_{123}+ A_{13} \\
     &=&-(2v\theta^\top x_q)\theta^{\top}(A+\theta b^{\top})x_q + \frac{(2v\theta^\top x_q)\theta^{\top}(A+\theta b^{\top})x_q\exp(x_q^{\top}Ax_q)}{D\exp(x_q^{\top}(A+\theta b^{\top})^{\top}(A+\theta b^{\top})x_q/2)} \\
    &&+ \frac{2v}{D}(\theta^\top x_q)\theta^\top(A+\theta b^\top)x_q \exp(x_q^{\top}(A+\theta b^{\top})^{\top}(A+\theta b^{\top})x_q). \\
   \end{eqnarray*}  
    In terms of the second-order term, since $x_i$s are independent of each other, we have
     \begin{eqnarray*}
         &&\mathbb{E}_{\{x_i\}_{i\in[D]}}A_2\\
    &=& \mathbb{E}_{\{x_i\}_{i\in[D]}}\left(\frac{ v \sum_{i=1}^D 
    \theta^{\top} x_i\exp(x_i^{\top}Ax_q+y_ib^{\top}x_q) }{D \mathbb{E}_{x_1} \exp(x_1^{\top}Ax_q+y_1b^{\top}x_q)
    +\exp(x_q^{\top}Ax_q)}\right)^2\\
        && -2\mathbb{E}_{\{x_i\}_{i\in[D]}}\frac{\left(v \sum_{i=1}^D 
    \theta^{\top} x_i\exp(x_i^{\top}Ax_q+y_ib^{\top}x_q) \right)^2}{\left(D \mathbb{E}_{x_1} \exp(x_1^{\top}Ax_q+y_1b^{\top}x_q)
    +\exp(x_q^{\top}Ax_q)\right)^3}\left( \sum \exp(x_i^{\top}Ax_q+y_ib^{\top}x_q) 
     - (D \mathbb{E}_{x_1} \exp(x_1^{\top}Ax_q+y_1b^{\top}x_q)
)\right)\\
        &&+ 3\mathbb{E}_{\{x_i\}_{i\in[D]}}\frac{\left(v \sum_{i=1}^D 
    \theta^{\top} x_i\exp(x_i^{\top}Ax_q+y_ib^{\top}x_q) \right)^2}{\left(D \mathbb{E}_{x_1} \exp(x_1^{\top}Ax_q+y_1b^{\top}x_q)
    +\exp(x_q^{\top}Ax_q)\right)^4}\left( \sum \exp(x_i^{\top}Ax_q+y_ib^{\top}x_q) 
     - (D \mathbb{E}_{x_1} \exp(x_1^{\top}Ax_q+y_1b^{\top}x_q)
)\right)^2\\
    &=&\frac{Dv^2\mathbb{E}_{x_1} 
    \theta^{\top} x_1x_1^{\top}\theta\exp(2x_1^{\top}(A+\theta b^{\top})x_q)}{(D \mathbb{E}_{x_1} \exp(x_1^{\top}Ax_q+y_1b^{\top}x_q)
    +\exp(x_q^{\top}Ax_q))^2} +\frac{D(D-1)v^2\mathbb{E}_{x_1,x_2} 
    \theta^{\top} x_1x_2^{\top}\theta\exp(x_1^{\top}(A+\theta b^{\top})x_q+x_2^{\top}(A+\theta b^{\top})x_q) }{(D \mathbb{E}_{x_1} \exp(x_1^{\top}Ax_q+y_1b^{\top}x_q)
    +\exp(x_q^{\top}Ax_q))^2} \\
 && -\mathbb{E}_{\{x_i\}_{i\in[D]}}  \frac{2D(D-1)v^2\theta^{\top} x_1x_2^{\top}\theta\exp(x_1^{\top}(A+\theta b^{\top})x_q+x_2^{\top}(A+\theta b^{\top})x_q) }{(D \mathbb{E}_{x_1} \exp(x_1^{\top}Ax_q+y_1b^{\top}x_q)
    +\exp(x_q^{\top}Ax_q))^3} \\
     && \qquad\qquad\qquad\qquad\qquad\qquad\times\left(\sum \exp(x_i^{\top}Ax_q+y_ib^{\top}x_q) 
     - (D \mathbb{E}_{x_1} \exp(x_1^{\top}Ax_q+y_1b^{\top}x_q)
) \right) \\
&&+  \mathbb{E}_{\{x_i\}_{i\in[D]}}\frac{3D(D-1)v^2
    \theta^{\top} x_1x_2^{\top}\theta\exp(x_1^{\top}(A+\theta b^{\top})x_q+x_2^{\top}(A+\theta b^{\top})x_q) }{(D \mathbb{E}_{x_1} \exp(x_1^{\top}Ax_q+y_1b^{\top}x_q)
    +\exp(x_q^{\top}Ax_q))^4} \\
     && \qquad\qquad\qquad\qquad\qquad\qquad\times\left(\sum \exp(x_i^{\top}Ax_q+y_ib^{\top}x_q) 
     - (D \mathbb{E}_{x_1} \exp(x_1^{\top}Ax_q+y_1b^{\top}x_q)
) \right)^2+o(\frac{1}{D})\\
&=&A_{21}+A_{22}+A_{23}+A_{24}.
    \end{eqnarray*}
For the terms $A_{21}$ to $A_{24}$, we have
     \begin{eqnarray*}
A_{21}
&=& \frac{Dv^2\theta^{\top}(I_d+4(A+\theta b^{\top})x_q x_q^{\top}(A+\theta b^{\top})^{\top})\theta\exp(2x_q^{\top}(A+\theta b^{\top})^{\top}(A+\theta b^{\top})x_q)}{(D \exp(x_q^{\top}(A+\theta b^{\top})^{\top}(A+\theta b^{\top})x_q/2)
    +\exp(x_q^{\top}Ax_q))^2} \\
&=& \frac{Dv^2\theta^{\top}(I_d+4(A+\theta b^{\top})x_q x_q^{\top}(A+\theta b^{\top})^{\top})\theta \exp(2x_q^{\top}(A+\theta b^{\top})^{\top}(A+\theta b^{\top})x_q)}{(D \exp(x_q^{\top}(A+\theta b^{\top})^{\top}(A+\theta b^{\top})x_q/2))^2}+o(\frac{1}{D})  \\
&=& \frac{v^2}{D}\theta^{\top}(I_d+4(A+\theta b^{\top})x_q x_q^{\top}(A+\theta b^{\top})^{\top})\theta \exp(x_q^{\top}(A+\theta b^{\top})^{\top}(A+\theta b^{\top})x_q)+o(\frac{1}{D}),
      \end{eqnarray*}
       \begin{eqnarray*}
       A_{22}&=&\frac{D(D-1)v^2\mathbb{E}_{x_1,x_2} 
    \theta^{\top} x_1x_2^{\top}\theta\exp(x_1^{\top}(A+\theta b^{\top})x_q+x_2^{\top}(A+\theta b^{\top})x_q) }{(D \mathbb{E}_{x_1} \exp(x_1^{\top}Ax_q+y_1b^{\top}x_q))^2} \\
    &&-\frac{2D(D-1)v^2\mathbb{E}_{x_1,x_2} 
    \theta^{\top} x_1x_2^{\top}\theta\exp(x_1^{\top}(A+\theta b^{\top})x_q+x_2^{\top}(A+\theta b^{\top})x_q)    \exp(x_q^{\top}Ax_q) }{(D \mathbb{E}_{x_1} \exp(x_1^{\top}Ax_q+y_1b^{\top}x_q))^3} +o(\frac{1}{D})\\
    &=&v^2(1-\frac{1}{D})(\theta^{\top}(A+\theta b^{\top})x_q)^2 - \frac{2v^2(\theta^{\top}(A+\theta b^{\top})x_q)^2 \exp{(x_q^{\top}Ax_q)}}{D \exp(x_q^{\top}(A+\theta b^{\top})^{\top}(A+\theta b^{\top})x_q/2)}+o(\frac{1}{D}),
       \end{eqnarray*}
       \begin{eqnarray*}
   A_{23}&=&    \frac{4v^2}{D}(\theta^{\top}(A+\theta b^{\top})x_q)^2 -\frac{8v^2}{D} (\theta^{\top}(A+\theta b^{\top})x_q)^2\exp(x_q^{\top}(A+\theta b^{\top})^{\top}(A+\theta b^{\top})x_q), 
       \end{eqnarray*}
       and
 \begin{eqnarray*}
A_{24}&=& \frac{3D^3(D-1)v^2\mathbb{E}_{x_1,x_2} 
    \theta^{\top} x_1x_2^{\top}\theta\exp(x_1^{\top}(A+\theta b^{\top})x_q+x_2^{\top}(A+\theta b^{\top})x_q)  (\mathbb{E}_{x_1} \exp(x_1^{\top}Ax_q+y_1b^{\top}x_q) )^2}{(D \mathbb{E}_{x_1} \exp(x_1^{\top}Ax_q+y_1b^{\top}x_q)
    +\exp(x_q^{\top}Ax_q))^4}\\
&+& \frac{3D(D-1)(D-2)v^2\mathbb{E}_{x_1,x_2} 
    \theta^{\top} x_1x_2^{\top}\theta\exp(x_1^{\top}(A+\theta b^{\top})x_q+x_2^{\top}(A+\theta b^{\top})x_q) \mathbb{E}_{x_3} \exp(2x_3^{\top}Ax_q+2y_3b^{\top}x_q)}{(D \mathbb{E}_{x_3} \exp(x_1^{\top}Ax_q+y_1b^{\top}x_q)
    +\exp(x_q^{\top}Ax_q))^4}\\
&+&\frac{12D(D-1)(D-2)v^2\mathbb{E}_{x_1,x_2} 
    \theta^{\top} x_1x_2^{\top}\theta\exp(2x_1^{\top}(A+\theta b^{\top})x_q+x_2^{\top}(A+\theta b^{\top})x_q) (\mathbb{E}_{x_1} \exp(x_1^{\top}Ax_q+y_1b^{\top}x_q) )}{(D \mathbb{E}_{x_1} \exp(x_1^{\top}Ax_q+y_1b^{\top}x_q)
    +\exp(x_q^{\top}Ax_q))^4}\\
&+& \frac{3D(D-1)(D-2)(D-3)v^2\mathbb{E}_{x_1,x_2} 
    \theta^{\top} x_1x_2^{\top}\theta\exp(x_1^{\top}(A+\theta b^{\top})x_q+x_2^{\top}(A+\theta b^{\top})x_q)}{(D \mathbb{E}_{x_1} \exp(x_1^{\top}Ax_q+y_1b^{\top}x_q)
    +\exp(x_q^{\top}Ax_q))^4}\\
    &&(\mathbb{E}_{x_3,x_4} 
   \exp(x_3^{\top}(A+\theta b^{\top})x_q+x_4^{\top}(A+\theta b^{\top})x_q))\\
&-&\frac{6D^2(D-1)(D-2)v^2\mathbb{E}_{x_1,x_2} 
    \theta^{\top} x_1x_2^{\top}\theta\exp(x_1^{\top}(A+\theta b^{\top})x_q+x_2^{\top}(A+\theta b^{\top})x_q) (\mathbb{E}_{x_1} \exp(x_1^{\top}Ax_q+y_1b^{\top}x_q) )^2}{(D \mathbb{E}_{x_3} \exp(x_1^{\top}Ax_q+y_1b^{\top}x_q)
    +\exp(x_q^{\top}Ax_q))^4}\\
&-&\frac{12D^2(D-1)v^2\mathbb{E}_{x_1,x_2} 
    \theta^{\top} x_1x_2^{\top}\theta\exp(2x_1^{\top}(A+\theta b^{\top})x_q+x_2^{\top}(A+\theta b^{\top})x_q) (\mathbb{E}_{x_1} \exp(x_1^{\top}Ax_q+y_1b^{\top}x_q) )}{(D \mathbb{E}_{x_3} \exp(x_1^{\top}Ax_q+y_1b^{\top}x_q)
    +\exp(x_q^{\top}Ax_q))^4}\\
    & =&-\frac{3v^2}{D}(\theta^{\top}(A+\theta b^{\top})x_q)^2 + \frac{3v^2}{D}(\theta^{\top}(A+\theta b^{\top})x_q)^2 \exp(x_q^{\top}(A+\theta b^{\top})^{\top}(A+\theta b^{\top})x_q)+o\left(\frac{1}{D}\right).
\end{eqnarray*}


 To sum up, 
 \begin{eqnarray*}
\mathbb{E}_{\{x_i,y_i\}_{i\in[D]}}A_2 
 &=& \frac{v^2}{D}\theta^{\top}(I_d-(A+\theta b^{\top})x_q x_q^{\top}(A+\theta b^{\top})^{\top})\theta \exp(x_q^{\top}(A+\theta b^{\top})^{\top}(A+\theta b^{\top})x_q) \\
 &+&v^2(\theta^{\top}(A+\theta b^{\top})x_q)^2 - \frac{2v^2(\theta^{\top}(A+\theta b^{\top})x_q)^2 \exp{(x_q^{\top}Ax_q)}}{D \exp(x_q^{\top}(A+\theta b^{\top})^{\top}(A+\theta b^{\top})x_q/2)} +o(\frac{1}{D}).\\
\end{eqnarray*} 
Based on the results of $A_1$ and $A_2$, we have 
        \begin{eqnarray}
    &&\mathbb{E}\left(y_q- (W^V_{d+1,:})^{\top}
        E\phi\left(E^{\top}(W^K)^{\top} W^Q \begin{bmatrix}
            x_q\\0
        \end{bmatrix} \right)\right)^2\nonumber\\
    &=&\mathbb{E}_{(x_q,\theta)}(x_q^{\top}\theta)^2+v^2(\theta^{\top}(A+\theta b^{\top})x_q)^2-2v(x_q^{\top}\theta)(\theta^{\top}(A+\theta b^{\top})x_q)+O\left(\frac{1}{D}\right)\nonumber\\
    &=&\mathbb{E}_{(x_q,\theta)}\left(\theta^{\top}(v(A+\theta b^{\top})-I_d)x_q\right)^2+O\left(\frac{1}{D}\right)\nonumber\\
    &=&\mathbb{E}_{(x_q,\theta)}\left[\left(\theta^{\top}(vA-I_d)x_q\right)^2+v^2(\|\theta\|^2b^{\top}x_q)^2+2v\theta^{\top}(vA-I_d)x_qx_q^{\top}b\|\theta\|^2\right]+O\left(\frac{1}{D}\right)\nonumber\\
    &=&\frac{1}{d}tr\left((vA-I_d)^2\right)+v^2\|b\|^2\mathbb{E}\|\theta\|^4+O\left(\frac{1}{D}\right).
        \end{eqnarray}
        Therefore, to minimize the loss, the optimal $A$ satisfies $tr\left((vA-I_d)^2\right)=O(d/D)$, and $\|b\|^2=O(1/D)$.

Furthermore, we have
        
    \begin{eqnarray}
&&\mathbb{E}\left(y_q- (W^V_{d+1,:})^{\top}
    E\phi\left(E^{\top}(W^K)^{\top} W^Q \begin{bmatrix}
        x_q\\0
    \end{bmatrix} \right)\right)^2 \nonumber \\
&=&\mathbb{E}_{(x_q,\theta)}\bigg[(x_q^{\top}\theta)^2 + A_{1}+A_{2}\bigg]\nonumber\\
&=&\mathbb{E}_{(x_q,\theta)}  \bigg[(x_q^{\top}\theta)^2-(2v\theta^\top x_q)\theta^{\top}(A+\theta b^{\top})x_q+ \frac{2v}{D}(\theta^\top x_q)\theta^\top(A+\theta b^\top)x_q \exp(x_q^{\top}(A+\theta b^{\top})^{\top}(A+\theta b^{\top})x_q) \nonumber\\
    &&+\frac{v^2}{D}\theta^{\top}(I_d-(A+\theta b^{\top})x_q x_q^{\top}(A+\theta b^{\top})^{\top})\theta \exp(x_q^{\top}(A+\theta b^{\top})^{\top}(A+\theta b^{\top})x_q)+v^2(\theta^{\top}(A+\theta b^{\top})x_q)^2\nonumber\\
    &&+ \frac{(2v\theta^\top x_q)\theta^{\top}(A+\theta b^{\top})x_q\exp(x_q^{\top}Ax_q)}{D\exp(x_q^{\top}(A+\theta b^{\top})^{\top}(A+\theta b^{\top})x_q/2)}- \frac{2v^2(\theta^{\top}(A+\theta b^{\top})x_q)^2 \exp{(x_q^{\top}Ax_q)}}{D \exp(x_q^{\top}(A+\theta b^{\top})^{\top}(A+\theta b^{\top})x_q/2)} +o(\frac{1}{D})\bigg]\nonumber\\
  &=& \frac{1}{d}tr\left((Av-I_d)^2\right)+v^2\|b\|^2\mathbb{E}\|\theta\|^4 + \mathbb{E}_{\theta} \bigg(\frac{v^2}{D} det (\Sigma_1)^{\frac{1}{2}}\|\theta\|^2 + \frac{2v}{D} det (\Sigma_1)^{\frac{1}{2}} \theta^{\top}(A+\theta b^{\top})\Sigma_1 \theta+ \frac{2v}{D} det (\Sigma_2)^{\frac{1}{2}} \theta^{\top}(A+\theta b^{\top})\Sigma_2 \theta\nonumber\\
  &&- \frac{v^2}{D} det (\Sigma_1)^{\frac{1}{2}} \theta^{\top}(A+\theta b^{\top})\Sigma_1 (A+\theta b^{\top})^{\top}\theta - \frac{2v^2}{D} det (\Sigma_2)^{\frac{1}{2}} \theta^{\top}(A+\theta b^{\top})\Sigma_2 (A+\theta b^{\top})^{\top}\theta\bigg) + o(\frac{1}{D}),
    \end{eqnarray}
    
where $\Sigma_1=(I-2(A+\theta b^{\top})^{\top}(A+\theta b^{\top}))^{-1}$ and  $\Sigma_2=(I+(A+\theta b^{\top})^{\top}(A+\theta b^{\top})-2A)^{-1}$.

\end{proof}

Assuming that $A^*=\frac{I_d}{v}+\Delta_A$, $b^*=\Delta_b$, where $\Delta_A=O(\frac{1}{\sqrt{D}})$ and $\Delta_b=O(\frac{1}{\sqrt{D}})$, we will have

\begin{eqnarray*}
&&{\frac{v^2}{D} det (\Sigma_1)^{\frac{1}{2}}\|\theta\|^2}\big|_{A=\frac{I_d}{v}+\Delta_A, b=\Delta_b} \\
&=& \frac{v^2}{D} det\left(\left(I-2(I_d/v+\Delta_A+\theta \Delta_b^{\top})^{\top}(I_d/v+\Delta_A+\theta \Delta_b^{\top})\right)^{-1}\right)^{\frac{1}{2}} \|\theta\|^2 \\ 
 &=& \frac{v^2}{D} det\left(\left((1-\frac{2}{v^2})I-\frac{2}{v}(\Delta_A+\theta \Delta_b^{\top})^{\top} -\frac{2}{v}(\Delta_A+\theta \Delta_b^{\top})-2(\Delta_A+\theta \Delta_b^{\top})^{\top}(\Delta_A+\theta \Delta_b^{\top})\right)^{-1}\right)^{\frac{1}{2}} \|\theta\|^2 
 \\
  &=& \frac{v^2}{D}(\frac{v^2-2}{v^2})^{\frac{d}{2}} det\left(\left((1-\frac{2}{v^2})I-\frac{2}{v}(\Delta_A+\theta \Delta_b^{\top})^{\top} -\frac{2}{v}(\Delta_A+\theta \Delta_b^{\top})-2(\Delta_A+\theta \Delta_b^{\top})^{\top}(\Delta_A+\theta \Delta_b^{\top})\right)^{-1}\right) \|\theta\|^2. \\
  \end{eqnarray*}
Therefore,
\begin{eqnarray*}
&&\mathbb E_{\theta} \left({\frac{v^2}{D} det (\Sigma_1)^{\frac{1}{2}}\|\theta\|^2}\big|_{A=\frac{I_d}{v}+\Delta_A, b=\Delta_b}-{\frac{v^2}{D} det (\Sigma_1)^{\frac{1}{2}}\|\theta\|^2}\big|_{A=\frac{I_d}{v}, b=0}\right)\\
&=& \mathbb E_{\theta} \frac{v^2}{D}(\frac{v^2-2}{v^2})^\frac{d}{2} \left(-(\frac{v^2}{v^2-2})^{d+1}tr(-\frac{2}{v}(\Delta_A+\theta \Delta_b^{\top})^{\top} -\frac{2}{v}(\Delta_A+\theta \Delta_b^{\top})-2(\Delta_A+\theta \Delta_b^{\top})^{\top}(\Delta_A+\theta \Delta_b^{\top}))\right) \|\theta\|^2  \\
&=&\mathbb E_{\theta} \frac{v^2}{D} (\frac{v^2}{v^2-2})^{\frac{d}{2}+1}(\frac{4}{v}tr(\Delta_A) + 2{\|\ \Delta_A\|\ }_F^2+ 2\|\Delta_b\|^2\|\theta\|^2)\|\theta\|^2  = o(\frac{1}{D}).
 \end{eqnarray*}

Furthermore, we have:
  \begin{eqnarray*}
  &&\Sigma_2\big|_{A=\frac{I_d}{v}+\Delta_A, b=\Delta_b}=\left(I+(I_d/v+\Delta_A+\theta \Delta_b^{\top})^{\top}(I_d/v+\Delta_A+\theta\Delta_b^{\top})-2(I_d/v+\Delta_A)\right)^{-1} \\
  &=&\left( \frac{(v-1)^2}{v^2}(I +\frac{v}{(v-1)^2}(\Delta_A+\theta\Delta_b^{\top}) +\frac{v}{(v-1)^2}(\Delta_A+\theta\Delta_b^{\top})^{\top}-\frac{v^2}{(v-1)^2}2\Delta_A + \frac{v^2}{(v-1)^2} (\Delta_A+\theta \Delta_b^{\top})^{\top}(\Delta_A+\theta \Delta_b^{\top}))\right)^{-1} \\
   &=& \frac{v^2}{(v-1)^2}\left(I -\frac{v}{(v-1)^2}(\Delta_A+\theta\Delta_b^{\top}) -\frac{v}{(v-1)^2}(\Delta_A+\theta\Delta_b^{\top})^{\top}+\frac{v^2}{(v-1)^2}2\Delta_A - \frac{v^2}{(v-1)^2} (\Delta_A+\theta \Delta_b^{\top})^{\top}(\Delta_A+\theta \Delta_b^{\top}))\right)\\
   &=&\Sigma_2^*.
   \end{eqnarray*} Therefore,
   \begin{eqnarray*}
   &&- \frac{2v^2}{D} \left(\det(\Sigma_2)^{\frac{1}{2}} \theta^{\top}(A+\theta b^{\top})\Sigma_2 (A+\theta b^{\top})^{\top}\theta\big|_{A=\frac{I_d}{v}+\Delta_A, b=\Delta_b} -\det(\Sigma_2)^{\frac{1}{2}} \theta^{\top}(A+\theta b^{\top})\Sigma_2 (A+\theta b^{\top})^{\top}\theta\big|_{A=\frac{I_d}{v}, b=0} \right)\\
    &=& - \frac{2v^2}{D} \left(\det(\Sigma_2^*)^{\frac{1}{2}} \theta^{\top}(I/v+\Delta_A+\theta \Delta_b^{\top})\Sigma_2^* (I/v+\Delta_A+\theta \Delta_b^{\top})^{\top}\theta -\det(\Sigma_2^*)^{\frac{1}{2}} \frac{1}{(v-1)^2}\theta^\top \theta\right)\\
    && - \frac{2v^2}{D} \left(\det(\Sigma_2^*)^{\frac{1}{2}} \frac{1}{(v-1)^2}\theta^\top \theta -\left[\det(\frac{v^2}{(v-1)^2}I)^{\frac{1}{2}}\right] \frac{1}{(v-1)^2}\theta^\top \theta\right) \\
    &=& o(\frac{1}{D}).
   \end{eqnarray*}
   
We can obtain similar results for other terms. Therefore, we have $L(A^* , b^*)-L(I_d/v , 0)= o(\frac{1}{D})$.
  
  When $A=\frac{I_d}{v}$ and $b=0$,
    \begin{eqnarray}
&&\mathbb{E}\left(y_q- (W^V_{d+1,:})^{\top}
    E\phi\left(E^{\top}(W^K)^{\top} W^Q \begin{bmatrix}
        x_q\\0
    \end{bmatrix} \right)\right)^2\nonumber\\
    &=&\mathbb{E}_{(x_q,\theta)}  \bigg[(\frac{1}{D}(\theta^\top x_q)\theta^\top x_q \exp(x_q^{\top}x_q/v^2)+\frac{v^2}{D}\theta^{\top}\theta \exp(x_q^{\top}x_q/v^2)\bigg]\nonumber\\
    &=& \frac{v^2}{D}(\frac{v^2}{v^2-2})^\frac{d}{2} + \frac{v^2}{D(v^2-2)}(\frac{v^2}{v^2-2})^\frac{d}{2}  + o(\frac{1}{D}),
    \end{eqnarray}
and $v$ should satisfies  $v^2>2$.

\subsection{Theorem \ref{thm:multi_head}}\label{sec:proof:multi_head}
\begin{proof}[Proof of Theorem \ref{thm:multi_head}]
 \begin{eqnarray*}
 &&\mathbb{E}\left(y_q- f(E)_{d+1,D+1} 
 \right)^2\\
&=&   \mathbb{E}\left(y_q-vm E_{d+1,:}\phi((W_1^KE)^{\top}W_1^QE_{:,D+1})+vn E_{d+1,:}\phi((W_2^KE)^{\top}W_2^QE_{:,D+1}) \right)^2 \\
&=&\mathbb{E}\left(y_q- vm \begin{bmatrix}
        y_1,y_2,\ldots,y_D,0
    \end{bmatrix}\phi\left(E^{\top}(W_1^K)^{\top} W_1^Q \begin{bmatrix}
        x_q\\0
    \end{bmatrix} \right) + vn \begin{bmatrix}
        y_1,y_2,\ldots,y_D,0
    \end{bmatrix}\phi\left(E^{\top}(W_2^K)^{\top} W_2^Q \begin{bmatrix}
        x_q\\0
    \end{bmatrix} \right)\right)^2\\
    &=& \mathbb{E}\left(y_q-\frac{vm \sum_{i=1}^D 
    \theta^{\top} x_i\exp(x_i^{\top}(A_1+\theta b_1^{\top})x_q) }{\sum \exp(x_i^{\top}A_1x_q+y_ib_1^{\top}x_q) 
    +\exp(x_q^{\top}A_1x_q)}+\frac{vn \sum_{i=1}^D 
    \theta^{\top} x_i\exp(x_i^{\top}(A_2+\theta b_2^{\top})x_q) }{\sum \exp(x_i^{\top}A_2x_q+y_ib_2^{\top}x_q) 
    +\exp(x_q^{\top}A_2x_q)}\right)^2\\
    &=& \mathbb{E}_{(x_q,\theta)}\mathbb{E}_{\{x_i\}_{i\in[D]}}\left(y_q^2+\underbrace{ \left(\frac{vm\sum_{i=1}^D 
    \theta^{\top} x_i\exp(x_i^{\top}A_1x_q+y_ib_1^{\top}x_q) }{\sum \exp(x_i^{\top}A_1x_q+y_ib_1^{\top}x_q) +\exp(x_q^{\top}A_1x_q)}\right)^2}_{B_1} +\underbrace{\left(\frac{vn\sum_{i=1}^D
    \theta^{\top} x_i\exp(x_i^{\top}A_2x_q+y_ib_2^{\top}x_q) }{\sum \exp(x_i^{\top}A_2x_q+y_ib_2^{\top}x_q) 
    +\exp(x_q^{\top}A_2x_q)}\right)^2 }_{B_2}
    \right) \\
     &+& \mathbb{E}_{(x_q,\theta)}\mathbb{E}_{\{x_i\}_{i\in[D]}}\left(\underbrace{-2y_q\left(\frac{vm\sum_{i=1}^D 
    \theta^{\top} x_i\exp(x_i^{\top}A_1x_q+y_ib_1^{\top}x_q) }{\sum \exp(x_i^{\top}A_1x_q+y_ib_1^{\top}x_q) 
    +\exp(x_q^{\top}A_1x_q)}\right)
    }_{B_2}
    +\underbrace{2y_q\left(\frac{vn\sum_{i=1}^D 
    \theta^{\top} x_i\exp(x_i^{\top}A_2x_q+y_ib_2^{\top}x_q) }{\sum \exp(x_i^{\top}A_2x_q+y_ib_2^{\top}x_q) 
    +\exp(x_q^{\top}A_2x_q)}\right)
    }_{B_4}\right). \\  
 &+&\mathbb{E}_{(x_q,\theta)}\mathbb{E}_{\{x_i\}_{i\in[D]}}\left(\underbrace{-\frac{2vm\sum_{i=1}^D 
    \theta^{\top} x_i\exp(x_i^{\top}A_1x_q+y_ib_1^{\top}x_q) }{\sum \exp(x_i^{\top}A_1x_q+y_ib_1^{\top}x_q) 
    +\exp(x_q^{\top}A_1x_q)} \frac{vn\sum_{i=1}^D 
    \theta^{\top} x_i\exp(x_i^{\top}A_2x_q+y_ib_2^{\top}x_q) }{\sum \exp(x_i^{\top}A_2x_q+y_ib_2^{\top}x_q) 
    +\exp(x_q^{\top}A_2x_q)}
   }_{B_5} \right). \\  
        \end{eqnarray*}
Similar to the $\mathbb{E}_{\{x_i\}_{i\in[D]}}A_2$  of \ref{sec:proof:optimal}, we have 
\begin{eqnarray*}
\mathbb{E}_{\{x_i\}_{i\in[D]}}B_1&=& \frac{v^2m^2}{D}\theta^{\top}(I_d-(A_1+\theta b_1^{\top})x_q x_q^{\top}(A_1+\theta b_1^{\top})^{\top})\theta \exp(x_q^{\top}(A_1+\theta b_1^{\top})^{\top}(A_1+\theta b_1^{\top})x_q) \\
 &&+v^2m^2(\theta^{\top}(A_1+\theta b_1^{\top})x_q)^2 - \frac{2v^2m^2(\theta^{\top}(A_1+\theta b_1^{\top})x_q)^2 \exp{(x_q^{\top}A_1x_q)}}{D \exp(x_q^{\top}(A_1+\theta b_1^{\top})^{\top}(A_1+\theta b_1^{\top})x_q/2)} +o(\frac{1}{D}),  \\
\mathbb{E}_{\{x_i\}_{i\in[D]}}B_2&=& \frac{v^2n^2}{D}\theta^{\top}(I_d-(A_2+\theta b_2^{\top})x_q x_q^{\top}(A_2+\theta b_2^{\top})^{\top})\theta \exp(x_q^{\top}(A_2+\theta b_2^{\top})^{\top}(A_2+\theta b_2^{\top})x_q) \\
 &&+v^2n^2(\theta^{\top}(A_2+\theta b_2^{\top})x_q)^2 - \frac{2v^2n^2(\theta^{\top}(A_2+\theta b_2^{\top})x_q)^2 \exp{(x_q^{\top}A_2x_q)}}{D \exp(x_q^{\top}(A_2+\theta b_2^{\top})^{\top}(A_2+\theta b_2^{\top})x_q/2)}  +o(\frac{1}{D}),\\
 \mathbb{E}_{\{x_i\}_{i\in[D]}}B_3&=& -(2mv\theta^\top x_q)\theta^{\top}(A_1+\theta b_1^{\top})x_q + \frac{(2vm\theta^\top x_q)\theta^{\top}(A_1+\theta b_1^{\top})x_q\exp(x_q^{\top}A_1x_q)}{D\exp(x_q^{\top}(A_1+\theta b_1^{\top})^{\top}(A_1+\theta b_1^{\top})x_q/2)} \\
    &&+ \frac{2mv}{D}(\theta^\top x_q)\theta^\top(A_1+\theta b_1^\top)x_q \exp(x_q^{\top}(A_1+\theta b_1^{\top})^{\top}(A_1+\theta b_1^{\top})x_q) +o(\frac{1}{D}), \\
  \mathbb{E}_{\{x_i\}_{i\in[D]}}B_4&=& +(2nv\theta^\top x_q)\theta^{\top}(A_2+\theta b_2^{\top})x_q- \frac{(2vn\theta^\top x_q)\theta^{\top}(A_2+\theta b_2^{\top})x_q\exp(x_q^{\top}A_2x_q)}{D\exp(x_q^{\top}(A_2+\theta b_2^{\top})^{\top}(A_2+\theta b_2^{\top})x_q/2)} \\
    &&- \frac{2nv}{D}(\theta^\top x_q)\theta^\top(A_2+\theta b+2^\top)x_q \exp(x_q^{\top}(A_2+\theta b_2^{\top})^{\top}(A_2+\theta b_2^{\top})x_q) +o(\frac{1}{D}),\\   
    \end{eqnarray*}
    and
\begin{eqnarray*}
&&\mathbb{E}_{\{x_i\}_{i\in[D]}}B_5 \\
&=&-\frac{2v^2mnD(D-1)(\mathbb{E}_{x_1}\theta^{\top} x_1 \exp(x_1^{\top}A_1x_q+y_1b_1^{\top}x_q) (\mathbb{E}_{x_2}\theta^{\top} x_2 \exp(x_2^{\top}A_2x_q+y_2b_2^{\top}x_q)
    )}{(D \mathbb{E}_{x_1} \exp(x_1^{\top}A_1x_q+y_1b_1^{\top}x_q)
    +\exp(x_q^{\top}A_1x_q))(D \mathbb{E}_{x_2} \exp(x_2^{\top}A_2x_q+y_1b_2^{\top}x_q)
    +\exp(x_q^{\top}A_2x_q))} \\
&-& \frac{2v^2mnD(\mathbb{E}_{x_1}(\theta^{\top} x_1)^2 \exp(x_1^{\top}A_1x_q+y_1b_1^{\top}x_q) \exp(x_1^{\top}A_2x_q+y_1b_2^{\top}x_q)
    )}{(D \mathbb{E}_{x_1} \exp(x_1^{\top}A_1x_q+y_1b_1^{\top}x_q)
    +\exp(x_q^{\top}A_1x_q))(D \mathbb{E}_{x_2} \exp(x_2^{\top}A_2x_q+y_2b_2^{\top}x_q)
    +\exp(x_q^{\top}A_2x_q))} \\
&+& \mathbb{E}_{\{x_i\}_{i\in[D]}} \left(\frac{2v^2mnD(D-1)(\theta^{\top} x_1 \exp(x_1^{\top}A_1x_q+y_1b_1^{\top}x_q) (\theta^{\top} x_2 \exp(x_2^{\top}A_2x_q+y_2b_2^{\top}x_q)
    )}{(D \mathbb{E}_{x_1} \exp(x_1^{\top}A_1x_q+y_1b_1^{\top}x_q)
    +\exp(x_q^{\top}A_1x_q))(D \mathbb{E}_{x_2} \exp(x_2^{\top}A_2x_q+y_2b_2^{\top}x_q)
    +\exp(x_q^{\top}A_2x_q))^2}  \right)\\
&& \times\left( \sum_{i=1}^D \exp(x_i^{\top}A_2x_q+y_ib_2^{\top}x_q) - D \mathbb{E}_{x_2} \exp(x_2^{\top}A_2x_q+y_2b_2^{\top}x_q) \right) \\
&+& \mathbb{E}_{\{x_i\}_{i\in[D]}} \left(\frac{2v^2mnD(D-1)(\theta^{\top} x_1 \exp(x_1^{\top}A_1x_q+y_1b_1^{\top}x_q) (\theta^{\top} x_2 \exp(x_2^{\top}A_2x_q+y_2b_2^{\top}x_q)
    )}{(D \mathbb{E}_{x_1} \exp(x_1^{\top}A_1x_q+y_1b_1^{\top}x_q)
    +\exp(x_q^{\top}A_1x_q))^2(D \mathbb{E}_{x_2} \exp(x_2^{\top}A_2x_q+y_2b_2^{\top}x_q)
    +\exp(x_q^{\top}A_2x_q))}  \right)\\
&&\times \left( \sum_{i=1}^D \exp(x_i^{\top}A_1x_q+y_ib_1^{\top}x_q) - D \mathbb{E}_{x_1} \exp(x_1^{\top}A_1x_q+y_1b_1^{\top}x_q) \right) \\
&-&\mathbb{E}_{\{x_i\}_{i\in[D]}} \frac{2v^2mnD(D-1)(\theta^{\top} x_1 \exp(x_1^{\top}A_1x_q+y_1b_1^{\top}x_q) (\theta^{\top} x_2 \exp(x_2^{\top}A_2x_q+y_2b_2^{\top}x_q)
    )}{(D \mathbb{E}_{x_1} \exp(x_1^{\top}A_1x_q+y_1b_1^{\top}x_q)
    +\exp(x_q^{\top}A_1x_q))^2(D \mathbb{E}_{x_2} \exp(x_2^{\top}A_2x_q+y_1b_2^{\top}x_q)
    +\exp(x_q^{\top}A_2x_q))^2} \\
 &&\times\bigg(\sum_{i=1}^D \exp(x_i^{\top}A_1x_q+y_ib_1^{\top}x_q) - D \mathbb{E}_{x_1} \exp(x_1^{\top}A_1x_q+y_1b_1^{\top}x_q) \bigg)  \\
 &&\times\bigg( \sum_{i=1}^D \exp(x_i^{\top}A_2x_q+y_ib_2^{\top}x_q) - D \mathbb{E}_{x_2} \exp(x_2^{\top}A_2x_q+y_2b_2^{\top}x_q) \bigg)\\
 &-&\mathbb{E}_{\{x_i\}_{i\in[D]}} \frac{2v^2mnD(D-1)(\theta^{\top} x_1 \exp(x_1^{\top}A_1x_q+y_1b_1^{\top}x_q) (\theta^{\top} x_2 \exp(x_2^{\top}A_2x_q+y_2b_2^{\top}x_q)
    )}{(D \mathbb{E}_{x_1} \exp(x_1^{\top}A_1x_q+y_1b_1^{\top}x_q)
    +\exp(x_q^{\top}A_1x_q))(D \mathbb{E}_{x_2} \exp(x_2^{\top}A_2x_q+y_1b_2^{\top}x_q)
    +\exp(x_q^{\top}A_2x_q))^3} \\
 &&\times\left( \sum_{i=1}^D \exp(x_i^{\top}A_2x_q+y_ib_2^{\top}x_q) - D \mathbb{E}_{x_2} \exp(x_2^{\top}A_2x_q+y_2b_2^{\top}x_q) \right)^2 \\
  &-&\mathbb{E}_{\{x_i\}_{i\in[D]}} \frac{2v^2mnD(D-1)(\theta^{\top} x_1 \exp(x_1^{\top}A_1x_q+y_1b_1^{\top}x_q) (\theta^{\top} x_2 \exp(x_2^{\top}A_2x_q+y_2b_2^{\top}x_q)
    )}{(D \mathbb{E}_{x_1} \exp(x_1^{\top}A_1x_q+y_1b_1^{\top}x_q)
    +\exp(x_q^{\top}A_1x_q))^3(D \mathbb{E}_{x_2} \exp(x_2^{\top}A_2x_q+y_1b_2^{\top}x_q)
    +\exp(x_q^{\top}A_2x_q))} \\
 &&\times\left( \sum_{i=1}^D \exp(x_i^{\top}A_1x_q+y_ib_1^{\top}x_q) - D \mathbb{E}_{x_1} \exp(x_1^{\top}A_1x_q+y_1b_1^{\top}x_q) \right)^2+ o(\frac{1}{D})\\
&=&B_{51}+B_{52}+B_{53}+B_{54}+B_{55}+B_{56}+B_{57}.
\end{eqnarray*}
For the terms $B_{51}$ to $B_{57}$, we have
\begin{eqnarray*}
 B_{51}  &=&  -2v^2mn (1-\frac{1}{D}) \theta^{\top}(A_1+\theta b_1)x_q \theta^{\top}(A_2+\theta b_2)x_q +2v^2mn\frac{1}{D}\frac{\exp(x_q^{\top}A_1x_q)\theta^{\top}(A_1+\theta b_1)x_q \theta^{\top}(A_2+\theta b_2)x_q}{\exp(x_q^{\top}(A_1+\theta b_1^{\top})^{\top}(A_1+\theta b_1^{\top})x_q/2)} \\
&&+  
2v^2mn\frac{1}{D} \frac{\exp(x_q^{\top}A_2x_q)\theta^{\top}(A_1+\theta b_1)x_q \theta^{\top}(A_2+\theta b_2)x_q}{\exp(x_q^{\top}(A_2+\theta b_2^{\top})^{\top}(A_2+\theta b_2^{\top})x_q/2)}, \\
\end{eqnarray*}

\begin{eqnarray*}
 B_{52} &=&  -\frac{2v^2mn}{D} \theta^{\top}(I+(A_1+\theta b_1^{\top}+A_2+\theta b_2^{\top})x_q^\top x_q(A_1+\theta b_1^{\top}+A_2+\theta b_2^{\top})^{\top})\theta\\
&& \qquad\qquad\qquad\exp(x_q^\top(A_1+\theta b_1^{\top})^{\top}(A_2+\theta b_2^{\top})x_q/2+x_q^\top(A_2+\theta b_2^{\top})^{\top}(A_1+\theta b_1^{\top})x_q/2),
\end{eqnarray*}

\begin{eqnarray*}
B_{53} 
&=&\frac{2v^2mnD(D-1)(\mathbb{E}_{x_1}\theta^{\top} x_1 \exp(x_1^{\top}A_1x_q+y_1b_1^{\top}x_q) (\mathbb{E}_{x_2}\theta^{\top} x_2 \exp(2x_2^{\top}A_2x_q+2y_2b_2^{\top}x_q)
    )}{(D \mathbb{E}_{x_1} \exp(x_1^{\top}A_1x_q+y_1b_1^{\top}x_q)
    +\exp(x_q^{\top}A_1x_q))(D \mathbb{E}_{x_2} \exp(x_2^{\top}A_2x_q+y_2b_2^{\top}x_q)
    +\exp(x_q^{\top}A_2x_q))^2}  \\
&&+ \frac{2v^2mnD(D-1)(\mathbb{E}_{x_1}\theta^{\top} x_1 \exp(x_1^{\top}(A_1+A_2) x_q+y_1(b_1+b_2)^{\top}x_q) (\mathbb{E}_{x_2}\theta^{\top} x_2 \exp(x_2^{\top}A_2x_q+y_2b_2^{\top}x_q)
    )}{(D \mathbb{E}_{x_1} \exp(x_1^{\top}A_1x_q+y_1b_1^{\top}x_q)
    +\exp(x_q^{\top}A_1x_q))(D \mathbb{E}_{x_2} \exp(x_2^{\top}A_2x_q+y_2b_2^{\top}x_q)
    +\exp(x_q^{\top}A_2x_q))^2}  \\
    &&- \frac{4v^2mnD(D-1)(\mathbb{E}_{x_1}\theta^{\top} x_1 \exp(x_1^{\top}A_1x_q+y_1b_1^{\top}x_q) (\mathbb{E}_{x_2}\theta^{\top} x_2 \exp(x_2^{\top}A_2x_q+y_2b_2^{\top}x_q)\mathbb{E}_{x_2} \exp(x_2^{\top}A_2x_q+y_2b_2^{\top}x_q)
    )}{(D \mathbb{E}_{x_1} \exp(x_1^{\top}A_1x_q+y_1b_1^{\top}x_q)
    +\exp(x_q^{\top}A_1x_q))(D \mathbb{E}_{x_2} \exp(x_2^{\top}A_2x_q+y_2b_2^{\top}x_q)
    +\exp(x_q^{\top}A_2x_q))^2}   \\
&=& \frac{2v^2mn}{D}\theta^{\top}(A_1+\theta b_1)x_q \theta^{\top}(A_2+\theta b_2)x_q \left(2\exp(x_q^{\top}(A_2+\theta b_2^{\top})^{\top}(A_2+\theta b_2^{\top})x_q)-2\right)\\
&& +\frac{2v^2mn}{D}\theta^{\top}(A_1+\theta b_1+A_2+\theta b_2)x_q \theta^{\top}(A_2+\theta b_2)x_q  \\
&&\qquad\qquad\times\exp(x_q^\top(A_1+\theta b_1^{\top})^{\top}(A_2+\theta b_2^{\top})x_q/2+x_q^\top(A_2+\theta b_2^{\top})^{\top}(A_1+\theta b_1^{\top})x_q/2), \\
\end{eqnarray*}
\begin{eqnarray*}
B_{54}
&=& \frac{2v^2mn}{D}\theta^{\top}(A_1+\theta b_1)x_q \theta^{\top}(A_2+\theta b_2)x_q \left(2\exp(x_q^{\top}(A_1+\theta b_1^{\top})^{\top}(A_1+\theta b_1^{\top})x_q)-2\right) \\
&&+\frac{2v^2mn}{D}\theta^{\top}(A_1+\theta b_1)x_q \theta^{\top}(A_1+\theta b_1+A_2+\theta b_2)x_q  \\
&&\qquad\qquad\times\
\exp(x_q^\top(A_1+\theta b_1^{\top})^{\top}(A_2+\theta b_2^{\top})x_q/2+x_q^\top(A_2+\theta b_2^{\top})^{\top}(A_1+\theta b_1^{\top})x_q/2),
\end{eqnarray*}
\begin{eqnarray*}
    B_{55}&=&\frac{-2v^2mnD^3(D-1)(\mathbb{E}_{x_1}\theta^{\top} x_1 \exp(x_1^{\top}A_1x_q+y_1b_1^{\top}x_q) (\mathbb{E}_{x_2}\theta^{\top} x_2 \exp(x_2^{\top}A_2x_q+y_2b_2^{\top}x_q)
    )(\mathbb{E}_{x_2} \exp(x_2^{\top}A_2x_q+y_2b_2^{\top}x_q))^2}{(D \mathbb{E}_{x_1} \exp(x_1^{\top}A_1x_q+y_1b_1^{\top}x_q)
    +\exp(x_q^{\top}A_1x_q))(D \mathbb{E}_{x_2} \exp(x_2^{\top}A_2x_q+y_1b_2^{\top}x_q)
    +\exp(x_q^{\top}A_2x_q))^3} \\
   &&+\frac{4v^2mnD^2(D-1)(D-2)(\mathbb{E}_{x_1}\theta^{\top} x_1 \exp(x_1^{\top}A_1x_q+y_1b_1^{\top}x_q) (\mathbb{E}_{x_2}\theta^{\top} x_2 \exp(x_2^{\top}A_2x_q+y_2b_2^{\top}x_q)
    )}{(D \mathbb{E}_{x_1} \exp(x_1^{\top}A_1x_q+y_1b_1^{\top}x_q)
    +\exp(x_q^{\top}A_1x_q))(D \mathbb{E}_{x_2} \exp(x_2^{\top}A_2x_q+y_1b_2^{\top}x_q)
    +\exp(x_q^{\top}A_2x_q))^3} \\
    &&\times\mathbb{E}_{x_2,x_3}\exp(x_2^{\top}A_2x_q+y_2b_2^{\top}x_q+x_3^{\top}A_2x_q+y_3b_2^{\top}x_q)\\
&&-\frac{2v^2mnD(D-1)(D-2)(\mathbb{E}_{x_1}\theta^{\top} x_1 \exp(x_1^{\top}A_1x_q+y_1b_1^{\top}x_q) (\mathbb{E}_{x_2}\theta^{\top} x_2 \exp(x_2^{\top}A_2x_q+y_2b_2^{\top}x_q)
    )}{(D \mathbb{E}_{x_1} \exp(x_1^{\top}A_1x_q+y_1b_1^{\top}x_q)
    +\exp(x_q^{\top}A_1x_q))(D \mathbb{E}_{x_2} \exp(x_2^{\top}A_2x_q+y_1b_2^{\top}x_q)
    +\exp(x_q^{\top}A_2x_q))^3} \\
    &&\mathbb{E}_{x_3} \exp(2x_3^{\top}A_2x_q+2y_3b_2^{\top}x_q)\\
&&-\frac{2v^2mnD(D-1)(D-2)(D-3)(\mathbb{E}_{x_1}\theta^{\top} x_1 \exp(x_1^{\top}A_1x_q+y_1b_1^{\top}x_q) (\mathbb{E}_{x_2}\theta^{\top} x_2 \exp(x_2^{\top}A_2x_q+y_2b_2^{\top}x_q)
    )}{(D \mathbb{E}_{x_1} \exp(x_1^{\top}A_1x_q+y_1b_1^{\top}x_q)
    +\exp(x_q^{\top}A_1x_q))(D \mathbb{E}_{x_2} \exp(x_2^{\top}A_2x_q+y_1b_2^{\top}x_q)
    +\exp(x_q^{\top}A_2x_q))^3} \\
    &&\mathbb{E}_{x_3} \exp(x_3^{\top}A_2x_q+y_3b_2^{\top}x_q+x_4^{\top}A_2x_q+y_4b_2^{\top}x_q)\\
    &=&2v^2mn\frac{1}{D}\theta^{\top}(A_1+\theta b_1)x_q \theta^{\top}(A_2+\theta b_2)x_q\\
    &&-2v^2mn\frac{1}{D}\theta^{\top}(A_1+\theta b_1)x_q \theta^{\top}(A_2+\theta b_2)x_q\exp(x_q^{\top}(A_2+\theta b_2^{\top})^{\top}(A_2+\theta b_2^{\top})x_q),
\end{eqnarray*}
\begin{eqnarray*}
    B_{56}&=& 2v^2mn\frac{1}{D}\theta^{\top}(A_1+\theta b_1)x_q \theta^{\top}(A_2+\theta b_2)x_q\\
    &&-2v^2mn\frac{1}{D}\theta^{\top}(A_1+\theta b_1)x_q \theta^{\top}(A_2+\theta b_2)x_q\exp(x_q^{\top}(A_1+\theta b_1^{\top})^{\top}(A_1+\theta b_1^{\top})x_q),\\
    B_{57}&=& 2v^2mn\frac{1}{D}\theta^{\top}(A_1+\theta b_1)x_q \theta^{\top}(A_2+\theta b_2)x_q\\
    &&-2v^2mn\frac{1}{D}\theta^{\top}(A_1+\theta b_1)x_q \theta^{\top}(A_2+\theta b_2)x_q \exp\left(x_q^\top(A_1+\theta b_1^{\top})^{\top}(A_2+\theta b_2^{\top})x_q/2+x_q^\top(A_2+\theta b_2^{\top})^{\top}(A_1+\theta b_1^{\top})x_q/2\right).
\end{eqnarray*}

Based on the results of $B_1$ to $B_5$, we have 
\begin{eqnarray*}
&&\mathbb{E}\left(y_q- f(E)_{d+1,D+1} 
 \right)^2\\
&=& \mathbb{E}_{(x_q,\theta)}\mathbb{E}_{\{x_i\}_{i\in[D]}} (x_q^{\top}\theta)^2 + v^2 m^2(\theta^{\top}(A_1+\theta b_1^{\top})x_q)^2+v^2n^2(\theta^{\top}(A_2+\theta b_2^{\top})x_q)^2 - 2vm(x_q^{\top}\theta)(\theta^{\top}(A_1+\theta b_1^{\top})x_q) \\
&& +2vn(x_q^{\top}\theta)(\theta^{\top}(A_2+\theta b_2^{\top})x_q)-2v^2mn\theta^{\top}(A_1+\theta b_1^{\top})x_q \theta^{\top}(A_2+\theta b_2^{\top})x_q + O(\frac{1}{D})\\
&=& 1 + \frac{v^2m^2}{d}tr(A_1 A_1^{\top}) + v^2m^2\|b_1\|^2\mathbb{E}\|\theta\|^4 + \frac{v^2n^2}{d} tr(A_2 A_2^{\top}) + v^2n^2\|b_2\|^2\mathbb{E}\|\theta\|^4-\frac{2vm}{d}\cdot tr(A_1)\\
&&+\frac{2vn}{d}\cdot tr(A_2) -\frac{2v^2mn}{d} \cdot tr(A_1A_2^{\top}) -2v^2mn b_1^{\top}b_2\mathbb {E}\|\theta\|^4 + O(\frac{1}{D}) \\
&=& \frac{1}{d} tr(vmA_1-vnA_2-I)^2 + v^2(mb_1-nb_2)^2 \mathbb{E}\|\theta\|^4 + O(\frac{1}{D}).
\end{eqnarray*}

Therefore, the optimal solutions satisfies $\|vmA_1-vnA_2\|_F^2 = O(\frac{d}{D})$ and $\|mb_1-nb_2\|^2=O(\frac{1}{D}).$

Furthermore, we have
\begin{eqnarray*}
&&\mathbb{E}\left(y_q- f(E)_{d+1,D+1} 
 \right)^2\\
 &=& \mathbb{E}_{(x_q,\theta)}\left[ x_q^{\top}\theta)^2 +B_1+B_2+B_3+B_4+B_{51}+B_{52}+B_{53}+B_{54}+B_{55}+B_{56}+B_{57}\right]\\
&=& \mathbb{E}_{(x_q,\theta)}\Bigg[ (x_q^{\top}\theta)^2 + \frac{v^2m^2}{D}\theta^{\top}(I_d-(A_1+\theta b_1^{\top})x_q x_q^{\top}(A_1+\theta b_1^{\top})^{\top})\theta \exp(x_q^{\top}(A_1+\theta b_1^{\top})^{\top}(A_1+\theta b_1^{\top})x_q)  \\
 && +\frac{v^2n^2}{D}\theta^{\top}(I_d-(A_2+\theta b_2^{\top})x_q x_q^{\top}(A_2+\theta b_2^{\top})^{\top})\theta \exp(x_q^{\top}(A_2+\theta b_2^{\top})^{\top}(A_2+\theta b_2^{\top})x_q)  \\
 &&  + v^2m^2(\theta^{\top}(A_1+\theta b_1^{\top})x_q)^2 +v^2n^2(\theta^{\top}(A_2+\theta b_2^{\top})x_q)^2 -(2mv\theta^\top x_q)\theta^{\top}(A_1+\theta b_1^{\top})x_q +(2nv\theta^\top x_q)\theta^{\top}(A_2+\theta b_2^{\top})x_q\\
 && - \frac{2v^2m^2(\theta^{\top}(A_1+\theta b_1^{\top})x_q)^2\exp{(x_q^{\top}A_1x_q)}}{D \exp(x_q^{\top}(A_1+\theta b_1^{\top})^{\top}(A_1+\theta b_1^{\top})x_q/2)}- \frac{2v^2n^2(\theta^{\top}(A_2+\theta b_2^{\top})x_q)^2\exp{(x_q^{\top}A_2x_q)}}{D \exp(x_q^{\top}(A_2+\theta b_2^{\top})^{\top}(A_2+\theta b_2^{\top})x_q/2)} \\
 &&+ \frac{(2vm\theta^\top x_q)\theta^{\top}(A_1+\theta b_1^{\top})x_q\exp(x_q^{\top}A_1x_q)}{D\exp(x_q^{\top}(A_1+\theta b_1^{\top})^{\top}(A_1+\theta b_1^{\top})x_q/2)} - \frac{(2vn\theta^\top x_q)\theta^{\top}(A_2+\theta b_2^{\top})x_q\exp(x_q^{\top}A_2x_q)}{D\exp(x_q^{\top}(A_2+\theta b_2^{\top})^{\top}(A_2+\theta b_2^{\top})x_q/2)} \\
    &&+ \frac{2mv}{D}(\theta^\top x_q)\theta^\top(A_1+\theta b_1^\top)x_q \exp(x_q^{\top}(A_1+\theta b_1^{\top})^{\top}(A_1+\theta b_1^{\top})x_q)  \\
    &&- \frac{2nv}{D}(\theta^\top x_q)\theta^\top(A_2+\theta b_2^\top)x_q \exp(x_q^{\top}(A_2+\theta b_2^{\top})^{\top}(A_2+\theta b_2^{\top})x_q) -2v^2mn\theta^{\top}(A_1+\theta b_1)x_q \theta^{\top}(A_2+\theta b_2)x_q \\
&& +
2v^2mn\frac{1}{D} \frac{\exp(x_q^{\top}A_1x_q)\theta^{\top}(A_1+\theta b_1)x_q \theta^{\top}(A_2+\theta b_2)x_q}{\exp(x_q^{\top}(A_1+\theta b_1^{\top})^{\top}(A_1+\theta b_1^{\top})x_q/2)} +
2v^2mn\frac{1}{D} \frac{\exp(x_q^{\top}A_2x_q)\theta^{\top}(A_1+\theta b_1)x_q \theta^{\top}(A_2+\theta b_2)x_q}{\exp(x_q^{\top}(A_2+\theta b_2^{\top})^{\top}(A_2+\theta b_2^{\top})x_q/2)} \\
&& - \frac{2v^2mn}{D} \theta^{\top}\theta \exp(x_q^\top(A_1+\theta b_1^{\top})^{\top}(A_2+\theta b_2^{\top})x_q/2+x_q^\top(A_2+\theta b_2^{\top})^{\top}(A_1+\theta b_1^{\top})x_q/2) \\
&&-\frac{2v^2mn}{D} \exp(x_q^\top(A_1+\theta b_1^{\top})^{\top}(A_2+\theta b_2^{\top})x_q/2+x_q^\top(A_2+\theta b_2^{\top})^{\top}(A_1+\theta b_1^{\top})x_q/2)\left((\theta^{\top}(A_1+\theta b_1)^{\top}x_q \theta^{\top}(A_2+\theta b_2)^{\top}x_q) \right)\\
&&+ \frac{2v^2mn}{D}\theta^{\top}(A_1+\theta b_1)x_q \theta^{\top}(A_2+\theta b_2)x_q \exp(x_q^{\top}(A_2+\theta b_2^{\top})^{\top}(A_2+\theta b_2^{\top})x_q) \\
&&+\frac{2v^2mn}{D}\theta^{\top}(A_1+\theta b_1)x_q \theta^{\top}(A_2+\theta b_2)x_q\exp(x_q^{\top}(A_1+\theta b_1^{\top})^{\top}(A_1+\theta b_1^{\top})x_q)\Bigg]+o(\frac{1}{D}).\\
\end{eqnarray*}
When $m,n, A_1, A_2, b_1, b_2, v$ satisfies $vmA_1=vnA_2$ and $mb_1=mb_2$, we have

\begin{eqnarray*}
&&\mathbb{E}\left(y_q- f(E)_{d+1,D+1} 
 \right)^2\\
&=& 
\mathbb{E}_{(x_q,\theta)}\bigg[\frac{v^2m^2}{D}\exp(x_q^{\top}(A_1+\theta b_1^{\top})^{\top}(A_1+\theta b_1^{\top})x_q)\| \theta\|^2+\frac{v^2n^2}{D}\exp(x_q^{\top}(A_2+\theta b_2^{\top})^{\top}(A_2+\theta b_2^{\top})x_q)\| \theta\|^2 \\
&&-\frac{2v^2mn}{D} \exp(x_q^\top(A_1+\theta b_1^{\top})^{\top}(A_2+\theta b_2^{\top})x_q/2+x_q^\top(A_2+\theta b_2^{\top})^{\top}(A_1+\theta b_1^{\top})x_q/2)\| \theta\|^2 \\
&&-\frac{2v^2mn}{D} \exp(x_q^\top(A_1+\theta b_1^{\top})^{\top}(A_2+\theta b_2^{\top})x_q/2+x_q^\top(A_2+\theta b_2^{\top})^{\top}(A_1+\theta b_1^{\top})x_q/2)\left((\theta^{\top}(A_1+\theta b_1)^{\top}x_q \theta^{\top}(A_2+\theta b_2)^{\top}x_q) \right)\\
&&+ 
\frac{v^2m^2}{D}(\theta^{\top}(A_1+\theta b_1^{\top})x_q)^2\exp(x_q^{\top}(A_1+\theta b_1^{\top})^{\top}(A_1+\theta b_1^{\top})x_q) \\
&&+\frac{v^2n^2}{D}(\theta^{\top}(A_2+\theta b_2^{\top})x_q)^2\exp(x_q^{\top}(A_2+\theta b_2^{\top})^{\top}(A_2+\theta b_2^{\top})x_q)\bigg]+o(\frac{1}{D}).
\end{eqnarray*}
Taking $m=2$, $n=1$, $A_1=\frac{c}{v}I$, $A_2=\frac{2c-1}{v}I$ and $b_1=b_2=0,$
\begin{eqnarray}
&&\mathbb{E}\left(y_q- f(E)_{d+1,D+1} 
 \right)^2\nonumber \\
  &=& \frac{4v^2}{D}\left( (\frac{v^2}{v^2-2c^2})^{\frac{d}{2}}- (\frac{v^2}{v^2-2c(2c-1)})^{\frac{d}{2}}\right)+\frac{v^2}{D}(\frac{v^2}{v^2-2(2c-1)^2})^{\frac{d}{2}}\nonumber \\ &+&  \frac{(2c-1)^2}{D}(\frac{v^2}{v^2-2(2c-1)^2})(\frac{v^2}{v^2-2(2c-1)^2})^{\frac{d}{2}}\nonumber \\
  &+& \frac{4c^2}{D}  (\frac{v^2}{v^2-2c^2})  (\frac{v^2}{v^2-2c^2})^{\frac{d}{2}}-\frac{4(2c-1)c}{D}  (\frac{v^2}{v^2-2c(2c-1)}) (\frac{v^2}{v^2-2c(2c-1)})^{\frac{d}{2}} + o(\frac{1}{D}).
 \end{eqnarray}
 Assuming that $v^2>\max\{2c^2, 2(2c-1)^2\}$,
 when $0<c<1$, we have 
 
 $\mathbb{E}\left(y_q- f_{multi}(E)_{d+1,D+1} 
 \right)^2<\mathbb{E}\left(y_q- f_{sing}(E)_{d+1,D+1} 
 \right)^2.$
\end{proof}

\subsection{Proposition~\ref{prop:multi}}\label{sec:proof:prop1}
\begin{figure}[!ht]
    \centering\vspace{-0.1in}
    \includegraphics[scale=0.6]{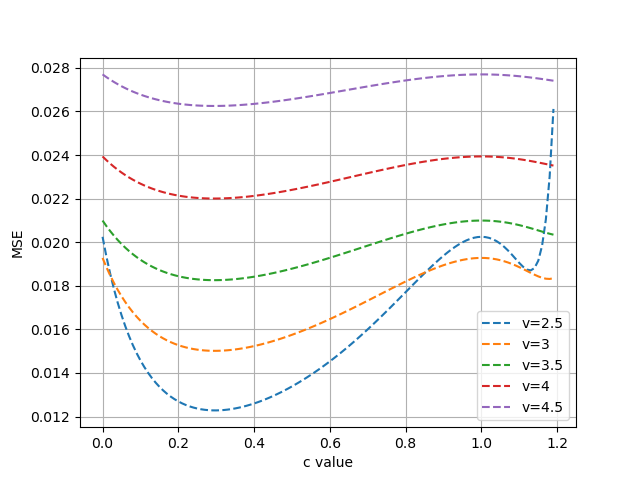}
    \caption{Theoretical loss of multi-head attention when taking different values of c. (D=1000, d=5) }
    \label{fig:prior_c}
\end{figure}

\begin{proof}[Proof of Proposition~\ref{prop:multi}] To differentiate the loss for single-head and multi-head attention, we use $L_{\text{sing}}$ and $L_{\text{multi}}$ to denote them respectively.

When $c=1$, the loss of multi-head attention indicated by Theorem~\ref{thm:multi_head} can be reduced to the optimal loss of single-head attention:
\begin{eqnarray*}
&&L_{\text{multi}}(A_1,A_2,b_1,b_2,v)\big|_{c=1}=\frac{v^2}{D}\left(\frac{v^2}{v^2-2}\right)^\frac{d}{2} + \frac{v^2}{D(v^2-2)}\left(\frac{v^2}{v^2-2}\right)^\frac{d}{2}+o(\frac{1}{D})\approx L_{\text{sing}}(A^*,b^*,v).
    \end{eqnarray*}
Upon differentiation, we have $\frac{\partial}{\partial c} L_{\text{multi}}\left( A_1, A_2, b_1, b_2, v \right)\big|_{c=1}=0$, and 
 \begin{eqnarray*}
&&\frac{\partial^2}{\partial c^2} L_{\text{multi}}\left( A_1, A_2, b_1, b_2, v \right)\big|_{c=1}\\
&=&-\frac{4v^2\left(d+2\right)^2}{D}\left(\frac{(\frac{v^2}{v^2-2})^{d/2}}{(v^2-2)^2}\right)-\frac{16v^2\left(d+2\right)}{D}\left(\frac{(\frac{v^2}{v^2-2})^{d/2}}{
(v^2-2)^3}\right)-\frac{8v^4\left(d+2\right)d}{D}\left(\frac{(\frac{v^2}{v^2-2})^{d/2}}{
(v^2-2)^4}\right)<0.
\end{eqnarray*} 
Therefore, when fixing other parameters, $c=1$ is a local maximum of the loss function, indicating that there must exist some $0<c^*<1$ such that $L_{\text{multi}}(A_1,A_2,b_1,b_2,v)\big|_{c=c^*} < L_{\text{multi}}(A_1,A_2,b_1,b_2,v)\big|_{c=1}\approx L_{\text{sing}}(A^*,b^*,v).$ 

In Figure \ref{fig:prior_c}, we also plot the value of $L_{\text{multi}}$ when changing $c$. One can see that when $c=1$, for all choices of $v$, 
$L_{\text{multi}}(\cdot,\cdot,\cdot,\cdot,v)\big|_{c=1}$ achieves its local maximum.
\end{proof}

\subsection{Prior Knowledge}\label{sec:proof:prior}

\begin{proof}[Proof of Theorem \ref{thm:prior}]

     When taking infinite many training samples (prompts), the loss function becomes
    \begin{eqnarray*}
    &&\mathbb{E}\left(y_q- (W^V_{d+1,:})^{\top}
    E\phi\left(E^{\top}(W^K)^{\top} W^Q \begin{bmatrix}
        x_q\\0
    \end{bmatrix} \right)\right)^2\\
    &=&\mathbb{E}\left(y_q-  \begin{bmatrix}
        u^{\top}x_1+vy_1,u^{\top}x_2+vy_2,\ldots,u^{\top}x_D+vy_D,u^{\top}x_q
    \end{bmatrix}\phi\left(E^{\top}(W^K)^{\top} W^Q \begin{bmatrix}
        x_q\\0
    \end{bmatrix} \right)\right)^2\\
    &=&\mathbb{E}_{(x_q,\theta)}\mathbb{E}_{\{x_i\}_{i\in[D]}}\left(y_q- \begin{bmatrix}
        u^{\top}x_1+vy_1,u^{\top}x_2+vy_2,\ldots,u^{\top}x_D+vy_D,u^{\top}x_q
    \end{bmatrix}\phi\left(\begin{bmatrix}
        x_1^{\top}Ax_q+y_1b^{\top}x_q\\\ldots\\x_q^{\top}A x_q+0
    \end{bmatrix} \right)\right)^2\\
    &=&\mathbb{E}_{(x_q,\theta)}\mathbb{E}_{\{x_i\}_{i\in[D]}}
    \left(y_q- \frac{ \sum_{i=1}^D 
  (u^\top + v\theta^{\top} )x_i\exp(x_i^{\top}Ax_q+y_ib^{\top}x_q) + u^\top x_q \exp(x_q^{\top}Ax_q)}{\sum \exp(x_i^{\top}Ax_q+y_ib^{\top}x_q) 
    +\exp(x_q^{\top}Ax_q)}
    \right)^2 \\
    &=&\mathbb{E}_{(x_q,\theta)}\mathbb{E}_{\{x_i\}_{i\in[D]}}
    \bigg(y_q^2\underbrace{-2y_q\left(\frac{ \sum_{i=1}^D 
  (u^\top + v\theta^{\top} )x_i\exp(x_i^{\top}Ax_q+y_ib^{\top}x_q) + u^\top x_q \exp(x_q^{\top}Ax_q)}{\sum \exp(x_i^{\top}Ax_q+y_ib^{\top}x_q) 
    +\exp(x_q^{\top}Ax_q)} \right)}_{A_1}
        \bigg)\\
    &+&\mathbb{E}_{(x_q,\theta)}\mathbb{E}_{\{x_i\}_{i\in[D]}}
   \underbrace{\left(\frac{ \sum_{i=1}^D 
  (u^\top + v\theta^{\top} )x_i\exp(x_i^{\top}Ax_q+y_ib^{\top}x_q) + u^\top x_q \exp(x_q^{\top}Ax_q)}{\sum \exp(x_i^{\top}Ax_q+y_ib^{\top}x_q) 
    +\exp(x_q^{\top}Ax_q)} \right)^2}_{A_2}.
    \end{eqnarray*}
    When fixing $x_q$ and $\theta$, the terms $A_1$ becomes
        \begin{eqnarray*}
         &&\mathbb{E}_{y_q}\mathbb{E}_{\{x_i,y_i\}_{i\in[D]}}A_1 \\
        &=&\mathbb{E}_{y_q}\mathbb{E}_{\{x_i,y_i\}_{i\in[D]}} -2y_q\bigg(\frac{\sum_{i=1}^D 
        (u^\top + v\theta^{\top} ) x_i\exp(x_i^{\top}Ax_q+y_ib^{\top}x_q)+u^\top x_q \exp(x_q^{\top}Ax_q) }{\sum \exp(x_i^{\top}Ax_q+y_ib^{\top}x_q) 
        +\exp(x_q^{\top}Ax_q)}\bigg)\\
        &=&\mathbb{E}_{\{x_i,y_i\}_{i\in[D]}} -2\theta^\top x_q \bigg(\frac{u^\top x_q \exp(x_q^{\top}Ax_q)}{\sum \exp(x_i^{\top}Ax_q+y_ib^{\top}x_q)     +\exp(x_q^{\top}Ax_q)}\bigg)- 2\theta^\top x_q \bigg(\frac{\sum_{i=1}^D 
         (u^\top + v\theta^{\top} ) x_i\exp(x_i^{\top}Ax_q+y_ib^{\top}x_q) }{\exp(x_q^{\top}Ax_q)+D\mathbb{E}\exp({x_1}^{\top}Ax_q+y_1b^{\top}x_q)}\bigg)\\
    &+&2\theta^\top x_q\left(\frac{\sum_{i=1}^D 
    (u^\top + v\theta^{\top} ) x_i\exp(x_i^{\top}Ax_q+y_ib^{\top}x_q) }{(\exp(x_q^{\top}Ax_q)+D\mathbb{E}\exp({x_1}^{\top}Ax_q+y_1b^{\top}x_q))^2}\left(\sum \exp(x_i^{\top}Ax_q+y_ib^{\top}x_q)-D\mathbb{E}\exp({x_1}^{\top}Ax_q+y_1b^{\top}x_q)\right)\right)\\
    &-&2\theta^\top x_q\left(\frac{\sum_{i=1}^D (u^\top + v\theta^{\top} ) x_i\exp(x_i^{\top}Ax_q+y_ib^{\top}x_q) }{(\exp(x_q^\top A x_q)+D\mathbb{E}\exp({x_1}^{\top}Ax_q+y_1b^{\top}x_q))^3}\left(\sum \exp(x_i^{\top}Ax_q+y_ib^{\top}x_q)-D\mathbb{E}\exp({x_1}^{\top}Ax_q+y_1b^{\top}x_q)\right)^2\right)\\
    &=& A_{11}+A_{12}+A_{13}+A_{14}.
    \end{eqnarray*}
For the terms $A_{11}$ to $A_{14}$, we have
   \begin{eqnarray*}
A_{11}&=&
-2\theta^\top x_q\left(\frac{u^\top x_q \exp(x_q^{\top}Ax_q)}{D\exp(x_q^{\top}(A+\theta b^{\top})^{\top}(A+\theta b^{\top}) x_q/2)} \right)+o(\frac{1}{D}), 
\end{eqnarray*}
    \begin{eqnarray*}
A_{12}&=& -2\theta^\top x_q(u^\top + v\theta^{\top} )(A+\theta b^{\top})x_q +\frac{2\theta^\top x_q(u^\top + v\theta^{\top} )(A+\theta b^{\top})x_q \exp(x_q^{\top}Ax_q)}{D\exp(x_q^{\top}(A+\theta b^{\top})^{\top}(A+\theta b^{\top})x_q/2)}+o(\frac{1}{D}),
\end{eqnarray*}
    \begin{eqnarray*}
A_{13}&=& -\frac{2}{D} \theta^\top x_q(u^\top + v\theta^{\top} )(A+\theta b^{\top})x_q +\frac{4}{D}\theta^\top x_q((u^\top+v\theta^{\top})(A+\theta b^{\top})x_q) \exp(x_q^{\top}(A+\theta b^{\top})^{\top}(A+\theta b^{\top})x_q)),
\end{eqnarray*}
and
    \begin{eqnarray*}
A_{14}&=& \frac{2}{D} \theta^\top x_q(u^\top + v\theta^{\top} )(A+\theta b^{\top})x_q -\frac{2}{D}\theta^\top x_q((u^\top+v\theta^{\top})(A+\theta b^{\top})x_q) \exp(x_q^{\top}(A+\theta b^{\top})^{\top}(A+\theta b^{\top})x_q)).
\end{eqnarray*}
In terms of $A_2$, when fixing $x_q$ and $\theta$, we have
\begin{eqnarray*}
&&\mathbb{E}_{\{x_i,y_i\}_{i\in[D]}}A_2\\
&=&\mathbb{E}_{\{x_i,y_i\}_{i\in[D]}}\left(\frac{ \sum_{i=1}^D 
  (u^\top + v\theta^{\top} )x_i\exp(x_i^{\top}Ax_q+y_ib^{\top}x_q) + u^\top x_q \exp(x_q^{\top}Ax_q)}{\sum \exp(x_i^{\top}Ax_q+y_ib^{\top}x_q) 
    +\exp(x_q^{\top}Ax_q)} \right)^2 \\
&=&\mathbb{E}_{\{x_i,y_i\}_{i\in[D]}}\left(\frac{ \sum_{i=1}^D 
  (u^\top + v\theta^{\top} )x_i\exp(x_i^{\top}Ax_q+y_ib^{\top}x_q)}{\sum \exp(x_i^{\top}Ax_q+y_ib^{\top}x_q)+\exp(x_q^{\top}Ax_q)} \right)^2 \\
  &+& \mathbb{E}_{\{x_i,y_i\}_{i\in[D]}}\left(\frac{u^\top x_q \exp(x_q^{\top}Ax_q)}{\sum \exp(x_i^{\top}Ax_q+y_ib^{\top}x_q)+\exp(x_q^{\top}Ax_q)} \right)^2 \\
  &+&\mathbb{E}_{\{x_i,y_i\}_{i\in[D]}}\bigg(2u^\top x_q \exp(x_q^{\top}Ax_q)\bigg)\frac{ \sum_{i=1}^D 
  (u^\top + v\theta^{\top} )x_i\exp(x_i^{\top}Ax_q+y_ib^{\top}x_q)}{\left(\sum \exp(x_i^{\top}Ax_q+y_ib^{\top}x_q)+\exp(x_q^{\top}Ax_q)\right)^2}  \\
  &=&A_{21}+A_{22}+A_{23}, 
    \end{eqnarray*}
    where
    \begin{eqnarray*}
    A_{21}&=& \frac{1}{D}(u^\top+v\theta^{\top})(u+v\theta) \exp(x_q^{\top}(A+\theta b^{\top})^{\top}(A+\theta b^{\top})x_q) +\bigg((u^\top+v\theta^{\top})(A+\theta b^{\top})x_q\bigg)^2 \\
  &-& \frac{2\left((u^\top +v\theta^{\top})(A+\theta b^{\top})x_q\right)^2\exp{(x_q^{\top}Ax_q)}}{D \exp(x_q^{\top}(A+\theta b^{\top})^{\top}(A+\theta b^{\top})x_q/2)} - \frac{1}{D}\left((u^\top+v\theta^{\top})(A+\theta b^{\top})x_q\right)^2\exp(x_q^{\top}(A+\theta b^{\top})^{\top}(A+\theta b^{\top})x_q),\\
    \end{eqnarray*}
        \begin{eqnarray*}
    &&A_{22}= o(\frac{1}{D}),
    \end{eqnarray*}
        \begin{eqnarray*}
    A_{23}&=&
    \frac{2u^\top x_q\exp(x_q^{\top}Ax_q)(u^\top + v\theta^{\top} )(A+\theta b^{\top})x_q}{D\exp(x_q^{\top}(A+\theta b^{\top})^{\top}(A+\theta b^{\top})x_q/2)}+o(\frac{1}{D}).\\ 
    \end{eqnarray*}
        As a result,
        \begin{eqnarray*}
    &&\mathbb{E}\left(y_q- (W^V_{d+1,:})^{\top}
        E\phi\left(E^{\top}(W^K)^{\top} W^Q \begin{bmatrix}
            x_q\\0
        \end{bmatrix} \right)\right)^2\\
    &=&\mathbb{E}_{(x_q,\theta)}(x_q^{\top}\theta)^2+((u^\top + v\theta^{\top} )(A+\theta b^{\top})x_q)^2-2(x_q^{\top}\theta)((u^\top + v\theta^{\top} )(A+\theta b^{\top})x_q)+O\left(\frac{1}{D}\right)\\
    &=&\mathbb{E}_{(x_q,\theta)}\bigg([(u^\top + v\theta^{\top} )(A+\theta b^{\top})-\theta^\top)]x_q\bigg)^2+O\left(\frac{1}{D}\right)\\
    &=& u^\top AA^\top u +\mathbb{E}_{\theta}\bigg( v^2\theta^\top A A^\top \theta- 2v\theta^\top A^\top \theta+2\theta^\top (vA-I) A^\top u+\|\theta\|^2 +2u^\top A b\theta^\top (u+v\theta) \bigg)\\
    &&+ \mathbb{E}_{\theta}\bigg(2\theta^\top(vA-I)b \theta^\top(u+v\theta)\bigg)+\|b\|^2 \mathbb{E}_{\theta} \bigg( (u^\top+v\theta^\top )\theta \theta^\top (u+v\theta)\bigg) \\
    &=&\|A^\top u\|^2 +\frac{\sigma^2}{d} tr((vA-I)^2)+\theta_0^\top ((vA-I)^2)\theta_0+2\theta_0^\top (vA-I) A^\top u + \underbrace{2(\theta_0^\top u+v\|\theta_0\|^2+v\sigma^2)u^\top}_{b_1^\top} Ab \\
    &&+(    \underbrace{\frac{4v\sigma^2}{d}\theta_0^\top+2v\sigma^2\theta_0^\top +2v\|\theta_0\|^2\theta_0^\top+\frac{2\sigma^2}{d}u^\top+2u^\top\theta_0\theta_0^\top) }_{b_2^\top}(vA-I)b  \\
    &&+ \|b\|^2\underbrace{(\frac{\sigma^2}{d}\|u\|^2+\|u^\top\theta_0\|^2+
    \frac{2v^2\sigma^4}{d}+\frac{4v^2\sigma^2}{d}\|\theta_0\|^2+    v^2(\sigma^2+\|\theta_0\|^2)^2+2v u^\top (\frac{2\sigma^2}{d}\theta_0+\sigma^2\theta_0+\|\theta_0\|^2\theta_0))}_{a_1} \\
    &&+O(\frac{1}{D}).
        \end{eqnarray*}
Therefore, to minimize the loss, assuming that $A$, $u$, $v$ are fixed, the optimal $b$ satisfies 

$b^*=-\frac{1}{2a_1}\bigg(A^\top b_1+(vA-I)^\top b_2\bigg)$.

Then we have
    \begin{eqnarray*}
    &&\mathbb{E}\left(y_q- (W^V_{d+1,:})^{\top}
        E\phi\left(E^{\top}(W^K)^{\top} W^Q \begin{bmatrix}
            x_q\\0
        \end{bmatrix} \right)\right)^2\\
    &=& \|A^\top u\|+\frac{\sigma^2}{d} tr((vA-I)^2)+\theta_0^\top ((vA-I)^2)\theta_0+2\theta_0^\top (vA-I) A^\top u \\
    && -\frac{1}{4a_1}\bigg[ b_1^\top AA^\top b_1+b_2^\top (vA-I)^2b_2 +2b_1^\top A(vA-I)^\top b_2 \bigg]+ O(\frac{1}{D}). \\
    \end{eqnarray*}

Assuming that $v$ and $\theta_0$ are fixed, the optimal $A^*$ should satisfies
 \begin{eqnarray*}
\bigg[\left(u^\top u+v\theta_0 u^\top -\frac{1}{4a_1}b_1 b_1^\top-\frac{v}{4a_1}b_1 b_2^\top\right)&+&v \left(v\frac{\sigma^2}{d}I+v\theta_0 \theta_0^\top +u\theta_0^\top-\frac{v}{4a_1}b_2b_2^\top-\frac{1}{4a_1}b_2b_1^\top\right)\bigg]A^*\\
&-&\left(v\frac{\sigma^2}{d}I+v\theta_0 \theta_0^\top +u\theta_0^\top-\frac{v}{4a_1}b_2b_2^\top-\frac{1}{4a_1}b_2b_1^\top\right)=O(\frac{1}{D}).
 \end{eqnarray*}

 \begin{itemize}
     \item When $\sigma^2 \gg O(\frac{1}{D})$: 
     \\
     If there exist an optimal $A^*$ which can minimize $\mathbb{E}\left(y_q- f(E)_{d+1,D+1} 
 \right)^2$ , 
 it is required that $vA^*-I=O(\frac{1}{D})$ and $\left(u^\top u+v\theta_0 u^\top -\frac{1}{4a_1}b_1 b_1^\top-\frac{v}{4a_1}b_1 b_2^\top\right)=O(\frac{1}{D})$.
 From
 $\left(u^\top u+v\theta_0 u^\top -\frac{1}{4a_1}b_1 b_1^\top-\frac{v}{4a_1}b_1 b_2^\top\right)=O(\frac{1}{D})$, we have $\left(\|u\|^2+2v^2\theta_0^2+2vu^\top \theta_0 \right)\left( uu^\top + v\theta_0u^\top\right)-v\left(\theta_0^\top u +v\theta_0^2+v\sigma^2 \right)u u^\top=O(\frac{1}{D}),$ which indicates that $u\parallel \theta_0$. 
 \\\\
 If we let $u=c_u\theta_0$, we have $c_u(c_u+2v)^2(c_u+v)\|\theta_0\|^2+c_u^2v^2\sigma^2=O(\frac{1}{D}).$
 \begin{itemize}
  \item{If $c_u^2= O(\frac{1}{D})$, substituting $u=c_u\theta_0$ and $vA-I=O(\frac{1}{D})$ to $\mathbb{E}\left(y_q- f(E)_{d+1,D+1}\right)^2$, we have 
      
       $\mathbb{E}\left(y_q- f(E)_{d+1,D+1}\right)^2=\underbrace{\frac{c^2\frac{\sigma^2}{d}\|\theta_0\|^2\left(\|\theta_0\|^2(c+2v)^2+2v^2\sigma^2\right)}{v^2\left((c+v)^2\|\theta_0\|^4+\frac{\sigma^2}{d}\|\theta_0\|^2(c+2v)^2+v^2\sigma^4(1+\frac{2}{d})+2v(c+v)\sigma^2\|\theta_0\|^2\right)}}_{=O(\frac{1}{D})}+O(\frac{1}{D})=O(\frac{1}{D}).$
 \item{If $c_u^2 >  O(\frac{1}{D})$, we have $(c_u+v)\|\theta_0\|^2+\frac{c_u v^2\sigma^2}{(c+2v)^2}=O(\frac{1}{D})$. }

 $\mathbb{E}\left(y_q- f(E)_{d+1,D+1}\right)^2=\underbrace{\frac{c^2\frac{\sigma^2}{d}\|\theta_0\|^2\left(\|\theta_0\|^2(c+2v)^2+2v^2\sigma^2\right)}{v^2\left((c+v)^2\|\theta_0\|^4+\frac{\sigma^2}{d}\|\theta_0\|^2(c+2v)^2+v^2\sigma^4\frac{2}{d}+\frac{v^2\sigma^4(c+v)^2}{(c+2v)^2}\right)}}_{>0 \text{ and } > O(\frac{1}{D})}+O(\frac{1}{D})$.}
    \end{itemize}

 Therefore, in order to minimize $\mathbb{E}\left(y_q- f(E)_{d+1,D+1}\right)^2$, it is required that $c_u=O(\frac{1}{\sqrt{D}})$. Then we have $b=O(\frac{1}{\sqrt{D}})$.

 \item When $\sigma^2 \ll O(\frac{1}{D})$: As long as $A$, $b$, $u$ and $v$ satisfies 
 \begin{eqnarray}
 2a_1b+\left(A^\top b_1+(vA-I)^\top b_2\right)=0,
 \label{eqn:appendix:prior:condition}    
 \end{eqnarray}
  we have $\mathbb{E}\left(y_q- f(E)_{d+1,D+1} 
 \right)^2=O(\frac{1}{D})$.
 \end{itemize}

 \begin{figure}[!ht]
    \centering\vspace{-0.1in}
    \includegraphics[scale=0.3]{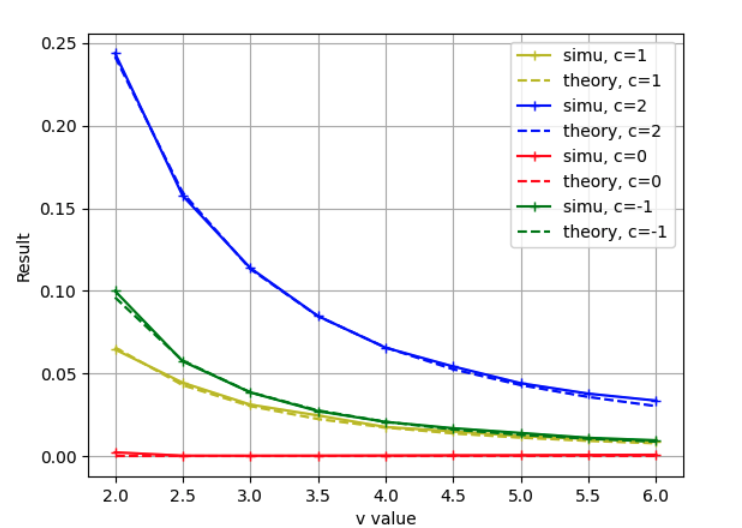}\vspace{-0.1in}
    \caption{Simultion: when A=I/v, the loss is minimized at c=0. }
    \label{fig:prior}
\end{figure}

   
When taking $A=I_d/v$, $b=0$ and $u=0$ we have
\begin{eqnarray*}
A_{11}&=& o(\frac{1}{D}), \\
A_{12}&=& -2 \mathbb{E} (\theta^\top x_qx_q^\top \theta)+\frac{2}{D}\mathbb{E}\frac{\theta^\top x_qx_q^\top \theta \exp(x_q^{\top}x_q/v)}{\exp(x_q^{\top}x_q/2v^2)}+o(\frac{1}{D}), \\
A_{13}+A_{14}&=&\frac{2}{D}\mathbb{E}\theta^\top x_qx_q^\top \theta \exp(x_q^{\top}x_q/v^2) + o(\frac{1}{D}),\\
A_{21}&=& \frac{v^2}{D} \mathbb{E} \theta^\top \theta \exp(x_q^{\top}x_q/v^2)+\mathbb{E}\theta^\top x_q x_q^\top \theta - \mathbb{E}\frac{2}{D}\frac{\left( \theta^\top x_q x_q^\top \theta \right)\exp{(x_q^{\top}x_q/v)}}{ \exp(x_q^{\top}x_q/2v^2)} - \frac{1}{D}\mathbb{E}\left(\theta^\top x_q x_q^\top \theta\right)\exp(x_q^{\top}x_q/v^2)+o(\frac{1}{D}),\\
A_{22}&=& o(\frac{1}{D}), \\
A_{23}&=& o(\frac{1}{D}). \\
\end{eqnarray*}
As a result,
\begin{eqnarray*}
    &&\mathbb{E}\left(y_q- (W^V_{d+1,:})^{\top}
    E\phi\left(E^{\top}(W^K)^{\top} W^Q \begin{bmatrix}
        x_q\\0
    \end{bmatrix} \right)\right)^2\\
    &=&\mathbb{E} (\theta^\top x_qx_q^\top \theta)+A_{12}+A_{13}+A_{14}+A_{21}\\
    &=& \frac{1}{D}\mathbb{E}\theta^\top x_qx_q^\top \theta \exp(x_q^{\top}x_q/v^2) + \frac{v^2}{D} \mathbb{E} \theta^\top \theta \exp(x_q^{\top}x_q/v^2)+ o(\frac{1}{D}). \\
    &=&\frac{v^2(\sigma^2+\|\theta_0\|^2)}{D}(\frac{v^2}{v^2-2})^\frac{d}{2} + \frac{v^2(\sigma^2+\|\theta_0\|^2)}{D(v^2-2)}(\frac{v^2}{v^2-2})^\frac{d}{2}. 
\end{eqnarray*}

Figure \ref{fig:simu_pri_singD}, \ref{fig:simu_pri_singd} below demonstrate the theoretical values and the corresponding simulation results, which indicates that the simulation of prediction loss aligns well with theoretical values. 

\begin{figure}[!ht]
    \centering\vspace{-0.1in}
    \includegraphics[scale=0.45]{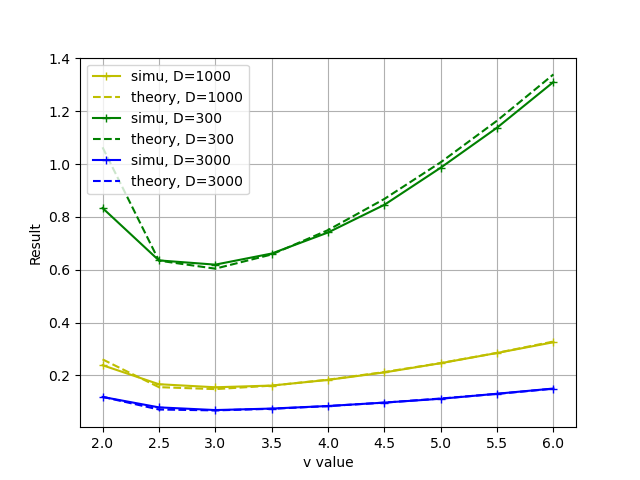}\vspace{-0.2in}
    \caption{ICL performance of single-head attention with prior knowledge, $(A,b,u)=(I_d/v,0,0)$ and $d=5$.}
    \label{fig:simu_pri_singD}
\end{figure}

\begin{figure}[!ht]
    \centering\vspace{-0.1in}
    \includegraphics[scale=0.45]{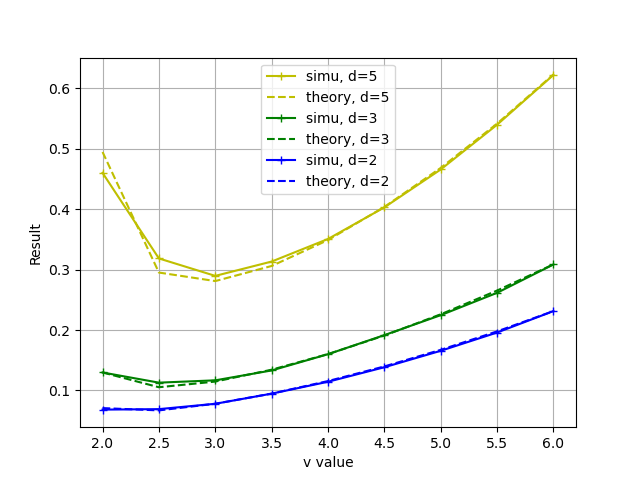}\vspace{-0.2in}
    \caption{ICL performance of single-head attention with prior knowledge, $(A,b,u)=(I_d/v,0,0)$ and $D=1000$.}
    \label{fig:simu_pri_singd}
\end{figure}

    \paragraph{ICL performance of multi-head attention}

    \begin{eqnarray*}
 &&\mathbb{E}\left(y_q- f(E)_{d+1,D+1} 
 \right)^2\\
&=&\mathbb{E}\bigg(y_q- m \begin{bmatrix}     u^{\top}x_1+vy_1,u^{\top}x_2+vy_2,\ldots,u^{\top}x_D+vy_D,u^{\top}x_q
    \end{bmatrix}\phi\left(E^{\top}(W_1^K)^{\top} W_1^Q \begin{bmatrix}
        x_q\\0
    \end{bmatrix} \right) \\
    && \qquad + n \begin{bmatrix}     u^{\top}x_1+vy_1,u^{\top}x_2+vy_2,\ldots,u^{\top}x_D+vy_D,u^{\top}x_q
    \end{bmatrix}\phi\left(E^{\top}(W_2^K)^{\top} W_2^Q \begin{bmatrix}
        x_q\\0
    \end{bmatrix} \right)\bigg)^2\\
    &=& \mathbb{E}\bigg(y_q-\frac{m \sum_{i=1}^D 
   (u^\top x_i+vy_i)\exp(x_i^{\top}(A_1+\theta b_1^{\top})x_q)+u^\top x_q \exp(x_q^{\top}A_1 x_q) }{\sum \exp(x_i^{\top}A_1x_q+y_ib_1^{\top}x_q) 
    +\exp(x_q^{\top}A_1x_q)}\\
    &&\qquad  +\frac{n\sum_{i=1}^D 
    (u^\top x_i+vy_i)\exp(x_i^{\top}(A_2+\theta b_2^{\top})x_q)+u^\top x_q \exp(x_q^{\top}A_2 x_q)}{\sum \exp(x_i^{\top}A_2x_q+y_ib_2^{\top}x_q) 
    +\exp(x_q^{\top}A_2x_q)}\bigg)^2\\
    &=& \mathbb{E}\left(y_q^2+ \left(\frac{m \sum_{i=1}^D 
   (u^\top x_i+vy_i)\exp(x_i^{\top}(A_1+\theta b_1^{\top})x_q)+u^\top x_q \exp(x_q^{\top}A_1 x_q) }{\sum \exp(x_i^{\top}A_1x_q+y_ib_1^{\top}x_q) 
    +\exp(x_q^{\top}A_1x_q)}\right)^2 \right)\\
     && \qquad+\mathbb{E}\left(\frac{n\sum_{i=1}^D 
    (u^\top x_i+vy_i)\exp(x_i^{\top}(A_2+\theta b_2^{\top})x_q)+u^\top x_q \exp(x_q^{\top}A_2 x_q)}{\sum \exp(x_i^{\top}A_2x_q+y_ib_2^{\top}x_q) 
    +\exp(x_q^{\top}A_2x_q)}\right)^2\\ &&\qquad-\mathbb{E} \left(2y_q\left(\frac{m \sum_{i=1}^D 
   (u^\top x_i+vy_i)\exp(x_i^{\top}(A_1+\theta b_1^{\top})x_q)+u^\top x_q \exp(x_q^{\top}A_1 x_q) }{\sum \exp(x_i^{\top}A_1x_q+y_ib_1^{\top}x_q) 
    +\exp(x_q^{\top}A_1x_q)}\right)\right)\\
&&\qquad+\mathbb{E} \left(2y_q\left(\frac{n\sum_{i=1}^D 
    (u^\top x_i+vy_i)\exp(x_i^{\top}(A_2+\theta b_2^{\top})x_q)+u^\top x_q \exp(x_q^{\top}A_2 x_q)}{\sum \exp(x_i^{\top}A_2x_q+y_ib_2^{\top}x_q) 
    +\exp(x_q^{\top}A_2x_q)}\right)\right)
\\  
 && \qquad-\mathbb{E}\left(\frac{m \sum_{i=1}^D 
   (u^\top x_i+vy_i)\exp(x_i^{\top}(A_1+\theta b_1^{\top})x_q)+u^\top x_q \exp(x_q^{\top}A_1 x_q) }{\sum \exp(x_i^{\top}A_1x_q+y_ib_1^{\top}x_q) 
    +\exp(x_q^{\top}A_1x_q)}\right)\\
    &&\qquad\times\mathbb{E}\left(\frac{n\sum_{i=1}^D 
    (u^\top x_i+vy_i)\exp(x_i^{\top}(A_2+\theta b_2^{\top})x_q)+u^\top x_q \exp(x_q^{\top}A_2 x_q)}{\sum \exp(x_i^{\top}A_2x_q+y_ib_2^{\top}x_q) 
    +\exp(x_q^{\top}A_2x_q)}\right)
    .
        \end{eqnarray*}
    When taking $m=2$, $n=1$, $A_1=\frac{c}{v}I$, $A_2=\frac{2c-1}{v}I$ and $u=b_1=b_2=0$, it becomes
 \begin{eqnarray*}
    &&\mathbb{E}\left(y_q- f(E)_{d+1,D+1} 
 \right)^2\\
 &=& \sigma^2+\|\theta_0\|^2+\mathbb{E}\left( \left(\frac{2v\sum_{i=1}^D 
    y_i\exp(x_i^{\top}x_q(c/v))}{\sum \exp(x_i^{\top}x_q(c/v)) 
    +\exp(\|x_q\|^2(c/v))}\right)^2 +\left(\frac{v\sum_{i=1}^D
    y_i\exp(x_i^{\top}x_q(2c-1)/v) }{\sum \exp(x_i^{\top}x_q(2c-1)/v) 
    +\exp(\|x_q\|^2(2c-1)/v)}\right)^2
    \right) \\
     &&+ \mathbb{E}\left( 
    2y_q\left(\frac{v\sum_{i=1}^D 
    y_i\exp(x_i^{\top}x_q(2c-1)/v) }{\sum \exp(x_i^{\top}x_q(2c-1)/v) 
    +\exp(\|x_q\|^2(2c-1)/v)}\right)    -2y_q\left(\frac{2v\sum_{i=1}^D 
    \theta^{\top} x_i\exp(x_i^{\top}x_q(c/v)) }{\sum \exp(x_i^{\top}x_q(c/v)) 
    +\exp(\|x_q\|^2(c/v))}\right)
    \right) \\  
 &&- \mathbb{E}\left(\frac{2v\sum_{i=1}^D 
    y_i\exp(x_i^{\top}x_q(c/v)) }{\sum \exp(x_i^{\top}x_q(c/v)) 
    +\exp(\|x_q\|^2(c/v))} \frac{2v\sum_{i=1}^D 
    y_i\exp(x_i^{\top}x_q(2c-1)/v) }{\sum \exp(x_i^{\top}x_q(2c-1)/v) 
    +\exp(\|x_q\|^2(2c-1)/v)}
    \right)\\
    &=&\sigma^2+\|\theta_0\|^2+B_1+B_2+B_3.
    \end{eqnarray*} 
Then we have
\begin{eqnarray*}
B_1&=&\mathbb{E}\left( \left(\frac{2v\sum_{i=1}^D 
    \theta^\top x_i\exp(x_i^{\top}x_q(c/v))}{\sum \exp(x_i^{\top}x_q(c/v)) 
    +\exp(\|x_q\|^2(c/v))}\right)^2 +\left(\frac{v\sum_{i=1}^D
   \theta^\top x_i\exp(x_i^{\top}x_q(2c-1)/v) }{\sum \exp(x_i^{\top}x_q(2c-1)/v) 
    +\exp(\|x_q\|^2(2c-1)/v)}\right)^2
    \right) \\
    &=& \frac{4}{D}v^2\|\theta\|^2 \exp(c^2\|x_q\|^2/v^2) +4c^2(\sigma^2+\|\theta_0\|^2) - \frac{8c^2(\theta^\top x_qx_q^\top\theta)\exp{(c^2\|x_q\|^2/v^2)}}{D \exp(c^2\|x_q\|^2/2v^2)} \\
    &&- \frac{4c}{D}\left(\theta^\top x_qx_q^\top \theta\right)\exp(c^2\|x_q\|^2/v^2)+\frac{1}{D}v^2\|\theta\|^2 \exp\left((2c-1)^2\|x_q\|^2/v^2\right) +(2c-1)^2(\sigma^2+\|\theta_0\|^2)  \\
    &&- \frac{2(2c-1)^2(\theta^\top x_qx_q^\top\theta)\exp{(((2c-1)^2\|x_q\|^2/v^2))}}{D \exp((2c-1)^2\|x_q\|^2/2v^2)}- \frac{2c-1}{D}\left(\theta^\top x_qx_q^\top \theta\right)\exp((2c-1)^2\|x_q\|^2/v^2),
   \end{eqnarray*} 
\begin{eqnarray*}
B_2&=&\mathbb{E}\left( 
    2y_q\left(\frac{v\sum_{i=1}^D 
    y_i\exp(x_i^{\top}x_q(2c-1)/v) }{\sum \exp(x_i^{\top}x_q(2c-1)/v) 
    +\exp(\|x_q\|^2(2c-1)/v)}\right)    -2y_q\left(\frac{2v\sum_{i=1}^D 
    \theta^{\top} x_i\exp(x_i^{\top}x_q(c/v)) }{\sum \exp(x_i^{\top}x_q(c/v)) 
    +\exp(\|x_q\|^2(c/v))}\right)
    \right)\\
&=&2(2c-1)\theta^\top x_qx_q^
\top \theta-\frac{2(2c-1)\theta^\top x_qx_q^
\top \theta \exp((2c-1)\|x_q\|^2/v)}{D\exp((2c-1)^2\|x_q\|^2/2v^2)}-\frac{2(2c-1)}{D}\theta^\top x_qx_q^
\top \theta \exp((2c-1)^2\|x_q\|^2/v^2))\\
&-&4c\theta^\top x_qx_q^
\top \theta+\frac{4c\theta^\top x_qx_q^
\top \theta \exp(c\|x_q\|^2/v)}{D\exp(c^2\|x_q\|^2/2v^2)}+\frac{4c}{D}\theta^\top x_qx_q^
\top \theta \exp(c^2\|x_q\|^2/v^2)),\\
\end{eqnarray*}
and
\begin{eqnarray*}
B_3&=&-4c(2c-1)(1-\frac{1}{D}) \theta^{\top}x_q x_q^\top \theta+
\frac{4}{D} \frac{c(2c-1)\theta^{\top}x_q x_q^\top \theta \exp(c\|x_q\|^2/v)}{\exp(c^2\|x_q\|^2/2v^2)} +
\frac{4}{D} \frac{c(2c-1)\theta^{\top}x_q x_q^\top \theta \exp((2c-1)\|x_q\|^2/v)}{\exp((2c-1)^2\|x_q\|^2/2v^2} \\
&+& \frac{4}{D}c(2c-1)\theta^{\top}x_qx_q^\top \theta \left(2\exp(c^2\|x_q\|^2/v^2)-2\right) + \frac{4}{D}c(2c-1)\theta^{\top}x_qx_q^\top \theta\left(2\exp((2c-1)^2\|x_q\|^2/v^2)-2\right) \\
&-&  \frac{4v^2}{D} \theta^{\top}\theta\exp((2c-1)c\|x_q\|^2/v^2)-\frac{4}{D}c(2c-1)(\theta^\top x_q x_q^\top \theta) \exp((2c-1)c\|x_q\|^2/v^2)\\ 
&+& \frac{4c^2}{D}(\theta^{\top}x_qx_q^\top\theta)\exp(c^2\|x_q\|^2/v^2) +\frac{1}{D}(2c-1)^2(\theta^{\top}x_qx_q^\top\theta)\exp((2c-1)^2\|x_q\|^2/v^2),\\
\end{eqnarray*}
To sum up, we have
    \begin{eqnarray*}
&&\mathbb{E}\left(y_q- f(E)_{d+1,D+1} 
 \right)^2\\
  &=& \frac{4}{D}v^2\|\theta\|^2 \exp(c^2\|x_q\|^2/v^2) +\frac{1}{D}v^2\|\theta\|^2 \exp\left((2c-1)^2\|x_q\|^2/v^2\right)\\
  &+& \frac{4}{D}c^2 \theta^\top x_qx_q^\top \theta \exp(c^2\|x_q\|^2/v^2) +\frac{1}{D}(2c-1)^2 \theta^\top x_qx_q^\top \theta \exp\left((2c-1)^2\|x_q\|^2/v^2\right)\\ &+&4v^2\|\theta\|^2\exp((2c-1)c\|x_q\|^2/v^2)-\frac{4}{D}c(2c-1)(\theta^\top x_q x_q^\top \theta) \exp((2c-1)c\|x_q\|^2/v^2)\\
    &=& \frac{4v^2(\sigma^2+\|\theta_0\|^2)}{D}\left( (\frac{v^2}{v^2-2c^2})^{\frac{d}{2}}- (\frac{v^2}{v^2-2c(2c-1)})^{\frac{d}{2}}\right)+\frac{v^2}{D}(\sigma^2+\|\theta_0\|^2)(\frac{v^2}{v^2-2(2c-1)^2})^{\frac{d}{2}} \\ &+&  \frac{(2c-1)^2}{D}(\sigma^2+\|\theta_0\|^2)(\frac{v^2}{v^2-2(2c-1)^2})(\frac{v^2}{v^2-2(2c-1)^2})^{\frac{d}{2}} \\
  &+& \frac{4c^2}{D} (\sigma^2+\|\theta_0\|^2) (\frac{v^2}{v^2-2c^2})  (\frac{v^2}{v^2-2c^2})^{\frac{d}{2}}-\frac{4(2c-1)c}{D}(\sigma^2+\|\theta_0\|^2)  (\frac{v^2}{v^2-2c(2c-1)}) (\frac{v^2}{v^2-2c(2c-1)})^{\frac{d}{2}}.
 \end{eqnarray*}

 Figure \ref{fig:simu_pri_multi} below demonstrates the alignment between the theoretical values and the corresponding simulation results.
 \begin{figure}[!ht]
    \centering\vspace{-0.1in}
    \includegraphics[scale=0.45]{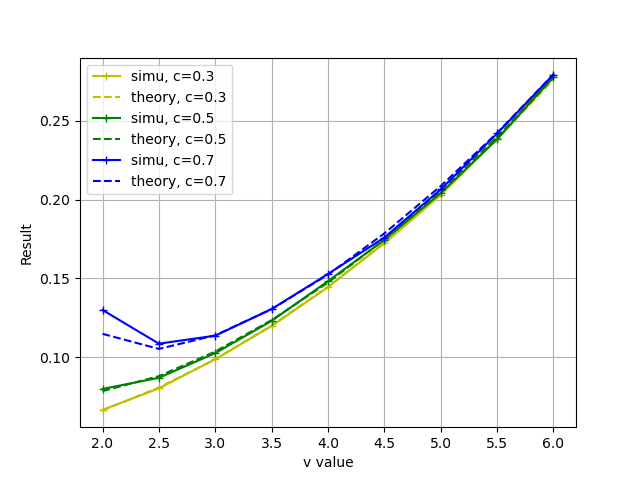}
    \vspace{-0.1in}
    \caption{ICL performance of multi-head attention with prior knowledge $(A_1,A_2,b_1,b_2)=\left((c/v)I_d,((2c-1)/v)I_d,0,0\right)$, $(m,n)=(2,1)$, and $(d,D)=(5,1000)$.}
    \label{fig:simu_pri_multi}
\end{figure}
\end{proof}

\subsection{Noisy Response: Theorem 
\ref{thm:noise}}\label{sec:proof:noisy}

\begin{proof}[Proof of Theorem \ref{thm:noise}]
    The main logic of the proof is the same as Theorem \ref{thm:optimal} for single-head attention and Theorem \ref{thm:multi_head} for multi-head attention.

    \paragraph{Optimal solution for single-head attention}
\begin{eqnarray*}
    &&\mathbb{E}\left(y_q- (W^V_{d+1,:})^{\top}
    E\phi\left(E^{\top}(W^K)^{\top} W^Q \begin{bmatrix}
        x_q\\0
    \end{bmatrix} \right)\right)^2\\
    &=&\mathbb{E}_{(x_q,y_q)}\mathbb{E}_{\{x_i,y_i\}_{i\in[D]}}\left(y_q- v\begin{bmatrix}
        y_1,y_2,\ldots,y_D,0
    \end{bmatrix}\phi\left(\begin{bmatrix}
        x_1^{\top}Ax_q+y_1b^{\top}x_q\\\ldots\\x_q^{\top}A x_q+0
    \end{bmatrix} \right)\right)^2\\
    &=&\mathbb{E}_{(x_q,y_q)}\mathbb{E}_{\{x_i,y_i\}_{i\in[D]}}\bigg(y_q^2+ \underbrace{\left(\frac{ v \sum_{i=1}^D 
    y_i\exp(x_i^{\top}Ax_q+y_ib^{\top}x_q) }{\sum \exp(x_i^{\top}Ax_q+y_ib^{\top}x_q) 
    +\exp(x_q^{\top}Ax_q)}\right)^2 }_{:=A_1}
    -\underbrace{2y_q\left(\frac{v\sum_{i=1}^D 
    y_i\exp(x_i^{\top}Ax_q+y_ib^{\top}x_q) }{\sum \exp(x_i^{\top}Ax_q+y_ib^{\top}x_q) 
    +\exp(x_q^{\top}Ax_q)}\right)}_{:=A_2}
    \bigg),
    \end{eqnarray*}
    where $\mathbb{E}y_q^2=1+\sigma_\epsilon^2$. 
    
    When fixing $x_q$ and $\theta$, the terms $A_1$ becomes
    \begin{eqnarray*}
        &&\mathbb{E}_{\{x_i,y_i\}_{i\in[D]}}A_1\\
        &=&v^2\mathbb{E}_{\{x_i,y_i\}_{i\in[D]}}\frac{ (\sum_{i=1}^D 
    y_i\exp(x_i^{\top}Ax_q+y_ib^{\top}x_q))^2 }{D^2\mathbb{E}^2\exp(x_i^{\top}Ax_q+y_ib^{\top}x_q)}\\
    &&-2v^2\mathbb{E}_{\{x_i,y_i\}_{i\in[D]}}\frac{ (\sum_{i=1}^D 
    y_i\exp(x_i^{\top}Ax_q+y_ib^{\top}x_q))^2 }{D^3\mathbb{E}^3\exp(x_i^{\top}Ax_q+y_ib^{\top}x_q)}\left[\sum_{i=1}^D 
    \exp(x_i^{\top}Ax_q+y_ib^{\top}x_q)-D\mathbb{E}\exp(x_i^{\top}Ax_q+y_ib^{\top}x_q)\right]\\
    &&+3v^2\mathbb{E}_{\{x_i,y_i\}_{i\in[D]}}\frac{ (\sum_{i=1}^D 
    y_i\exp(x_i^{\top}Ax_q+y_ib^{\top}x_q))^2 }{D^4\mathbb{E}^4\exp(x_i^{\top}Ax_q+y_ib^{\top}x_q)}\left[\sum_{i=1}^D 
    \exp(x_i^{\top}Ax_q+y_ib^{\top}x_q)-D\mathbb{E}\exp(x_i^{\top}Ax_q+y_ib^{\top}x_q)\right]^2\\
    &&-2v^2\mathbb{E}_{\{x_i,y_i\}_{i\in[D]}}\frac{ (\sum_{i=1}^D 
    y_i\exp(x_i^{\top}Ax_q+y_ib^{\top}x_q))^2 }{D^3\mathbb{E}^3\exp(x_i^{\top}Ax_q+y_ib^{\top}x_q)}\exp(x_q^{\top}Ax_q)+o\left(\frac{1}{D}\right)\\
    &:=&A_{11}+A_{12}+A_{13}+A_{14}+o\left(\frac{1}{D}\right).
    \end{eqnarray*}
    To figure out $A_{11}$ to $A_{13}$, we know that
    \begin{eqnarray*}
        &&\mathbb{E}y_i^2\exp(x_i^{\top}Ax_q+y_i b^{\top}x_q)^2\\
        &=&\mathbb{E}(x_i^{\top}\theta+\epsilon_i)^2\exp(2x_i^{\top}Ax_q+2x_i^{\top}\theta b^{\top}x_q+2\epsilon_ib^{\top}x_q)\\
        &=&\theta^{\top}\left( I+4(Ax_q+\theta b^{\top}x_q)(Ax_q+\theta b^{\top}x_q)^{\top}\right)\theta\exp(2\|Ax_q+\theta b^{\top}x_q\|^2+2\sigma_{\epsilon}^2(b^{\top}x_q)^2)\\
        &&+8\sigma_{\epsilon} (\theta^{\top}Ax_q)b^{\top}x_q \exp(2\|Ax_q+\theta b^{\top}x_q\|^2+2\sigma_{\epsilon}^2(b^{\top}x_q)^2 )\\
        &&+\sigma_{\epsilon}^2(1+4\sigma_{\epsilon}^2(b^{\top}x_q)^2)\exp(2\|Ax_q+\theta b^{\top}x_q\|^2+2\sigma_{\epsilon}^2(b^{\top}x_q)^2 ),
    \end{eqnarray*}
    and
    \begin{eqnarray*}
        \mathbb{E}y_i\exp(x_i^{\top}Ax_q+y_i b^{\top}x_q)=(\theta^{\top}Ax_q+\|\theta\|^2b^{\top}x_q+\sigma_{\epsilon}(b^{\top}x_q))\exp(\|Ax_q+\theta b^{\top}x_q\|^2/2+\sigma_{\epsilon}^2(b^{\top}x_q)^2 /2).
    \end{eqnarray*}
    As a result,
    \begin{eqnarray*}
        A_{11}&=&v^2\mathbb{E}_{\{x_i,y_i\}_{i\in[D]}}\frac{ (\sum_{i=1}^D 
    y_i\exp(x_i^{\top}Ax_q+y_ib^{\top}x_q))^2 }{D^2\mathbb{E}^2\exp(x_i^{\top}Ax_q+y_ib^{\top}x_q)}\\
    &=&v^2\frac{D\mathbb{E}y_i^2\exp(x_i^{\top}Ax_q+y_i b^{\top}x_q)^2 +D(D-1)\mathbb{E}^2y_i\exp(x_i^{\top}Ax_q+y_i b^{\top}x_q) }{D^2\exp( \|Ax_q+\theta b^{\top}x_q\|^2+\sigma_{\epsilon}^2(b^{\top}x_q)^2 )}\\
    &=&v^2\frac{1}{D}\left[\theta^{\top}\left( I+4(Ax_q+\theta b^{\top}x_q)(Ax_q+\theta b^{\top}x_q)^{\top}\right)\theta+8\sigma_{\epsilon} (\theta^{\top}Ax_q)b^{\top}x_q+\sigma_{\epsilon}^2(1+4\sigma_{\epsilon}^2(b^{\top}x_q)^2)\right]\\
    &&\qquad\qquad\qquad\qquad\qquad\times\exp(\|Ax_q+\theta b^{\top}x_q\|^2+\sigma_{\epsilon}^2(b^{\top}x_q)^2 )\\
    &&+v^2\frac{D-1}{D}(\theta^{\top}Ax_q+\|\theta\|^2b^{\top}x_q+\sigma_{\epsilon}(b^{\top}x_q))^2,
    \end{eqnarray*}
    \begin{eqnarray*}
        &&A_{12}\\
        &=&-2v^2\mathbb{E}_{\{x_i,y_i\}_{i\in[D]}}\frac{ (\sum_{i=1}^D 
    y_i\exp(x_i^{\top}Ax_q+y_ib^{\top}x_q))^2 }{D^3\mathbb{E}^3\exp(x_i^{\top}Ax_q+y_ib^{\top}x_q)}\left[\sum_{i=1}^D 
    \exp(x_i^{\top}Ax_q+y_ib^{\top}x_q)-D\mathbb{E}\exp(x_i^{\top}Ax_q+y_ib^{\top}x_q)\right]\\
    &=&-2v^2\mathbb{E}\frac{2D(D-1)}{D^3\mathbb{E}^3\exp(x_i^{\top}Ax_q+y_ib^{\top}x_q)}\left[ \mathbb{E}y_i\exp(x_i^{\top}Ax_q+y_ib^{\top}x_q)\mathbb{E}y_i\exp(x_i^{\top}Ax_q+y_ib^{\top}x_q)^2\right]\\
    &&+2v^2\mathbb{E}\frac{2D(D-1)}{D^3\mathbb{E}^3\exp(x_i^{\top}Ax_q+y_ib^{\top}x_q)}\left[ \mathbb{E}^2y_i\exp(x_i^{\top}Ax_q+y_ib^{\top}x_q)\mathbb{E}\exp(x_i^{\top}Ax_q+y_ib^{\top}x_q))\right]+o\left(\frac{1}{D}\right)\\
    &=&-4v^2\mathbb{E}\frac{2D(D-1)}{D^3\mathbb{E}^3\exp(x_i^{\top}Ax_q+y_ib^{\top}x_q)}(\theta^{\top}Ax_q+\|\theta\|^2b^{\top}x_q+\sigma_{\epsilon}(b^{\top}x_q))^2\exp(5\|Ax_q+\theta b^{\top}x_q\|^2/2+5\sigma_{\epsilon}^2(b^{\top}x_q)^2 /2)\\
    &&+2v^2\mathbb{E}\frac{2D(D-1)}{D^3\mathbb{E}^3\exp(x_i^{\top}Ax_q+y_ib^{\top}x_q)}(\theta^{\top}Ax_q+\|\theta\|^2b^{\top}x_q+\sigma_{\epsilon}(b^{\top}x_q))^2\exp(3\|Ax_q+\theta b^{\top}x_q\|^2/2+3\sigma_{\epsilon}^2(b^{\top}x_q)^2 /2)\\
    &=&-8v^2\frac{1}{D}(\theta^{\top}Ax_q+\|\theta\|^2b^{\top}x_q+\sigma_{\epsilon}(b^{\top}x_q))^2\exp(\|Ax_q+\theta b^{\top}x_q\|^2+\sigma_{\epsilon}^2(b^{\top}x_q)^2 )\\
    &&+4v^2\frac{1}{D}(\theta^{\top}Ax_q+\|\theta\|^2b^{\top}x_q+\sigma_{\epsilon}(b^{\top}x_q))^2+o\left(\frac{1}{D}\right),
    \end{eqnarray*}
    and
    \begin{eqnarray*}
        &&A_{13}\\
        &=&3v^2\mathbb{E}_{\{x_i,y_i\}_{i\in[D]}}\frac{ (\sum_{i=1}^D 
    y_i\exp(x_i^{\top}Ax_q+y_ib^{\top}x_q))^2 }{D^4\mathbb{E}^4\exp(x_i^{\top}Ax_q+y_ib^{\top}x_q)}\left[\sum_{i=1}^D 
    \exp(x_i^{\top}Ax_q+y_ib^{\top}x_q)-D\mathbb{E}\exp(x_i^{\top}Ax_q+y_ib^{\top}x_q)\right]^2\\
    &=&3v^2\frac{\mathbb{E}^2y_i\exp(x_i^{\top}Ax_q+y_ib^{\top}x_q)}{D\mathbb{E}^4\exp(x_i^{\top}Ax_q+y_ib^{\top}x_q)}\left[\mathbb{E}\exp(x_i^{\top}Ax_q+y_ib^{\top}x_q)^2-\mathbb{E}^2\exp(x_i^{\top}Ax_q+y_ib^{\top}x_q)\right]+o\left(\frac{1}{D}\right)\\
    &=&\frac{3v^2}{D}(\theta^{\top}Ax_q+\|\theta\|^2b^{\top}x_q+\sigma_{\epsilon}(b^{\top}x_q))^2\left( \exp(\|Ax_q+\theta b^{\top}x_q\|^2+\sigma_{\epsilon}^2(b^{\top}x_q)^2 ) -1 \right)+o\left(\frac{1}{D}\right),
    \end{eqnarray*}
    with
    \begin{eqnarray*}
        A_{14}&=&-2v^2\frac{1}{D}(\theta^{\top}Ax_q+\|\theta\|^2b^{\top}x_q+\sigma_{\epsilon}(b^{\top}x_q))^2\exp(x_q^{\top}Ax_q-\|Ax_q+\theta b^{\top}x_q\|^2/2-\sigma_{\epsilon}^2(b^{\top}x_q)^2/2 ).
    \end{eqnarray*}
    In terms of $A_2$, when fixing $x_q$ and $\theta$, we have
    \begin{eqnarray*}
        &&\mathbb{E}_{y_q}\mathbb{E}_{\{x_i,y_i\}_{i\in[D]}}A_2\\
        &=&\mathbb{E}_{y_q}\mathbb{E}_{\{x_i,y_i\}_{i\in[D]}}2y_q\left(\frac{v\sum_{i=1}^D 
    y_i\exp(x_i^{\top}Ax_q+y_ib^{\top}x_q) }{\sum \exp(x_i^{\top}Ax_q+y_ib^{\top}x_q) 
    +\exp(x_q^{\top}Ax_q)}\right)\\
    &=&\mathbb{E}_{\{x_i,y_i\}_{i\in[D]}}2\theta^{\top}x_q\left(\frac{v\sum_{i=1}^D 
    y_i\exp(x_i^{\top}Ax_q+y_ib^{\top}x_q) }{D\mathbb{E}\exp(x_i^{\top}Ax_q+y_ib^{\top}x_q)}\right)\\
    &&-\mathbb{E}_{\{x_i,y_i\}_{i\in[D]}}2\theta^{\top}x_q\left(\frac{v\sum_{i=1}^D 
    y_i\exp(x_i^{\top}Ax_q+y_ib^{\top}x_q) }{D^2\mathbb{E}^2\exp(x_i^{\top}Ax_q+y_ib^{\top}x_q)}\right)\left(\sum_{i=1}^D 
  \exp(x_i^{\top}Ax_q+y_ib^{\top}x_q) - D\mathbb{E}\exp(x_i^{\top}Ax_q+y_ib^{\top}x_q)\right)\\
  &&+\mathbb{E}_{\{x_i,y_i\}_{i\in[D]}}2\theta^{\top}x_q\left(\frac{v\sum_{i=1}^D 
    y_i\exp(x_i^{\top}Ax_q+y_ib^{\top}x_q) }{D^3\mathbb{E}^3\exp(x_i^{\top}Ax_q+y_ib^{\top}x_q)}\right)\left(\sum_{i=1}^D 
  \exp(x_i^{\top}Ax_q+y_ib^{\top}x_q) - D\mathbb{E}\exp(x_i^{\top}Ax_q+y_ib^{\top}x_q)\right)^2\\
  &&-\mathbb{E}_{\{x_i,y_i\}_{i\in[D]}}2\theta^{\top}x_q\left(\frac{v\sum_{i=1}^D 
    y_i\exp(x_i^{\top}Ax_q+y_ib^{\top}x_q) }{D^2\mathbb{E}^2\exp(x_i^{\top}Ax_q+y_ib^{\top}x_q)}\right)\exp(x_q^{\top}Ax_q)+o\left(\frac{1}{D}\right)\\
&:=&A_{21}+A_{22}+A_{23}+A_{24}.
    \end{eqnarray*}
For $A_{21}$ to $A_{24}$, we have
\begin{eqnarray*}
    A_{21}=2v\theta^{\top}x_q(\theta^{\top}Ax_q+\|\theta\|^2b^{\top}x_q+\sigma_{\epsilon}(b^{\top}x_q)),
\end{eqnarray*}
\begin{eqnarray*}
    A_{22}=-2v\theta^{\top}x_q \frac{1}{D}(\theta^{\top}Ax_q+\|\theta\|^2b^{\top}x_q+\sigma_{\epsilon}(b^{\top}x_q))\left( 2\exp(\|Ax_q+\theta b^{\top}x_q\|^2+\sigma_{\epsilon}^2(b^{\top}x_q)^2 ) -1 \right)+o\left(\frac{1}{D}\right),
\end{eqnarray*}
\begin{eqnarray*}
    A_{23}=2v\theta^{\top}x_q\frac{1}{D}(\theta^{\top}Ax_q+\|\theta\|^2b^{\top}x_q+\sigma_{\epsilon}(b^{\top}x_q))\left( \exp(\|Ax_q+\theta b^{\top}x_q\|^2+\sigma_{\epsilon}^2(b^{\top}x_q)^2 ) -1 \right)+o\left(\frac{1}{D}\right),
\end{eqnarray*}
\begin{eqnarray*}
    A_{24}=-2v\theta^{\top}x_q\frac{1}{D}(\theta^{\top}Ax_q+\|\theta\|^2b^{\top}x_q+\sigma_{\epsilon}(b^{\top}x_q))\exp(x_q^{\top}Ax_q-\|Ax_q+\theta b^{\top}x_q\|^2/2-\sigma_{\epsilon}^2(b^{\top}x_q)^2/2 ).
\end{eqnarray*}
Inserting $A_{11}$ to $A_{24}$ into $A_1$ and $A_2$, we obtain
\begin{eqnarray}
    &&\mathbb{E}\left(y_q- (W^V_{d+1,:})^{\top}
    E\phi\left(E^{\top}(W^K)^{\top} W^Q \begin{bmatrix}
        x_q\\0
    \end{bmatrix} \right)\right)^2\nonumber\\
    &=&\mathbb{E}\left(x_q^{\top}\theta-v(\theta^{\top}Ax_q+\|\theta\|^2b^{\top}x_q+\sigma_{\epsilon}(b^{\top}x_q))\right)^2+O\left(\frac{1}{D}\right)\nonumber\\
    &=&\frac{1}{d}tr\left((I-vA)^2\right) + \mathbb{E}\|\theta\|^4\|b\|^2+\sigma_{\epsilon}^2\|b\|^2+O\left(\frac{1}{D}\right).\label{eqn:1/D}
\end{eqnarray}
As a result, the optimal solution of $A$ and $b$ satisfies that $tr\left((I-vA)^2\right)/d=O(1/D)$ and $\|b\|^2=O(1/D)$.  

In addition, similar to Theorem \ref{thm:optimal}, when taking $A=I_d/v$ and $b=0$, we have
\begin{eqnarray*}
A_{11}&=&\frac{1}{D}\mathbb{E}\left[ v^2\|\theta\|^2+4(x_q^{\top}\theta)^2+v^2\sigma_{\epsilon}^2 \right]\exp(\|x_q\|^2/v^2)+\frac{D-1}{D},\\
A_{12}&=&-\frac{8}{D}\mathbb{E}(x_q^{\top}\theta)^2\exp(\|x_q\|^2/v^2)+\frac{4}{D},\\
A_{13}&=&-\frac{3}{D}+\frac{3}{D}\mathbb{E}(x_q^{\top}\theta)^2\exp(\|x_q\|^2/v^2),\\
A_{14}&=&-\frac{2}{D}\mathbb{E}(x_q^{\top}\theta)^2\exp(\|x_q\|^2/v-\|x_q\|^2/v^2/2),\\
A_{21}&=&2,\\
A_{22}&=&-\frac{2}{D}\mathbb{E}(x_q^{\top}\theta)^2(2\exp(\|x_q\|^2/v^2)-1),\\
A_{23}&=&\frac{2}{D}\mathbb{E}(x_q^{\top}\theta)^2(\exp(\|x_q\|^2/v^2)-1),\\
A_{24}&=&-\frac{2}{D}\mathbb{E}(x_q^{\top}\theta)^2 \exp(\|x_q\|^2/v-\|x_q\|^2/v^2/2).
\end{eqnarray*}
As a result,
\begin{eqnarray*}
    &&\mathbb{E}\left(y_q- (W^V_{d+1,:})^{\top}
    E\phi\left(E^{\top}(W^K)^{\top} W^Q \begin{bmatrix}
        x_q\\0
    \end{bmatrix} \right)\right)^2\\
    &=&1+\sigma_{\epsilon}^2+A_{11}+A_{12}+A_{13}+A_{14}-A_{21}-A_{22}-A_{23}-A_{24}\\
    &=&\sigma_{\epsilon}^2+\frac{v^2(1+\sigma_{\epsilon}^2)}{D}\mathbb{E}\exp(\|x_q\|^2/v^2) +\frac{1}{D}\mathbb{E}(x_q^{\top}\theta)^2\exp(\|x_q\|^2/v^2)+o(\frac{1}{D}).
\end{eqnarray*}

    \paragraph{ICL performance of multi-head attention}
\begin{eqnarray*}
 &&\mathbb{E}\left(y_q- f(E)_{d+1,D+1} 
 \right)^2\\
&=&\mathbb{E}\left(y_q- vm \begin{bmatrix}
        y_1,y_2,\ldots,y_D,0
    \end{bmatrix}\phi\left(E^{\top}(W_1^K)^{\top} W_1^Q \begin{bmatrix}
        x_q\\0
    \end{bmatrix} \right) + vn \begin{bmatrix}
        y_1,y_2,\ldots,y_D,0
    \end{bmatrix}\phi\left(E^{\top}(W_2^K)^{\top} W_2^Q \begin{bmatrix}
        x_q\\0
    \end{bmatrix} \right)\right)^2\\
    &=& \mathbb{E}\left(y_q-\frac{vm \sum_{i=1}^D 
   y_i\exp(x_i^{\top}(A_1+\theta b_1^{\top})x_q) }{\sum \exp(x_i^{\top}A_1x_q+y_ib_1^{\top}x_q) 
    +\exp(x_q^{\top}A_1x_q)}+\frac{vn \sum_{i=1}^D 
    y_i\exp(x_i^{\top}(A_2+\theta b_2^{\top})x_q) }{\sum \exp(x_i^{\top}A_2x_q+y_ib_2^{\top}x_q) 
    +\exp(x_q^{\top}A_2x_q)}\right)^2\\
    &=& \mathbb{E}\left(y_q^2+ \left(\frac{vm\sum_{i=1}^D 
    y_i\exp(x_i^{\top}A_1x_q+y_ib_1^{\top}x_q) }{\sum \exp(x_i^{\top}A_1x_q+y_ib_1^{\top}x_q) 
    +\exp(x_q^{\top}A_1x_q)}\right)^2 
    -2y_q\left(\frac{vm\sum_{i=1}^D 
    \theta^{\top} x_i\exp(x_i^{\top}A_1x_q+y_ib_1^{\top}x_q) }{\sum \exp(x_i^{\top}A_1x_q+y_ib_1^{\top}x_q) 
    +\exp(x_q^{\top}A_1x_q)}\right)
    \right) \\
     &+& \mathbb{E}\left(\left(\frac{vn\sum_{i=1}^D
    y_i\exp(x_i^{\top}A_2x_q+y_ib_2^{\top}x_q) }{\sum \exp(x_i^{\top}A_2x_q+y_ib_2^{\top}x_q) 
    +\exp(x_q^{\top}A_2x_q)}\right)^2 
    +2y_q\left(\frac{vn\sum_{i=1}^D 
    y_i\exp(x_i^{\top}A_2x_q+y_ib_2^{\top}x_q) }{\sum \exp(x_i^{\top}A_2x_q+y_ib_2^{\top}x_q) 
    +\exp(x_q^{\top}A_2x_q)}\right)
    \right) \\  
 &-& \mathbb{E}\left(\frac{2vn\sum_{i=1}^D 
    y_i\exp(x_i^{\top}A_1x_q+y_ib_1^{\top}x_q) }{\sum \exp(x_i^{\top}A_1x_q+y_ib_1^{\top}x_q) 
    +\exp(x_q^{\top}A_1x_q)} \frac{vm\sum_{i=1}^D 
    y_i\exp(x_i^{\top}A_2x_q+y_ib_2^{\top}x_q) }{\sum \exp(x_i^{\top}A_2x_q+y_ib_2^{\top}x_q) 
    +\exp(x_q^{\top}A_2x_q)}
    \right).
        \end{eqnarray*}
    When taking $m=2$, $n=1$, $A_1=\frac{c}{v}I$, $A_2=\frac{2c-1}{v}I$ and $b_1=b_2=0$, it becomes
\begin{eqnarray*}
 &&\mathbb{E}\left(y_q- f(E)_{d+1,D+1} 
 \right)^2\\
    &=&1+\sigma_{\epsilon}^2+ \mathbb{E}\left( \left(\frac{2v\sum_{i=1}^D 
    y_i\exp(x_i^{\top}x_q(c/v))}{\sum \exp(x_i^{\top}x_q(c/v)) 
    +\exp(\|x_q\|^2(c/v))}\right)^2 -2y_q\left(\frac{2v\sum_{i=1}^D 
    \theta^{\top} x_i\exp(x_i^{\top}x_q(c/v)) }{\sum \exp(x_i^{\top}x_q(c/v)) 
    +\exp(\|x_q\|^2(c/v))}\right)
    \right) \\
     &&+ \mathbb{E}\left(\left(\frac{v\sum_{i=1}^D
    y_i\exp(x_i^{\top}x_q(2c-1)/v) }{\sum \exp(x_i^{\top}x_q(2c-1)/v) 
    +\exp(\|x_q\|^2(2c-1)/v)}\right)^2 
    +2y_q\left(\frac{v\sum_{i=1}^D 
    y_i\exp(x_i^{\top}x_q(2c-1)/v) }{\sum \exp(x_i^{\top}x_q(2c-1)/v) 
    +\exp(\|x_q\|^2(2c-1)/v)}\right)
    \right) \\  
 &&- \mathbb{E}\left(\frac{2v\sum_{i=1}^D 
    y_i\exp(x_i^{\top}x_q(c/v)) }{\sum \exp(x_i^{\top}x_q(c/v)) 
    +\exp(\|x_q\|^2(c/v))} \frac{2v\sum_{i=1}^D 
    y_i\exp(x_i^{\top}x_q(2c-1)/v) }{\sum \exp(x_i^{\top}x_q(2c-1)/v) 
    +\exp(\|x_q\|^2(2c-1)/v)}
    \right)\\
    &:=&1+\sigma_{\epsilon}^2+B_1+B_2+B_3.
    \end{eqnarray*}
    Similar to how we calculate $A_1$ and $A_2$, for $B_1$, the terms are similar. We follow the above proof and obtain
\begin{eqnarray*}
A_{11}&=&\frac{4}{D}\mathbb{E}\left[ v^2\|\theta\|^2+4c^2(x_q^{\top}\theta)^2+v^2\sigma_{\epsilon}^2 \right]\exp(c^2\|x_q\|^2/v^2)+\frac{D-1}{D}4c^2,\\
A_{12}&=&-\frac{32c^2}{D}\mathbb{E}(x_q^{\top}\theta)^2\exp(c^2\|x_q\|^2/v^2)+\frac{16c^2}{D},\\
A_{13}&=&-\frac{12c^2}{D}+\frac{12c^2}{D}\mathbb{E}(x_q^{\top}\theta)^2\exp(c^2\|x_q\|^2/v^2),\\
A_{14}&=&-\frac{8c^2}{D}\mathbb{E}(x_q^{\top}\theta)^2\exp(c\|x_q\|^2/v-c^2\|x_q\|^2/v^2/2),\\
A_{21}&=&4c,\\
A_{22}&=&-\frac{4c}{D}\mathbb{E}(x_q^{\top}\theta)^2(2\exp(c^2\|x_q\|^2/v^2)-1),\\
A_{23}&=&\frac{4c}{D}\mathbb{E}(x_q^{\top}\theta)^2(\exp(c^2\|x_q\|^2/v^2)-1),\\
A_{24}&=&-\frac{4c}{D}\mathbb{E}(x_q^{\top}\theta)^2 \exp(c\|x_q\|^2/v-c^2\|x_q\|^2/v^2/2),
\end{eqnarray*}
    thus
    \begin{eqnarray*}
        B_1
        &=&\mathbb{E}\left( \left(\frac{2v\sum_{i=1}^D 
    y_i\exp(x_i^{\top}x_q(c/v))}{\sum \exp(x_i^{\top}x_q(c/v)) 
    +\exp(\|x_q\|^2(c/v))}\right)^2 
    -2y_q\left(\frac{2v\sum_{i=1}^D 
    \theta^{\top} x_i\exp(x_i^{\top}x_q(c/v)) }{\sum \exp(x_i^{\top}x_q(c/v)) 
    +\exp(\|x_q\|^2(c/v))}\right)
    \right)\\
    &=&4c^2-4c+\frac{4v^2(1+\sigma_{\epsilon}^2)}{D}\mathbb{E}\exp(c^2\|x_q\|^2/v^2)-\frac{4c^2-4c}{D}\mathbb{E}(x_q^{\top}\theta)^2\exp(c^2\|x_q\|^2/v^2)\\
    &&-\frac{2(4c^2-2c)}{D}\mathbb{E}(x_q^{\top}\theta)^2 \exp(c\|x_q\|^2/v-c^2\|x_q\|^2/v^2/2).
    \end{eqnarray*}
For $B_2$, similarly, we obtain
\begin{eqnarray*}
            B_2
    &=&(2c-1)^2+2(2c-1)+\frac{v^2(1+\sigma_{\epsilon}^2)}{D}\mathbb{E}\exp((2c-1)^2\|x_q\|^2/v^2)\\
    &&-\frac{(2c-1)^2+2(2c-1)}{D}\mathbb{E}(x_q^{\top}\theta)^2\exp((2c-1)^2\|x_q\|^2/v^2)\\
    &&-\frac{2((2c-1)^2+(2c-1))}{D}\mathbb{E}(x_q^{\top}\theta)^2 \exp((2c-1)\|x_q\|^2/v-(2c-1)(2c-1)^2\|x_q\|^2/v^2/2).
\end{eqnarray*}
In terms of $B_3$, 
\begin{eqnarray*}
    &&B_3\\
    &=&- \mathbb{E}\left(\frac{2v\sum_{i=1}^D 
    y_i\exp(x_i^{\top}x_q(c/v)) }{\sum \exp(x_i^{\top}x_q(c/v)) 
    +\exp(\|x_q\|^2(c/v))} \frac{2v\sum_{i=1}^D 
    y_i\exp(x_i^{\top}x_q(2c-1)/v) }{\sum \exp(x_i^{\top}x_q(2c-1)/v) 
    +\exp(\|x_q\|^2(2c-1)/v)}
    \right)\\
    &=&-4v^2\mathbb{E}\frac{\left(\sum_{i=1}^D 
    y_i\exp(x_i^{\top}x_q(c/v))\right)\left(\sum_{i=1}^D 
    y_i\exp(x_i^{\top}x_q(2c-1)/v)\right)}{D^2\mathbb{E}\exp(x_i^{\top}x_q(c/v))\mathbb{E}\exp(x_i^{\top}x_q(2c-1)/v)}\\
    &&+4v^2\mathbb{E}\frac{\left(\sum_{i=1}^D 
    y_i\exp(x_i^{\top}x_q(c/v))\right)\left(\sum_{i=1}^D 
    y_i\exp(x_i^{\top}x_q(2c-1)/v)\right)}{D^3\mathbb{E}^2\exp(x_i^{\top}x_q(c/v))\mathbb{E}\exp(x_i^{\top}x_q(2c-1)/v)}\left( \sum_{i=1}^D  \exp(x_i^{\top}x_q(c/v))-D\mathbb{E}\exp(x_i^{\top}x_q(c/v))\right)\\
    &&+4v^2\mathbb{E}\frac{\left(\sum_{i=1}^D 
    y_i\exp(x_i^{\top}x_q(c/v))\right)\left(\sum_{i=1}^D 
    y_i\exp(x_i^{\top}x_q(2c-1)/v)\right)}{D^3\mathbb{E}\exp(x_i^{\top}x_q(c/v))\mathbb{E}^2\exp(x_i^{\top}x_q(2c-1)/v)}\left( \sum_{i=1}^D  \exp(x_i^{\top}x_q(2c-1)/v)-D\mathbb{E}\exp(x_i^{\top}x_q(2c-1)/v)\right)\\
    &&-4v^2\mathbb{E}\frac{\left(\sum_{i=1}^D 
    y_i\exp(x_i^{\top}x_q(c/v))\right)\left(\sum_{i=1}^D 
    y_i\exp(x_i^{\top}x_q(2c-1)/v)\right)}{D^4\mathbb{E}^3\exp(x_i^{\top}x_q(c/v))\mathbb{E}\exp(x_i^{\top}x_q(2c-1)/v)}\left( \sum_{i=1}^D  \exp(x_i^{\top}x_q(c/v))-D\mathbb{E}\exp(x_i^{\top}x_q(c/v))\right)^2\\
    &&-4v^2\mathbb{E}\frac{\left(\sum_{i=1}^D 
    y_i\exp(x_i^{\top}x_q(c/v))\right)\left(\sum_{i=1}^D 
    y_i\exp(x_i^{\top}x_q(2c-1)/v)\right)}{D^4\mathbb{E}\exp(x_i^{\top}x_q(c/v))\mathbb{E}^3\exp(x_i^{\top}x_q(2c-1)/v)}\\
    &&\qquad\qquad\times\left( \sum_{i=1}^D  \exp(x_i^{\top}x_q(2c-1)/v)-D\mathbb{E}\exp(x_i^{\top}x_q(2c-1)/v)\right)^2\\
    &&-4v^2\mathbb{E}\frac{\left(\sum_{i=1}^D 
    y_i\exp(x_i^{\top}x_q(c/v))\right)\left(\sum_{i=1}^D 
    y_i\exp(x_i^{\top}x_q(2c-1)/v)\right)}{D^4\mathbb{E}^2\exp(x_i^{\top}x_q(c/v))\mathbb{E}^2\exp(x_i^{\top}x_q(2c-1)/v)}\\
    &&\qquad\qquad\times \left( \sum_{i=1}^D  \exp(x_i^{\top}x_q(c/v))-D\mathbb{E}\exp(x_i^{\top}x_q(c/v))\right)\left( \sum_{i=1}^D  \exp(x_i^{\top}x_q(2c-1)/v)-D\mathbb{E}\exp(x_i^{\top}x_q(2c-1)/v)\right)\\
    &&+4v^2\mathbb{E}\frac{\left(\sum_{i=1}^D 
    y_i\exp(x_i^{\top}x_q(c/v))\right)\left(\sum_{i=1}^D 
    y_i\exp(x_i^{\top}x_q(2c-1)/v)\right)}{D^3\mathbb{E}^2\exp(x_i^{\top}x_q(c/v))\mathbb{E}\exp(x_i^{\top}x_q(2c-1)/v)}\exp(\|x_q\|^2c/v)\\
    &&+4v^2\mathbb{E}\frac{\left(\sum_{i=1}^D 
    y_i\exp(x_i^{\top}x_q(c/v))\right)\left(\sum_{i=1}^D 
    y_i\exp(x_i^{\top}x_q(2c-1)/v)\right)}{D^3\mathbb{E}\exp(x_i^{\top}x_q(c/v))\mathbb{E}^2\exp(x_i^{\top}x_q(2c-1)/v)}\exp(\|x_q\|^2(2c-1)/v)+o\left(\frac{1}{D}\right)\\
    &:=&\mathbb{E}(B_{31}+B_{32}+B_{33}+B_{34}+B_{35}+B_{36}+B_{37}+B_{38})+o\left(\frac{1}{D}\right).
\end{eqnarray*}
For $B_{31}$ to $B_{38}$, we have
\begin{eqnarray*}
    B_{31}&=&-4c(2c-1)\frac{D-1}{D}(x_q^{\top}\theta)^2 -4v^2\frac{1}{D}\frac{\left[\theta^{\top}(I_d+ (3c-1)^2 x_qx_q^{\top}/v^2 )\theta+\sigma_{\epsilon}^2  \right]\exp( (3c-1)^2\|x_q\|^2/(2v^2) )}{\exp( (c^2+(2c-1)^2)\|x_q\|^2/(2v^2) )}\\
    &=&-4c(2c-1)\frac{D-1}{D}(x_q^{\top}\theta)^2 -4v^2\frac{1}{D}\left[\theta^{\top}(I_d+ (3c-1)^2 x_qx_q^{\top}/v^2 )\theta+\sigma_{\epsilon}^2  \right]\exp( (2c^2-c)\|x_q\|^2/v^2 ),
\end{eqnarray*}
\begin{eqnarray*}
    B_{32}&=&4\frac{1}{D}(x_q^{\top}\theta)^2\frac{ 2c(2c-1)\exp( (4c^2+(2c-1)^2)\|x_q\|^2/(2v^2) )-c(2c-1)\exp( (2c^2+(2c-1)^2)\|x_q\|^2/(2v^2) ) }{\exp( (2c^2+(2c-1)^2)\|x_q\|^2/(2v^2) )}\\
    &&+4\frac{1}{D}(x_q^{\top}\theta)^2\frac{c(3c-1)\exp( (c^2+(3c-1)^2)\|x_q\|^2/(2v^2) )-c(2c-1)\exp( (c^2+2(2c-1)^2)\|x_q\|^2/(2v^2) )}{\exp( (2c^2+(2c-1)^2)\|x_q\|^2/(2v^2) )}
    +o\left(\frac{1}{D}\right)\\
    &=&\frac{4}{D}(x_q^{\top}\theta)^2[2c(2c-1)]\exp(\|x_q\|^2c^2/v^2)-\frac{8}{D}(x_q^{\top}\theta)^2[c(2c-1)]\\
    &&+\frac{4}{D}(x_q^{\top}\theta)^2[c(3c-1)]\exp((2c^2-c)\|x_q\|^2/v^2)
    +o\left(\frac{1}{D}\right),
\end{eqnarray*}
\begin{eqnarray*}
    B_{33}&=&4\frac{1}{D}(x_q^{\top}\theta)^2\frac{ 2c(2c-1)\exp( (c^2+4(2c-1)^2)\|x_q\|^2/(2v^2) )-c(2c-1)\exp( (c^2+2(2c-1)^2)\|x_q\|^2/(2v^2) ) }{\exp( (c^2+2(2c-1)^2)\|x_q\|^2/(2v^2) )}\\
    &&+4\frac{1}{D}(x_q^{\top}\theta)^2\frac{(3c-1)(2c-1)\exp( ((3c-1)^2+(2c-1)^2)\|x_q\|^2/(2v^2) )-c(2c-1)\exp( (c^2+2(2c-1)^2)\|x_q\|^2/(2v^2) )}{\exp( (c^2+2(2c-1)^2)\|x_q\|^2/(2v^2) )}\\
    &&+o\left(\frac{1}{D}\right)\\
    &=&\frac{4}{D}(x_q^{\top}\theta)^2[2c(2c-1)]\exp(\|x_q\|^2(2c-1)^2/v^2)-\frac{8v^2}{D}(x_q^{\top}\theta)^2[c(2c-1)]\\
    &&+\frac{4}{D}(x_q^{\top}\theta)^2[(3c-1)(2c-1)]\exp((2c^2-c)\|x_q\|^2/v^2)
    +o\left(\frac{1}{D}\right),
\end{eqnarray*}
\begin{eqnarray*}
    B_{34}=-4\frac{c(2c-1)}{D}(x_q^{\top}\theta)^2\left(\exp(c^2\|x_q\|^2/v^2)-1\right)+o\left(\frac{1}{D}\right),
\end{eqnarray*}
\begin{eqnarray*}
    B_{35}=-4\frac{c(2c-1)}{D}(x_q^{\top}\theta)^2\left(\exp((2c-1)^2\|x_q\|^2/v^2)-1\right)+o\left(\frac{1}{D}\right),
\end{eqnarray*}
\begin{eqnarray*}
    B_{36}&=&-4\frac{c(2c-1)}{D}(x_q^{\top}\theta)^2\left(\frac{\exp((3c-1)^2\|x_q\|^2/(2v^2))}{\exp(c^2\|x_q\|^2/(2v^2))\exp((2c-1)^2\|x_q\|^2/(2v^2))}-1\right)+o\left(\frac{1}{D}\right)\\
    &=&-4\frac{c(2c-1)}{D}(x_q^{\top}\theta)^2\left(\exp((2c^2-c)\|x_q\|^2/v^2)-1\right)+o\left(\frac{1}{D}\right),
\end{eqnarray*}
and
\begin{eqnarray*}
    B_{37}=\frac{4}{D}c(2c-1)\exp(\|x_q\|^2c/v-\|x_q\|^2c^2/(2v^2))+o\left(\frac{1}{D}\right),
\end{eqnarray*}
\begin{eqnarray*}
    B_{38}=\frac{4}{D}c(2c-1)\exp(\|x_q\|^2(2c-1)/v-\|x_q\|^2(2c-1)^2/(2v^2))+o\left(\frac{1}{D}\right).
\end{eqnarray*}
Putting everything together, we have
\begin{eqnarray*}
    B_3
    &=&B_{31}+B_{32}+B_{33}+B_{34}+B_{35}+B_{36}+B_{37}+B_{38}\\
    &=&-4c(2c-1)\frac{D-1}{D}(x_q^{\top}\theta)^2 -4v^2\frac{1}{D}\left[\theta^{\top}(I_d+ (3c-1)^2 x_qx_q^{\top}/v^2 )\theta+\sigma_{\epsilon}^2  \right]\exp( (2c^2-c)\|x_q\|^2/v^2 )\\
    &&+\frac{4}{D}(x_q^{\top}\theta)^2[2c(2c-1)]\exp(\|x_q\|^2c^2/v^2)-\frac{8}{D}(x_q^{\top}\theta)^2[c(2c-1)]\\
    &&+\frac{4}{D}(x_q^{\top}\theta)^2[c(3c-1)]\exp((2c^2-c)\|x_q\|^2/v^2)\\
    &&+\frac{4}{D}(x_q^{\top}\theta)^2[2c(2c-1)]\exp(\|x_q\|^2(2c-1)^2/v^2)-\frac{8}{D}(x_q^{\top}\theta)^2[c(2c-1)]\\
    &&+\frac{4}{D}(x_q^{\top}\theta)^2[(3c-1)(2c-1)]\exp((2c^2-c)\|x_q\|^2/v^2)\\
    &&-4\frac{c(2c-1)}{D}(x_q^{\top}\theta)^2\left(\exp(c^2\|x_q\|^2/v^2)-1\right)\\
    &&-4\frac{c(2c-1)}{D}(x_q^{\top}\theta)^2\left(\exp((2c-1)^2\|x_q\|^2/v^2)-1\right)\\
    &&-4\frac{c(2c-1)}{D}(x_q^{\top}\theta)^2\left(\exp((2c^2-c)\|x_q\|^2/v^2)-1\right)\\
    &&+\frac{4}{D}c(2c-1)\exp(\|x_q\|^2c/v-\|x_q\|^2c^2/(2v^2))\\
    &&+\frac{4}{D}c(2c-1)\exp(\|x_q\|^2(2c-1)/v-\|x_q\|^2(2c-1)^2/(2v^2))+o\left(\frac{1}{D}\right)\\
    &=&-(8c^2-4c)(x_q^{\top}\theta)^2-\frac{6}{D}(x_q^{\top}\theta)^2c(2c-1)-\frac{4v^2}{D}\left[\|\theta\|^2+\sigma_{\epsilon}^2   \right]\exp((2c^2-c)\|x_q\|^2/v^2)\\
    &&+\frac{4}{D}(x_q^{\top}\theta)^2\exp(\|x_q\|^2c^2/v^2)\left[2c(2c-1)-c(2c-1)\right]\\
    &&+\frac{4}{D}(x_q^{\top}\theta)^2\exp(\|x_q\|^2(2c-1)^2/v^2)\left[2c(2c-1)-c(2c-1)\right]\\
    &&+\frac{4}{D}(x_q^{\top}\theta)^2\exp(\|x_q\|^2(2c^2-c)/v^2)\left[ -2c(2c-1) \right]\\
    &&+\frac{4}{D}c(2c-1)\exp(\|x_q\|^2c/v-\|x_q\|^2c^2/(2v^2))\\
    &&+\frac{4}{D}c(2c-1)\exp(\|x_q\|^2(2c-1)/v-\|x_q\|^2(2c-1)^2/(2v^2))+o\left(\frac{1}{D}\right)\\
    &=&-4(2c^2-c)(x_q^{\top}\theta)^2-\frac{4v^2}{D}\left[\|\theta\|^2+\sigma_{\epsilon}^2    \right]\exp((2c^2-c)\|x_q\|^2/v^2)\\
    &&+\frac{4}{D}(x_q^{\top}\theta)^2\exp(\|x_q\|^2c^2/v^2)\left[c(2c-1)\right]\\
    &&+\frac{4}{D}(x_q^{\top}\theta)^2\exp(\|x_q\|^2(2c-1)^2/v^2)\left[c(2c-1)\right]\\
    &&+\frac{4}{D}(x_q^{\top}\theta)^2\exp(\|x_q\|^2(2c^2-c)/v^2)\left[-2c(2c-1) \right]\\
    &&+\frac{4}{D}c(2c-1)\exp(\|x_q\|^2c/v-\|x_q\|^2c^2/(2v^2))\\
    &&+\frac{4}{D}c(2c-1)\exp(\|x_q\|^2(2c-1)/v-\|x_q\|^2(2c-1)^2/(2v^2))+o\left(\frac{1}{D}\right).
\end{eqnarray*}
Finally,
\begin{eqnarray*}
 &&\mathbb{E}\left(y_q- f(E)_{d+1,D+1} 
 \right)^2\\
    &=&1+\sigma_{\epsilon}^2+B_1+B_2+\mathbb{E}(B_{31}+B_{32}+B_{33}+B_{34}+B_{35}+B_{36}+B_{37}+B_{38})+o\left(\frac{1}{D}\right)\\
    &=&1+\sigma_{\epsilon}^2+4c^2-4c+\frac{4v^2(1+\sigma_{\epsilon}^2)}{D}\mathbb{E}\exp(c^2\|x_q\|^2/v^2)-\frac{4c^2-4c}{D}\mathbb{E}(x_q^{\top}\theta)^2\exp(c^2\|x_q\|^2/v^2)\\
    &&-\frac{2(4c^2-2c)}{D}\mathbb{E}(x_q^{\top}\theta)^2 \exp(c\|x_q\|^2/v-c^2\|x_q\|^2/v^2/2)\\
    &&+(2c-1)^2+2(2c-1)+\frac{v^2(1+\sigma_{\epsilon}^2)}{D}\mathbb{E}\exp((2c-1)^2\|x_q\|^2/v^2)\\
    &&-\frac{(2c-1)^2+2(2c-1)}{D}\mathbb{E}(x_q^{\top}\theta)^2\exp((2c-1)^2\|x_q\|^2/v^2)\\
    &&-\frac{2((2c-1)^2+(2c-1))}{D}\mathbb{E}(x_q^{\top}\theta)^2 \exp((2c-1)\|x_q\|^2/v-(2c-1)(2c-1)^2\|x_q\|^2/v^2/2)\\
    &&-\mathbb{E}4(2c^2-c)(x_q^{\top}\theta)^2-\mathbb{E}\frac{4v^2}{D}\left[\|\theta\|^2+\sigma_{\epsilon}^2    \right]\exp((2c^2-c)\|x_q\|^2/v^2)\\
    &&+\mathbb{E}\frac{4}{D}(x_q^{\top}\theta)^2\exp(\|x_q\|^2c^2/v^2)\left[c(2c-1)\right]\\
    &&+\mathbb{E}\frac{4}{D}(x_q^{\top}\theta)^2\exp(\|x_q\|^2(2c-1)^2/v^2)\left[c(2c-1)\right]\\
    &&+\mathbb{E}\frac{4}{D}(x_q^{\top}\theta)^2\exp(\|x_q\|^2(2c^2-c)/v^2)\left[-2c(2c-1) \right]\\
    &&+\mathbb{E}\frac{4}{D}c(2c-1)\exp(\|x_q\|^2c/v-\|x_q\|^2c^2/(2v^2))\\
    &&+\mathbb{E}\frac{4}{D}c(2c-1)\exp(\|x_q\|^2(2c-1)/v-\|x_q\|^2(2c-1)^2/(2v^2))+o\left(\frac{1}{D}\right)\\
    &=&\sigma_{\epsilon}^2+\frac{v^2(1+\sigma_{\epsilon}^2)}{D}\mathbb{E}\left[4\exp(c^2\|x_q\|^2/v^2)+\exp((2c-1)^2\|x_q\|^2/v^2)-4\exp((2c^2-c)\|x_q\|^2/v^2)\right]\\
    &&+\frac{4c^2}{D}(x_q^{\top}\theta)^2\exp(\|x_q\|^2c^2/v^2)+\frac{(2c-1)^2}{D}(x_q^{\top}\theta)^2\exp(\|x_q\|^2(2c-1)^2/v^2)\\
    &&+\frac{4}{D}(x_q^{\top}\theta)^2\exp(\|x_q\|^2(2c^2-c)/v^2)\left[-2c(2c-1) \right]+o\left(\frac{1}{D}\right),
    \end{eqnarray*}
    thus
    \begin{eqnarray*}
&&\mathbb{E}\left(y_q- f(E)_{d+1,D+1} 
 \right)^2\\
  &=& \sigma_{\epsilon}^2+\frac{4v^2(1+\sigma_{\epsilon}^2)}{D}\left( (\frac{v^2}{v^2-2c^2})^{\frac{d}{2}}- (\frac{v^2}{v^2-2c(2c-1)})^{\frac{d}{2}}\right)+\frac{v^2(1+\sigma_{\epsilon}^2)}{D}(\frac{v^2}{v^2-2(2c-1)^2})^{\frac{d}{2}} \\ &&+  \frac{(2c-1)^2}{D}(\frac{v^2}{v^2-2(2c-1)^2})(\frac{v^2}{v^2-2(2c-1)^2})^{\frac{d}{2}} \\
  &&+  \frac{4c^2}{D}  (\frac{v^2}{v^2-2c^2})  (\frac{v^2}{v^2-2c^2})^{\frac{d}{2}}-\frac{4(2c-1)c}{D}  (\frac{v^2}{v^2-2c(2c-1)}) (\frac{v^2}{v^2-2c(2c-1)})^{\frac{d}{2}}+o(\frac{1}{D}).
 \end{eqnarray*}
\end{proof}
\subsection{Correlated Features: Theorem \ref{prop:correlated}}\label{sec:proof:correlated}

\begin{proof}[Proof of Theorem \ref{prop:correlated}]
To figure out the optimal solution of single-head attention, we firstly transform the problem from correlated features to the problem with isotropic features with a new $\theta$ distribution. After transforming the problem, since Theorem \ref{thm:optimal} only utilize the distribution of $\theta$ in its last derivation step, we can directly utilize the results in Theorem \ref{thm:optimal}.

To transform correlated features, denote $z\sim N(0,I_d)$ and $x=\Sigma^{1/2}z$. Recall that the attention score is calculated as
\begin{eqnarray*}
    \phi\left(  (W^KW_{in}E)^{\top}(W^QW_{in}E) \right)=\phi\left(  (W^KW_{in}E)^{\top}(W^QW_{in}E) \right)
\end{eqnarray*}
    
Based on Theorem \ref{thm:optimal}, we have
    \begin{eqnarray*}
&&\mathbb{E}\left(y_q- (W^V_{d+1,:})^{\top}
    E\phi\left(E^{\top}(W^K)^{\top} W^Q \begin{bmatrix}
        x_q\\0
    \end{bmatrix} \right)\right)^2 \\
&=&\mathbb{E}_{(x_q,\theta)}  (x_q^{\top}\theta)^2+\frac{v^2}{D}\theta^{\top}(I_d-4(A+\theta b^{\top})x_q x_q^{\top}(A+\theta b^{\top})^{\top})\theta \exp(x_q^{\top}(A+\theta b^{\top})^{\top}(A+\theta b^{\top})x_q) \\
  &+&   v^2(1+\frac{3}{D})(\theta^{\top}(A+\theta b^{\top})x_q)^2 - \frac{2v^2(\theta^{\top}(A+\theta b^{\top})x_q)^2\exp{(x_q^{\top}Ax_q)}}{D \exp(x_q^{\top}(A+\theta b^{\top})^{\top}(A+\theta b^{\top})x_q/2)} \\
   &-& \frac{3v^2}{D}(\theta^{\top}(A+\theta b^{\top})x_q)^2 + \frac{3v^2}{D}(\theta^{\top}(A+\theta b^{\top})x_q)^2 \exp(x_q^{\top}(A+\theta b^{\top})^{\top}(A+\theta b^{\top})x_q)\\
  &-& 2(x_q^{\top}\theta)\bigg(v\theta^{\top}(A+\theta b^{\top})x_q - \frac{v\theta^{\top}(A+\theta b^{\top})x_q\exp(x_q^{\top}Ax_q)}{D\exp(x_q^{\top}(A+\theta b^{\top})^{\top}(A+\theta b^{\top})x_q/2)} +  \frac{v}{D}\theta^{\top}(A+\theta b^{\top})x_q \\ 
  &&\qquad-\frac{2v}{D}\theta^{\top}(A+\theta b^{\top})x_q\exp(x_q^{\top}(A+\theta b^{\top})^{\top}(A+\theta b^{\top})x_q -\frac{v}{D}(\theta^{\top}(A+\theta b^{\top})x_q) \\
  &&\qquad  + \frac{v}{D}(\theta^{\top}(A+\theta b^{\top})x_q) \exp(x_q^{\top}(A+\theta b^{\top})^{\top}(A+\theta b^{\top})x_q))\bigg)  \color{black} + o(\frac{1}{D}),
    \end{eqnarray*}
from which the optimal solution satisfies $\mathbb{E}\theta^{\top}(I_d-vA)^2\theta=O(1/D)$ and $\|b\|^2\mathbb{E}\|\theta\|^4=O(1/D)$ where $\theta\sim N(0,\Sigma^{-1/2}/d)$.

For multi-head attention, the same argument applies, and we can also transform the correlated features problem to isotropic features with a new $\theta$ distribution. Further, due to the flexibility of multi-head attention, when each head is of full rank, i.e., $p/h>d$, the performance of multi-head attention is not worse than single-head attention. There always exists some $W_{out}$ such that the multi-head attention can be reduced to a single-head attention.
   
\end{proof}


\subsection{Local Examples: Theorem \ref{thm:local_local} and \ref{thm:local_population}}
\subsubsection{Theorem \ref{thm:local_local}}
\begin{proof}[Proof of Theorem \ref{thm:local_local}]
The proof of Theorem \ref{thm:local_local} is almost the same as Theorem \ref{thm:optimal}. The only difference is the change on the distribution of the examples $(x_i,y_i)$s.

When taking infinite many training prompts, the loss function becomes
    \begin{eqnarray*}
    &&\mathbb{E}\left(y_q- (W^V_{d+1,:})^{\top}
    E\phi\left(E^{\top}(W^K)^{\top} W^Q \begin{bmatrix}
        x_q\\0
    \end{bmatrix} \right)\right)^2\\
   &=&\mathbb{E}_{(x_q,\theta)}\mathbb{E}_{\{x_i\}_{i\in[D]}}\left(y_q^2
    \underbrace{-2y_q\left(\frac{v\sum_{i=1}^D 
    \theta^{\top} x_i\exp(x_i^{\top}Ax_q+y_ib^{\top}x_q) }{\sum \exp(x_i^{\top}Ax_q+y_ib^{\top}x_q)
    +\exp(x_q^{\top}Ax_q)}\right)}_{=A_1}+ \underbrace{\left(\frac{ v \sum_{i=1}^D 
    \theta^{\top} x_i\exp(x_i^{\top}Ax_q+y_ib^{\top}x_q) }{\sum \exp(x_i^{\top}Ax_q+y_ib^{\top}x_q)+\exp(x_q^{\top}Ax_q)}\right)^2}_{=A_2}
    \right).
    \end{eqnarray*}
    For $A_1$, we have
    \begin{eqnarray*}
    &&\mathbb{E}_{\{x_i,y_i\}_{i\in[D]}}A_1\\
    &=&\mathbb{E}_{\{x_i,y_i\}_{i\in[D]}}\frac{(-2v\theta^\top x_q) \sum_{i=1}^D 
    \theta^{\top} x_i\exp(x_i^{\top}Ax_q+y_ib^{\top}x_q) }{D\mathbb{E}\exp({x_1}^{\top}Ax_q+y_1b^{\top}x_q)}+o(\frac{1}{D})\\
    &&+ \mathbb{E}_{\{x_i,y_i\}_{i\in[D]}}\frac{(2v\theta^\top x_q) \sum_{i=1}^D 
    \theta^{\top} x_i\exp(x_i^{\top}Ax_q+y_ib^{\top}x_q) }{(D\mathbb{E}\exp({x_1}^{\top}Ax_q+y_1b^{\top}x_q))^2}\\&&\qquad\qquad\qquad\times \left(\exp(x_q^{\top}Ax_q)+\sum \exp(x_i^{\top}Ax_q+y_ib^{\top}x_q)-D\mathbb{E}\exp({x_1}^{\top}Ax_q+y_1b^{\top}x_q)\right)\\ 
    &&-\mathbb{E}_{\{x_i,y_i\}_{i\in[D]}}\frac{(2v\theta^\top x_q) \sum_{i=1}^D 
    \theta^{\top} x_i\exp(x_i^{\top}Ax_q+y_ib^{\top}x_q) }{D\mathbb{E}\exp({x_1}^{\top}Ax_q+y_1b^{\top}x_q))^3}\left(\sum \exp(x_i^{\top}Ax_q+y_ib^{\top}x_q)-D\mathbb{E}\exp({x_1}^{\top}Ax_q+y_1b^{\top}x_q)\right)^2\\
    &&+o\left(\frac{1}{D}\right)\\
    &=&A_{11}+A_{12}+A_{13}+o(\frac{1}{D}).
    \end{eqnarray*}
    Since $x_i\sim N(x_q,\sigma_x^2)$, we have
    \begin{eqnarray*}
        \mathbb{E}_{\{x_1,y_1\}}\exp(x_1^{\top}Ax_q+y_1b^{\top}x_q)&=&\mathbb{E}_{\{x_1,y_1\}}\exp\left(\left(\sigma_x\frac{x_1-x_q}{\sigma_x}\right)^{\top}(A+\theta b^{\top})x_q\right)\\
        &=&\exp\left(\frac{1}{2}\sigma_x^2 x_q^{\top}(A+\theta b^{\top})^{\top}(A+\theta b^{\top}) x_q+x_q^{\top}(A+\theta b^{\top}) x_q\right),\\
        \mathbb{E}_{\{x_1,y_1\}}x_1\exp(x^{\top}Ax_q+y_1b^{\top}x_q)
        &=&\left[\sigma_x^2(A+\theta b^{\top})x_q+x_q\right]\exp(x_q^{\top}(A+\theta b^{\top})^{\top}(A+\theta b^{\top})x_q/2+x_q^{\top}(A+\theta b^{\top}) x_q).
    \end{eqnarray*}
    Therefore,
    \begin{eqnarray*}
A_{11}
&=&\mathbb{E}_{\{x_1,y_1\}}\left(-\frac{D(2v\theta^\top x_q)\mathbb{E}\theta^{\top}x_1\exp(x_1^{\top}Ax_q+y_1b^{\top}x_q)}{D\mathbb{E}_{x_1}\exp(x_1^{\top}Ax_q+y_1b^{\top}x_q)}\right)\\
    &=&-(2v\theta^\top x_q)\theta^{\top}\left[\sigma_x^2(A+\theta b^{\top})x_q + x_q\right]  +o(\frac{1}{D}).\\
    \end{eqnarray*}
        \begin{eqnarray*}
       A_{12} &=&\mathbb{E}_{\{x_i,y_i\}_{i\in[D]}}\frac{(2v\theta^\top x_q \sum_{i=1}^D 
    \theta^{\top} x_i\exp(x_i^{\top}Ax_q+y_ib^{\top}x_q))(\exp(x_q^{\top}Ax_q)+\sum \exp(x_i^{\top}Ax_q+y_ib^{\top}x_q)) }{(D\mathbb{E}\exp({x_1}^{\top}Ax_q+y_1b^{\top}x_q))^2}\\
    &&-\mathbb{E}_{\{x_i,y_i\}_{i\in[D]}}\frac{(2v\theta^\top x_q\sum_{i=1}^D 
    \theta^{\top} x_i\exp(x_i^{\top}Ax_q+y_ib^{\top}x_q))(D\mathbb{E}\exp({x_1}^{\top}Ax_q+y_1b^{\top}x_q)) }{(D\mathbb{E}\exp({x_1}^{\top}Ax_q+y_1b^{\top}x_q))^2}\\
    &=&\frac{2v(\theta^{\top}y_q)}{D}\theta^{\top}\left[2\sigma_x^2(A+\theta b^{\top})x_q + x_q\right]\exp\left(x_q^{\top}(A+\theta b^{\top})^{\top}(A+\theta b^{\top})x_q\right)\\
    &&-\frac{2v}{D}(\theta^{\top}x_q)\theta^{\top}\left[\sigma_x^2(A+\theta b^{\top})x_q + x_q\right]+\frac{2v(\theta^{\top}\left[\sigma_x^2(A+\theta b^{\top})x_q + x_q\right])^2 \exp{(x_q^{\top}Ax_q)}}{D \exp(x_q^{\top}(A+\theta b^{\top})^{\top}(A+\theta b^{\top})x_q)}.
        \end{eqnarray*}
    
\begin{eqnarray*}
    A_{13}
=\frac{2v}{D}(\theta^\top x_q)\theta^\top\left[\sigma_x^2(A+\theta b^{\top})x_q + x_q\right] - \frac{2v}{D}(\theta^\top x_q)\theta^\top\left[\sigma_x^2(A+\theta b^{\top})x_q + x_q\right] \exp(x_q^{\top}(A+\theta b^{\top})^{\top}(A+\theta b^{\top})x_q). 
\end{eqnarray*}
To sum up, we have 
\begin{eqnarray*}
     A_1&=& A_{11}+ A_{12}+ A_{13} \\
     &=&-2(v\theta^\top x_q)\theta^{\top}\left[\sigma_x^2(A+\theta b^{\top})x_q + x_q\right] + \frac{(2v\theta^\top x_q)\theta^{\top}\left[\sigma_x^2(A+\theta b^{\top})x_q + x_q\right]\exp(x_q^{\top}Ax_q)}{D\exp(x_q^{\top}(A+\theta b^{\top})^{\top}(A+\theta b^{\top})x_q/2)} \\
    &&+ \frac{2v}{D}(\theta^\top x_q)\theta^\top\left[\sigma_x^2(A+\theta b^{\top})x_q \right] \exp(x_q^{\top}(A+\theta b^{\top})^{\top}(A+\theta b^{\top})x_q).
   \end{eqnarray*}  
    In terms of $A_2$, we have
    \begin{eqnarray*}
    &&\mathbb{E}x_ix_i^{\top}\exp(2x_i^{\top}(A+\theta b^{\top})x_q)\\
    &=&\mathbb{E}\left(\frac{x_i-x_q}{\sigma_x}\right)\left(\frac{x_i-x_q}{\sigma_x}\right)^{\top}\sigma_x^2\exp\left({2\sigma_x}\frac{(x_i-x_q)^{\top}(A+\theta b^{\top})x_q}{\sigma_x}+2 x_q^{\top}(A+\theta b^{\top})x_q\right)\\
    &&+x_qx_q^{\top}\exp(\frac{\sigma_x^2}{2v^2}\|x_q\|^2+2\|x_q\|^2/v)\\
    &=&\left(\sigma_x^2\left( I_d+{4\sigma_x^2}(A+\theta b^{\top})x_qx_q^{\top} (A+\theta b^{\top})^{\top}\right)+x_qx_q^{\top}\right)\exp\left({2\sigma_x}\frac{(x_i-x_q)^{\top}(A+\theta b^{\top})x_q}{\sigma_x}+2 x_q^{\top}(A+\theta b^{\top})x_q\right).
\end{eqnarray*}
    since $x_i$s are independent with each other, we have
     \begin{eqnarray*}
         &&\mathbb{E}_{\{x_i\}_{i\in[D]}}A_2\\
    &=& \mathbb{E}_{\{x_i\}_{i\in[D]}}\left(\frac{ v \sum_{i=1}^D 
    \theta^{\top} x_i\exp(x_i^{\top}Ax_q+y_ib^{\top}x_q) }{D \mathbb{E}_{x_1} \exp(x_1^{\top}Ax_q+y_1b^{\top}x_q)
    }\right)^2\\
        && -2\mathbb{E}_{\{x_i\}_{i\in[D]}}\frac{\left(v \sum_{i=1}^D 
    \theta^{\top} x_i\exp(x_i^{\top}Ax_q+y_ib^{\top}x_q) \right)^2}{\left(D \mathbb{E}_{x_1} \exp(x_1^{\top}Ax_q+y_1b^{\top}x_q)
    \right)^3}\left( \sum \exp(x_i^{\top}Ax_q+y_ib^{\top}x_q) 
     - (D \mathbb{E}_{x_1} \exp(x_1^{\top}Ax_q+y_1b^{\top}x_q)
)\right)\\
&& -2\mathbb{E}_{\{x_i\}_{i\in[D]}}\frac{\left(v \sum_{i=1}^D 
    \theta^{\top} x_i\exp(x_i^{\top}Ax_q+y_ib^{\top}x_q) \right)^2}{\left(D \mathbb{E}_{x_1} \exp(x_1^{\top}Ax_q+y_1b^{\top}x_q)
    \right)^3}\exp(x_q^{\top}Ax_q)\\
        &&+ 3\mathbb{E}_{\{x_i\}_{i\in[D]}}\frac{\left(v \sum_{i=1}^D 
    \theta^{\top} x_i\exp(x_i^{\top}Ax_q+y_ib^{\top}x_q) \right)^2}{\left(D \mathbb{E}_{x_1} \exp(x_1^{\top}Ax_q+y_1b^{\top}x_q)
    \right)^4}\left( \sum \exp(x_i^{\top}Ax_q+y_ib^{\top}x_q) 
     - (D \mathbb{E}_{x_1} \exp(x_1^{\top}Ax_q+y_1b^{\top}x_q)
)\right)^2\\
&=& 
\frac{ v^2 D\mathbb{E}(\theta^{\top}x_i)^2\exp(2x_i^{\top}Ax_q+2y_ib^{\top}x_q) + D(D-1)v^2 \mathbb{E}^2(\theta^{\top}x_i)\exp(x_i^{\top}Ax_q +y_i b^{\top}x_q)}{(D \mathbb{E}_{x_1} \exp(x_1^{\top}Ax_q+y_1b^{\top}x_q)
    )^2}\\
&&-\frac{ 4D(D-1)v^2 \mathbb{E}(\theta^{\top}x_i)\exp(x_i^{\top}Ax_q +y_i b^{\top}x_q)\mathbb{E}(\theta^{\top}x_i)\exp(2x_i^{\top}Ax_q +2y_i b^{\top}x_q)}{(D \mathbb{E}_{x_1} \exp(x_1^{\top}Ax_q+y_1b^{\top}x_q)
   )^3}\\
    \\
&&+\frac{ 4D(D-1)v^2 \mathbb{E}(\theta^{\top}x_i)\exp(x_i^{\top}Ax_q +y_i b^{\top}x_q)^2\mathbb{E}\exp(x_i^{\top}Ax_q +y_i b^{\top}x_q)}{(D \mathbb{E}_{x_1} \exp(x_1^{\top}Ax_q+y_1b^{\top}x_q)
    )^3}\\
    \\
&&-\frac{ D(D-1)v^2 \mathbb{E}(\theta^{\top}x_i)\exp(x_i^{\top}Ax_q +y_i b^{\top}x_q)^2\exp(x_q^{\top}Ax_q)}{(D \mathbb{E}_{x_1} \exp(x_1^{\top}Ax_q+y_1b^{\top}x_q)
    )^3}\\
&&+\frac{ 3D^2(D-1)v^2 \mathbb{E}^2(\theta^{\top}x_i)\exp(x_i^{\top}Ax_q +y_i b^{\top}x_q)\left[\mathbb{E}\exp(2x_i^{\top}Ax_q +2y_i b^{\top}x_q)-\mathbb{E}^2\exp(x_i^{\top}Ax_q +y_i b^{\top}x_q)\right]}{(D \mathbb{E}_{x_1} \exp(x_1^{\top}Ax_q+y_1b^{\top}x_q)
    )^4}+o\left(\frac{1}{D}\right)\\
&=& \frac{v^2}{D}\theta^{\top}(\sigma_x^2 I_d+4\sigma_x^4(A+\theta b^{\top})x_q x_q^{\top}(A+\theta b^{\top})^{\top}+x_qx_q^{\top}  )\theta \exp(x_q^{\top}(A+\theta b^{\top})^{\top}(A+\theta b^{\top})x_q)\\
&&+v^2(\theta^{\top}\left[\sigma_x^2(A+\theta b^{\top})x_q + x_q\right])^2 \\
 &&+ \frac{v^2}{D}\theta^{\top}(-4(2\sigma_x^2(A+\theta b^{\top})x_q+x_q)(\sigma_x^2(A+\theta b^{\top})x_q+x_q)^{\top}+3(\sigma_x^2(A+\theta b^{\top})x_q+x_q)(\sigma_x^2(A+\theta b^{\top})x_q+x_q)^{\top} )\theta \\
 &&\qquad\qquad\qquad\qquad\times\exp(x_q^{\top}(A+\theta b^{\top})^{\top}(A+\theta b^{\top})x_q) \\
 &&- \frac{2v^2(\theta^{\top}\left[\sigma_x^2(A+\theta b^{\top})x_q + x_q\right])^2 \exp{(x_q^{\top}Ax_q)}}{D \exp(x_q^{\top}(A+\theta b^{\top})^{\top}(A+\theta b^{\top})x_q)} +o(\frac{1}{D})\\
 &=& \frac{v^2}{D}\theta^{\top}(\sigma_x^2 I_d-\sigma_x^4(A+\theta b^{\top})x_q x_q^{\top}(A+\theta b^{\top})^{\top}  )\theta \exp(x_q^{\top}(A+\theta b^{\top})^{\top}(A+\theta b^{\top})x_q)\\
&&+v^2(\theta^{\top}\left[\sigma_x^2(A+\theta b^{\top})x_q + x_q\right])^2 - \frac{2v^2(\theta^{\top}\left[\sigma_x^2(A+\theta b^{\top})x_q + x_q\right])^2 \exp{(x_q^{\top}Ax_q)}}{D \exp(x_q^{\top}(A+\theta b^{\top})^{\top}(A+\theta b^{\top})x_q)} +o(\frac{1}{D})
\end{eqnarray*} 
Based on the results of $A_1$ and $A_2$, we have 
        
    \begin{eqnarray*}
&&\mathbb{E}\left(y_q- (W^V_{d+1,:})^{\top}
    E\phi\left(E^{\top}(W^K)^{\top} W^Q \begin{bmatrix}
        x_q\\0
    \end{bmatrix} \right)\right)^2 \nonumber \\
&=&\mathbb{E}_{(x_q,\theta)}\bigg[(x_q^{\top}\theta)^2 + A_{1}+A_{2}\bigg]\nonumber\\
&=&\mathbb{E}_{(x_q,\theta)}  \bigg[(x_q^{\top}\theta)^2-2(v\theta^\top x_q)\theta^{\top}\left[\sigma_x^2(A+\theta b^{\top})x_q + x_q\right] + \frac{(2v\theta^\top x_q)\theta^{\top}\left[\sigma_x^2(A+\theta b^{\top})x_q + x_q\right]\exp(x_q^{\top}Ax_q)}{D\exp(x_q^{\top}(A+\theta b^{\top})^{\top}(A+\theta b^{\top})x_q/2)} \\
    &&+ \frac{2v}{D}(\theta^\top x_q)\theta^\top\left[\sigma_x^2(A+\theta b^{\top})x_q \right] \exp(x_q^{\top}(A+\theta b^{\top})^{\top}(A+\theta b^{\top})x_q)\\
    &&+\frac{v^2}{D}\theta^{\top}(\sigma_x^2 I_d-\sigma_x^4(A+\theta b^{\top})x_q x_q^{\top}(A+\theta b^{\top})^{\top}  )\theta \exp(x_q^{\top}(A+\theta b^{\top})^{\top}(A+\theta b^{\top})x_q)\\
&&+v^2(\theta^{\top}\left[\sigma_x^2(A+\theta b^{\top})x_q + x_q\right])^2 - \frac{2v^2(\theta^{\top}\left[\sigma_x^2(A+\theta b^{\top})x_q + x_q\right])^2 \exp{(x_q^{\top}Ax_q)}}{D \exp(x_q^{\top}(A+\theta b^{\top})^{\top}(A+\theta b^{\top})x_q)}\bigg] +o\left(\frac{1}{D}\right).
    \end{eqnarray*}
    From the above formulation, one can see that the optimal solution satisfies
    \begin{eqnarray*}
        v[\sigma_x^2(A+\theta b^{\top})+I_d] \approx I_d.
    \end{eqnarray*}
Taking $v=1$, $A=0_{d\times d}$, $b=\textbf{0}$, we have
\begin{eqnarray*}
\mathbb{E}\left(y_q- (W^V_{d+1,:})^{\top}
    E\phi\left(E^{\top}(W^K)^{\top} W^Q \begin{bmatrix}
        x_q\\0
    \end{bmatrix} \right)\right)^2 = O\left(\frac{\sigma_x^2}{D}\right)+o\left(\frac{1}{D}\right).
    \end{eqnarray*}

\end{proof}
\subsubsection{Theorem \ref{thm:local_population}}
\begin{proof}[Proof of Theorem \ref{thm:local_population}]
    Recall that for single-head attention, we take $A=I_d/v$ and $b=0$. Following the proof of Theorem \ref{thm:optimal}, the prediction risk becomes
 \begin{eqnarray*}
    &&\mathbb{E}\left(y_q- (W^V_{d+1,:})^{\top}
    E\phi\left(E^{\top}(W^K)^{\top} W^Q \begin{bmatrix}
        x_q\\0
    \end{bmatrix} \right)\right)^2\\
&=&\mathbb{E}_{(x_q,\theta)}\mathbb{E}_{\{x_i\}_{i\in[D]}}\left(y_q^2
    {-2y_q\left(\frac{v\sum_{i=1}^D 
    \theta^{\top} x_i\exp(x_i^{\top}Ax_q+y_ib^{\top}x_q) }{\sum \exp(x_i^{\top}Ax_q+y_ib^{\top}x_q)
    +\exp(x_q^{\top}Ax_q)}\right)}+ {\left(\frac{ v \sum_{i=1}^D 
    \theta^{\top} x_i\exp(x_i^{\top}Ax_q+y_ib^{\top}x_q) }{\sum \exp(x_i^{\top}Ax_q+y_ib^{\top}x_q)+\exp(x_q^{\top}Ax_q)}\right)^2}
    \right)\\
    &=&\mathbb{E}_{(x_q,\theta)}\mathbb{E}_{\{x_i\}_{i\in[D]}}\bigg(y_q^2
    \underbrace{-2y_q\left(\frac{v\sum_{i=1}^D 
    \theta^{\top} x_i\exp(x_i^{\top}x_q/v) }{\sum \exp(x_i^{\top}x_q/v)
    +\exp(x_q^{\top}x_q/v)}\right)}_{=A_1}+ \underbrace{\left(\frac{ v \sum_{i=1}^D 
    \theta^{\top} x_i\exp(x_i^{\top}x_q/v) }{\sum \exp(x_i^{\top}x_q/v)+\exp(x_q^{\top}x_q/v)}\right)^2}_{=A_2}
    \bigg).
    \end{eqnarray*}
    Recall that in the testing stage, $x_i\sim N(x_q,\sigma_x^2)$. In this case,
    \begin{eqnarray*}
        \mathbb{E}\exp(x_i^{\top}x_q/v)=\mathbb{E}\exp\left( \frac{(x_i-x_q)^{\top}x_q}{\sigma_x v}\sigma_x + \|x_q\|^2/v \right)=\exp\left(\frac{\sigma_x^2}{2v^2}\|x_q\|^2+\|x_q\|^2/v\right),
    \end{eqnarray*}
    and
    \begin{eqnarray*}
        \mathbb{E}x_i\exp(x_i^{\top}x_q/v)=\frac{\sigma_x^2}{v}x_q\exp\left(\frac{\sigma_x^2}{2v^2}\|x_q\|^2+\|x_q\|^2/v\right)+x_q\exp\left(\frac{\sigma_x^2}{2v^2}\|x_q\|^2+\|x_q\|^2/v\right).
    \end{eqnarray*}
    Consequently, fixing $x_q$ and $\theta$,
    \begin{eqnarray*}
        \mathbb{E}A_1&=&-2y_q\mathbb{E}\left(\frac{v\sum_{i=1}^D 
    \theta^{\top} x_i\exp(x_i^{\top}x_q/v) }{\sum \exp(x_i^{\top}x_q/v)
    +\exp(x_q^{\top}x_q/v)}\right)\\
    &=&-2y_q\mathbb{E}\left(\frac{v\sum_{i=1}^D 
    \theta^{\top} x_i\exp(x_i^{\top}x_q/v) }{\mathbb{E}\sum \exp(x_i^{\top}x_q/v)
    +\exp(x_q^{\top}x_q/v)}\right)\\
    &&+2y_q\mathbb{E}\left(\frac{v\sum_{i=1}^D 
    \theta^{\top} x_i\exp(x_i^{\top}x_q/v) }{(\mathbb{E}\sum \exp(x_i^{\top}x_q/v)
    +\exp(x_q^{\top}x_q/v))^2}\right)\left( \sum \exp(x_i^{\top}x_q/v)-\mathbb{E}\sum \exp(x_i^{\top}x_q/v) \right)\\
    &&-2y_q\mathbb{E}\left(\frac{v\sum_{i=1}^D 
    \theta^{\top} x_i\exp(x_i^{\top}x_q/v) }{(\mathbb{E}\sum \exp(x_i^{\top}x_q/v)
    +\exp(x_q^{\top}x_q/v))^3}\right)\left( \sum \exp(x_i^{\top}x_q/v)-\mathbb{E}\sum \exp(x_i^{\top}x_q/v) \right)^2+o\left(\frac{1}{D}\right)\\
    &:=&A_{11}+A_{12}+A_{13},
    \end{eqnarray*}
    where
    \begin{eqnarray*}
        A_{11}=-2y_q^2\left( \sigma_x^2+v \right)\frac{D\exp(\sigma_x^2\|x_q\|^2/(2v^2))}{D\exp(\sigma_x^2\|x_q\|^2/(2v^2))+1}=-2y_q^2(\sigma_x^2+v)+O\left(\frac{1}{D}\right),
    \end{eqnarray*}
    \begin{eqnarray*}
        A_{12}=2y_q^2\left( {2\sigma_x^2}\exp\left(\frac{2\sigma_x^2}{v^2}\|x_q\|^2\right)-{\sigma_x^2}\exp\left(\frac{\sigma_x^2}{v^2}\|x_q\|^2\right) \right)\frac{D}{(D\exp(\sigma_x^2\|x_q\|^2/(2v^2))+1)^2}=O\left(\frac{1}{D}\right),
    \end{eqnarray*}
    and
    \begin{eqnarray*}
        A_{13}=-2y_q^2(\sigma_x^2+v)\frac{D\exp(\sigma_x^2\|x_q\|^2/(2v^2))}{(D\exp(\sigma_x^2\|x_q\|^2/(2v^2))+1)^3} D\left( \exp\left(\frac{2\sigma_x^2}{v^2}\|x_q\|^2\right) -\exp\left(\frac{\sigma_x^2}{v^2}\|x_q\|^2\right) \right)  +o\left(\frac{1}{D}\right)=O\left(\frac{1}{D}\right).
    \end{eqnarray*}
    For $A_2$, when fixing $x_q$ and $\theta$, we have
    \begin{eqnarray*}
        A_2&=&\mathbb{E}\left(\frac{ v \sum_{i=1}^D 
    \theta^{\top} x_i\exp(x_i^{\top}x_q/v) }{\sum \exp(x_i^{\top}x_q/v)+\exp(x_q^{\top}x_q/v)}\right)^2=\mathbb{E}\left(\frac{ v \sum_{i=1}^D 
    \theta^{\top} x_i\exp(x_i^{\top}x_q/v) }{\mathbb{E}\sum \exp(x_i^{\top}x_q/v)+\exp(x_q^{\top}x_q/v)}\right)^2+O\left(\frac{1}{D}\right)\\
    &=&\frac{D(D-1)}{(D\exp(\sigma_x^2\|x_q\|^2/(2v^2))+1)^2}\left[(\sigma_x^2+v)x_q^{\top}\theta\right]^2\exp\left(\frac{\sigma_x^2}{v^2}\|x_q\|^2\right)+O\left(\frac{1}{D}\right)\\
    &=&(\sigma_x^2+v)^2y_q^2++O\left(\frac{1}{D}\right).
    \end{eqnarray*}
    To conclude, when fixing $x_q$ and $\theta$, we obtain
    \begin{eqnarray*}
        \mathbb{E}\left(y_q- (W^V_{d+1,:})^{\top}
    E\phi\left(E^{\top}(W^K)^{\top} W^Q \begin{bmatrix}
        x_q\\0
    \end{bmatrix} \right)\right)^2=(\sigma_x^2+v-1)^2y_q^2+O\left(\frac{1}{D}\right).
    \end{eqnarray*}
\end{proof}

\newpage
\section{Simulation and Experiment Details}
\label{apd:experiment}
\subsection{Visualization of Single-Head Attention Score}

\begin{figure}[!ht]
\centering
\vspace{-0.2cm}
  \begin{subfigure}{.45\textwidth}
        \centering
\includegraphics[scale=0.45]{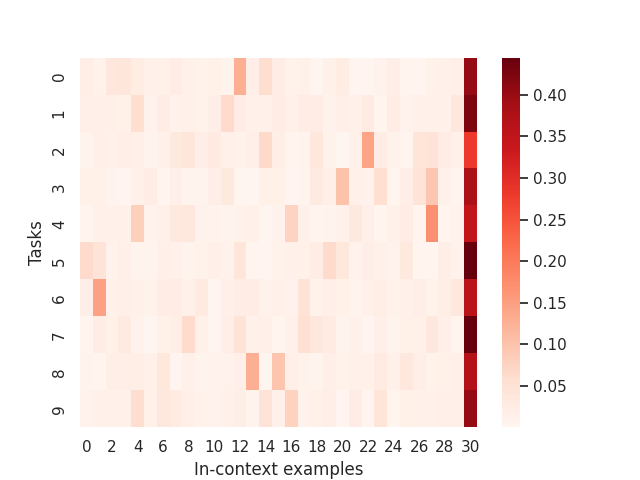}
    \caption{$v>0$}
    \label{fig:atten_score_v+}
    \end{subfigure}\hspace{.1in}
      \begin{subfigure}{.45\textwidth}
    \centering
    \includegraphics[scale=0.45]{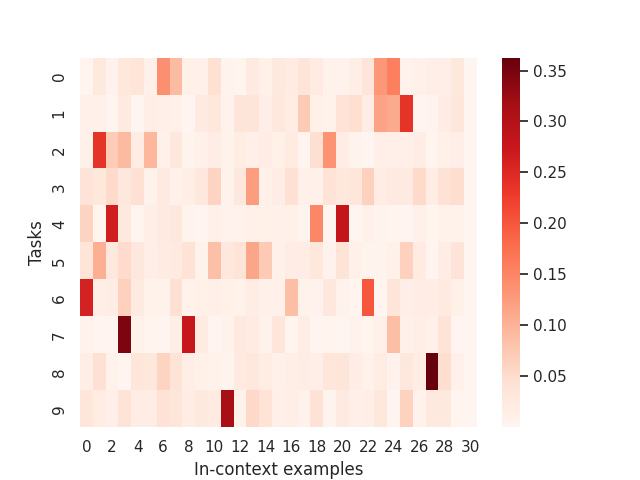}
    \caption{$v<0$}
    \label{fig:atten_score_v-}
    \end{subfigure}\hspace{.1in}
    \vspace{-0.1in}
    \caption{Single-Head Attention Score for 10 tasks.}
    \label{fig:atten_score}
\end{figure}

Based on Theorem \ref{thm:optimal}, the optimal $A$ is in the format of $I_d/v+o$. As a result, there are two possible cases. (i) When $v>0$, the attention score of $x_q$ against itself is usually the largest one as $x_q^{\top}Ax_q=\|x_q\|^2/v$ is always positive. (ii) When $v<0$, the attention score of $x_q$ against itself is always small. Figure \ref{fig:atten_score} shows these two cases correspondingly.

\vspace{-0.1in}






\subsection{Noisy Response and Correlated Features}

\begin{figure}[!ht]
    \centering
    \begin{subfigure}{.42\textwidth}
        \centering
        \includegraphics[width=.9\linewidth]{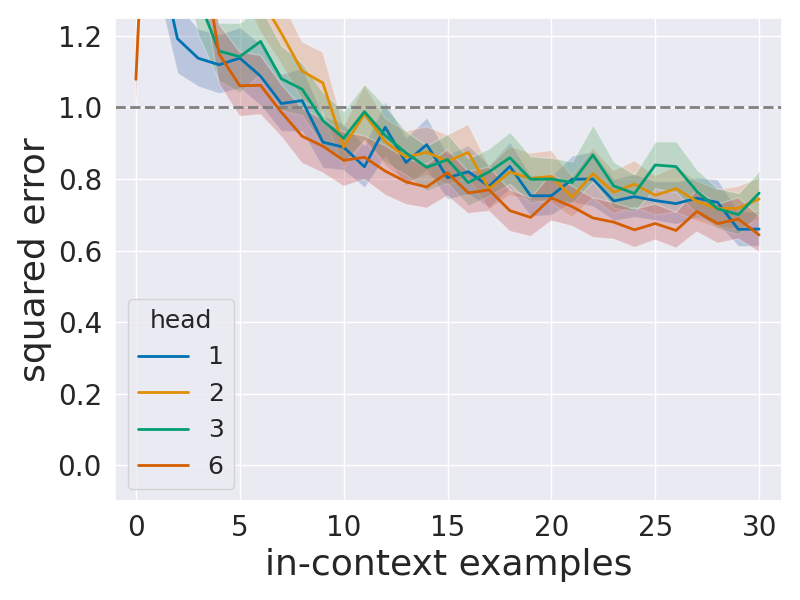}
        \caption{No read-in layer}
    \end{subfigure}\hspace{.1in}
    \begin{subfigure}{.42\textwidth}
        \centering
        \includegraphics[width=.9\linewidth]{figures/append/noise_no_readin.png}
        \caption{$p=6$}
    \end{subfigure}
    \begin{subfigure}{.42\textwidth}
        \centering
        \includegraphics[width=.9\linewidth]{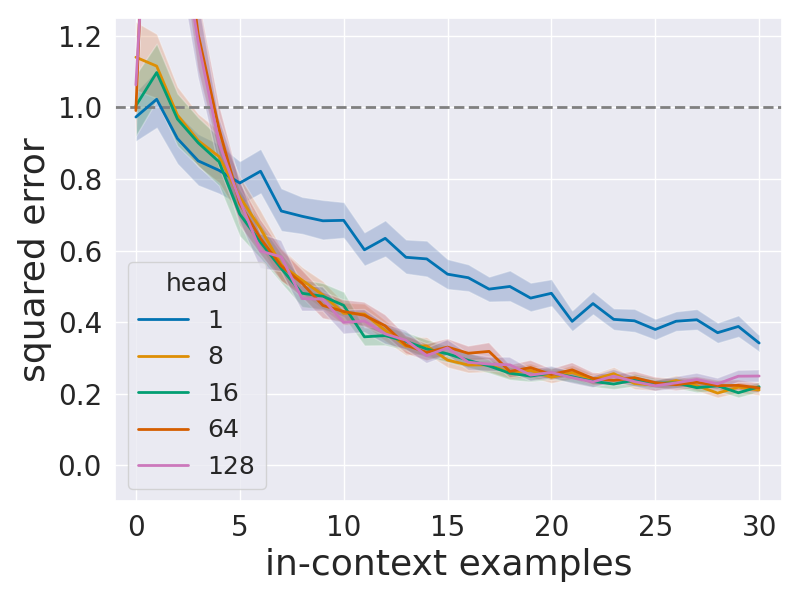}
        \caption{$p=128$}
    \end{subfigure}\hspace{.1in}
    \begin{subfigure}{.42\textwidth}
        \centering
        \includegraphics[width=.9\linewidth]{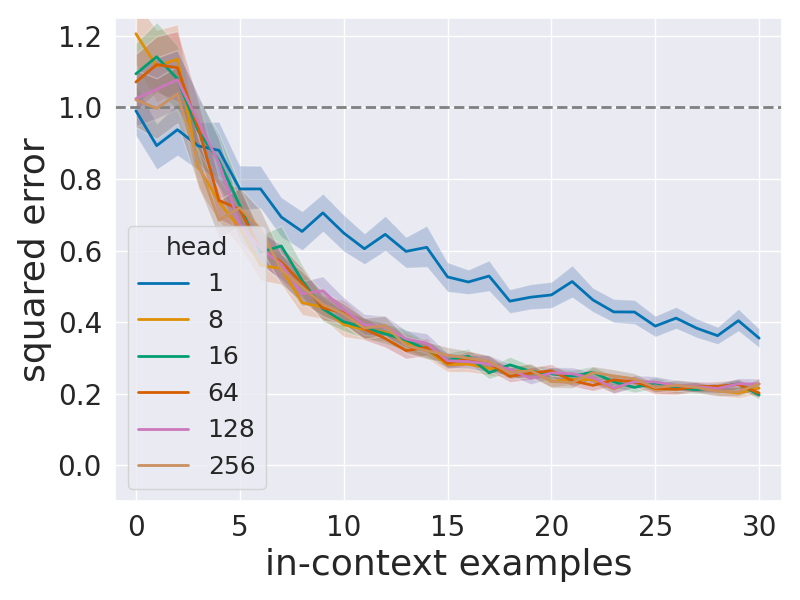}
        \caption{$p=256$}
    \end{subfigure}
    \caption{ICL performance with noisy responses.}
    \label{fig:append_noise}
\end{figure}

For noisy response and correlated features, we conduct experiments to verify the effectiveness of multi-head attention. The results for noisy label can be found in Figure~\ref{fig:append_noise}. While the best prediction loss is away from zero, one can still see that with sufficient input embedding dimension, multi-head attention improves the performance.

\begin{figure}
    \centering
    \begin{subfigure}{.42\textwidth}
        \centering
        \includegraphics[width=.9\linewidth]{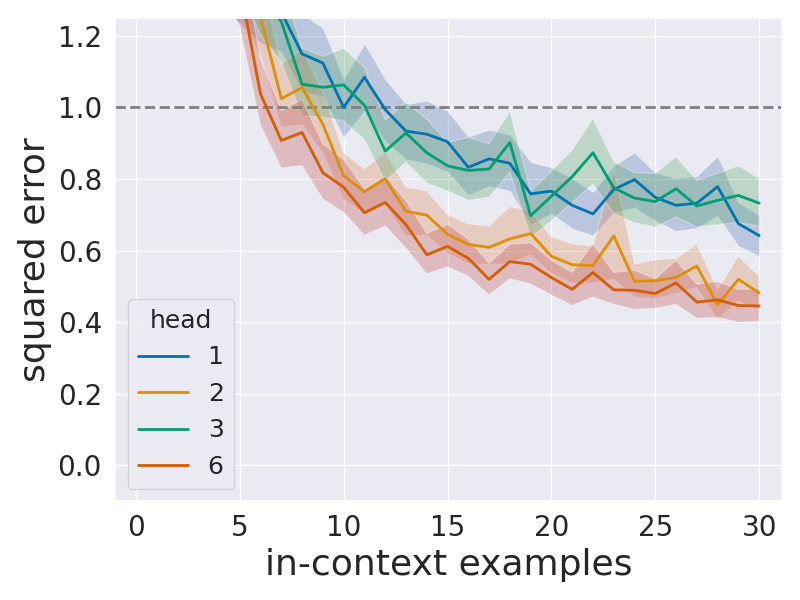}
        \caption{No read-in layer}
    \end{subfigure}\hspace{.1in}
    \begin{subfigure}{.42\textwidth}
        \centering
        \includegraphics[width=.9\linewidth]{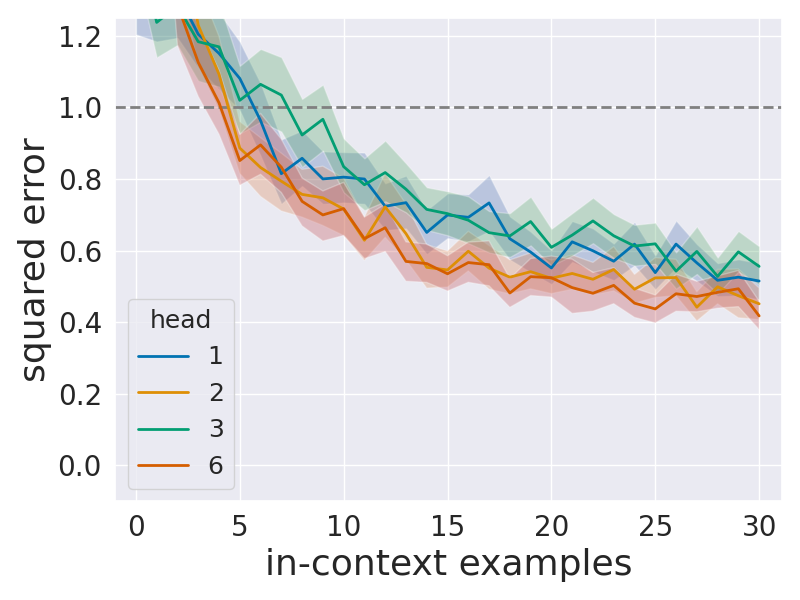}
        \caption{$p=6$}
    \end{subfigure}
    \begin{subfigure}{.42\textwidth}
        \centering
        \includegraphics[width=.9\linewidth]{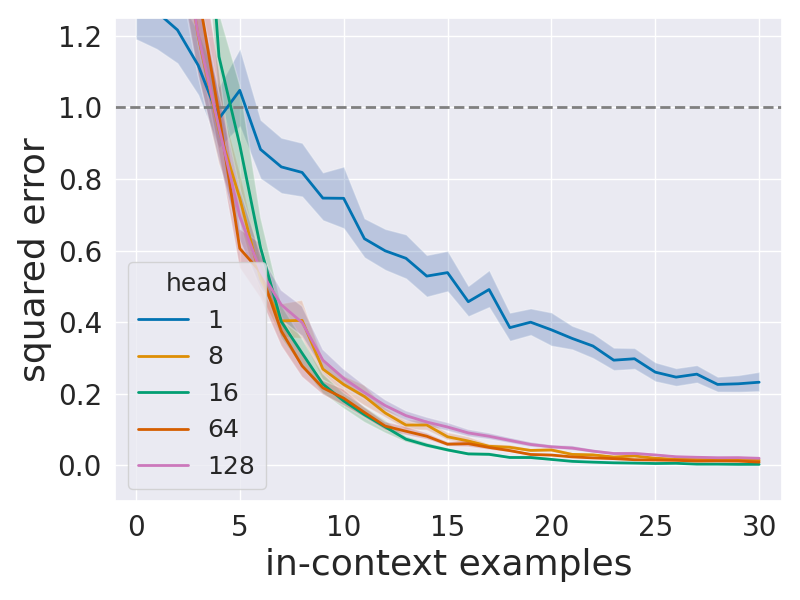}
        \caption{$p=128$}
    \end{subfigure}\hspace{.1in}
    \begin{subfigure}{.42\textwidth}
        \centering
        \includegraphics[width=.9\linewidth]{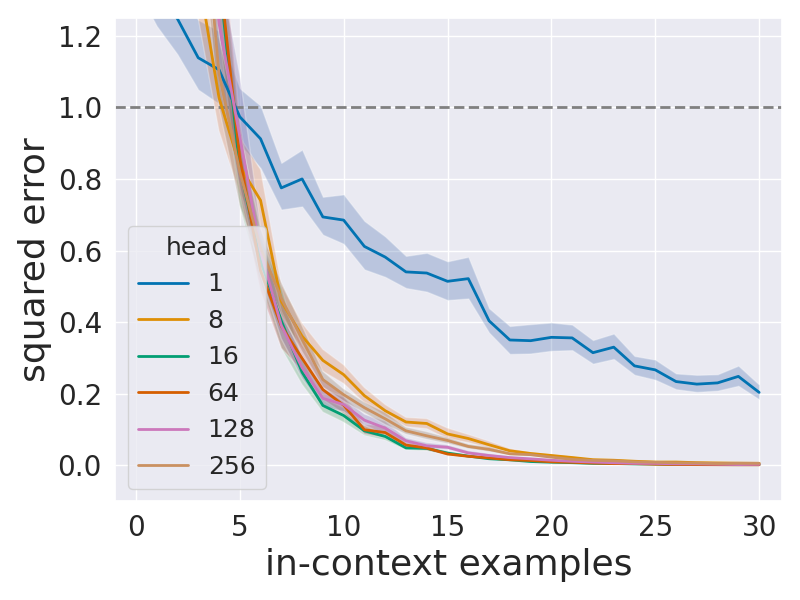}
        \caption{$p=256$}
    \end{subfigure}
    \caption{ICL Performance with correlated features.}
    \label{fig:append_correlated}
\end{figure}

For correlated features, to generate $\Sigma$, we follow the procedure in \citet{zhang2023trained} and take the diagonal elements following exp(1) distribution. For the off diagonal elements, we take all of them as 0.1. From Figure \ref{fig:append_correlated} we can see that multi-head attention with $p/h>d$ is better than single-head attention.

\newpage

\subsection{Other Figures}

\label{apd:fig}
\begin{figure}[!ht]
    \centering
    \begin{subfigure}{.42\textwidth}
        \centering
        \includegraphics[width=.9\linewidth]{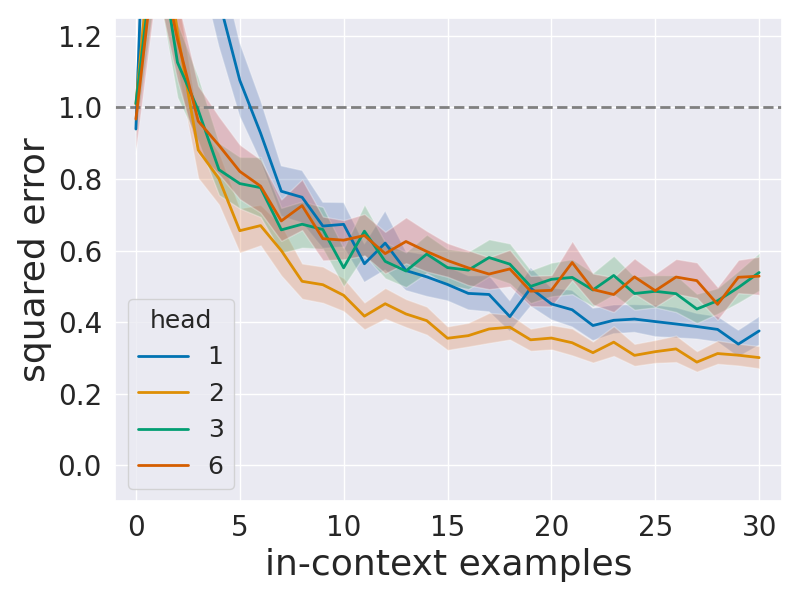}
        \caption{$p=6$}
        \label{fig:emb6}
    \end{subfigure}\hspace{.1in}
    \begin{subfigure}{.42\textwidth}
        \centering
        \includegraphics[width=.9\linewidth]{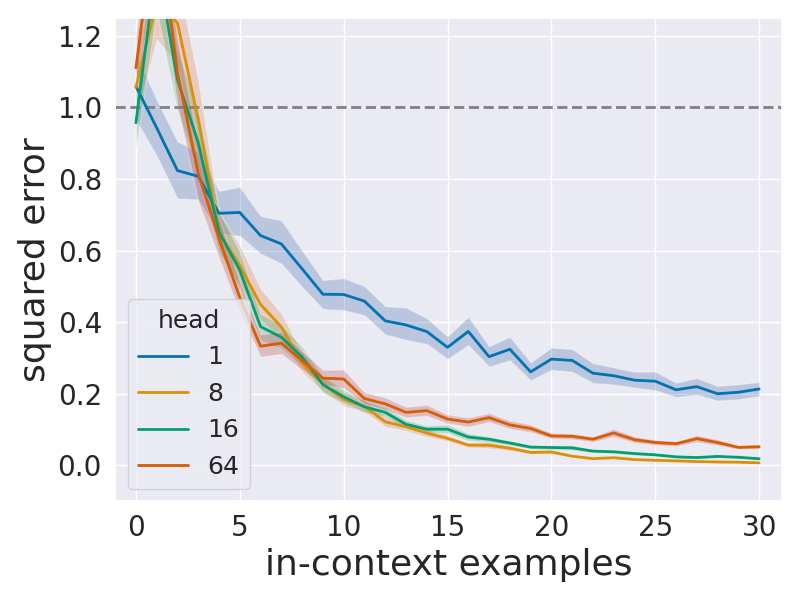}
        \caption{$p=64$}
        \label{fig:emb64}
    \end{subfigure}
    \begin{subfigure}{.42\textwidth}
        \centering
        \includegraphics[width=.9\linewidth]{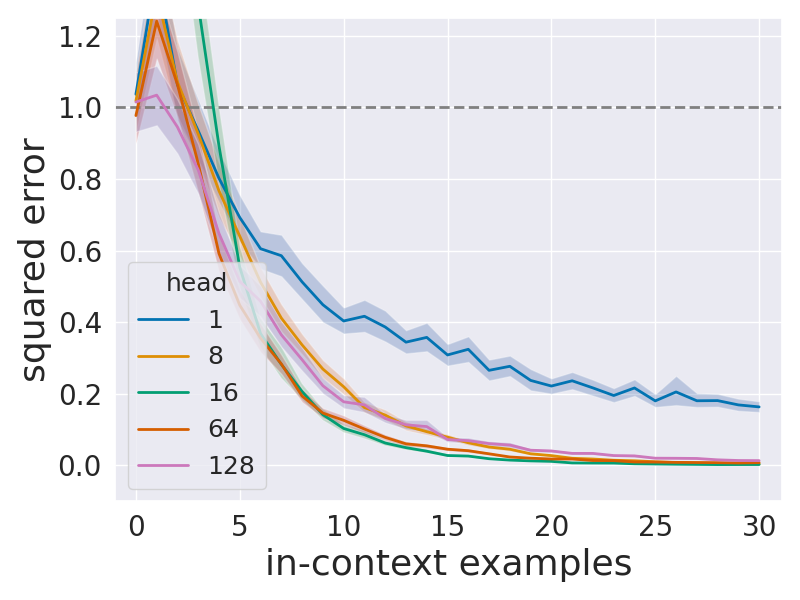}
        \caption{$p=128$}
        \label{fig:emb128}
    \end{subfigure}\hspace{.1in}
    \begin{subfigure}{.42\textwidth}
        \centering
        \includegraphics[width=.9\linewidth]{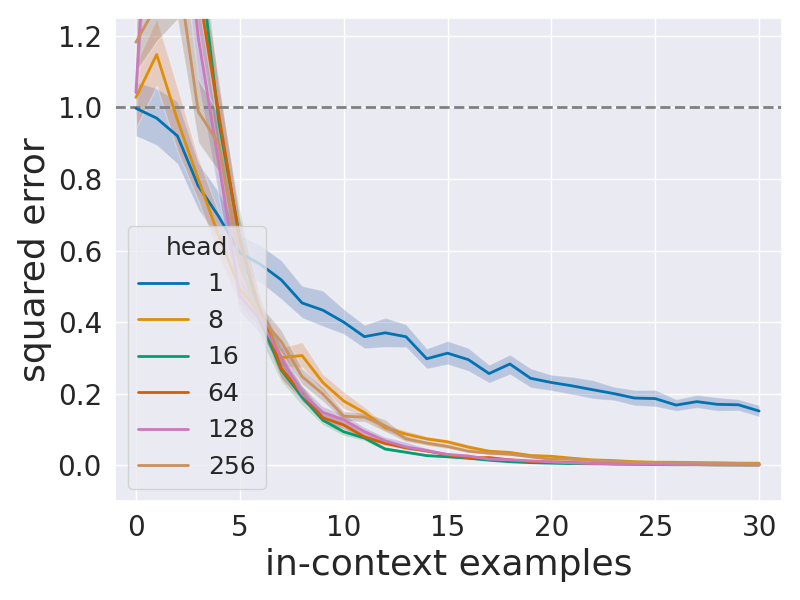}
        \caption{$p=256$}
        \label{fig:emb256}
    \end{subfigure}
    \caption{Standard experiment.}
    \label{fig:append_standard}
\end{figure}

\begin{figure}[htp]
    \centering
    \begin{subfigure}{.42\textwidth}
        \centering
        \includegraphics[width=.9\linewidth]{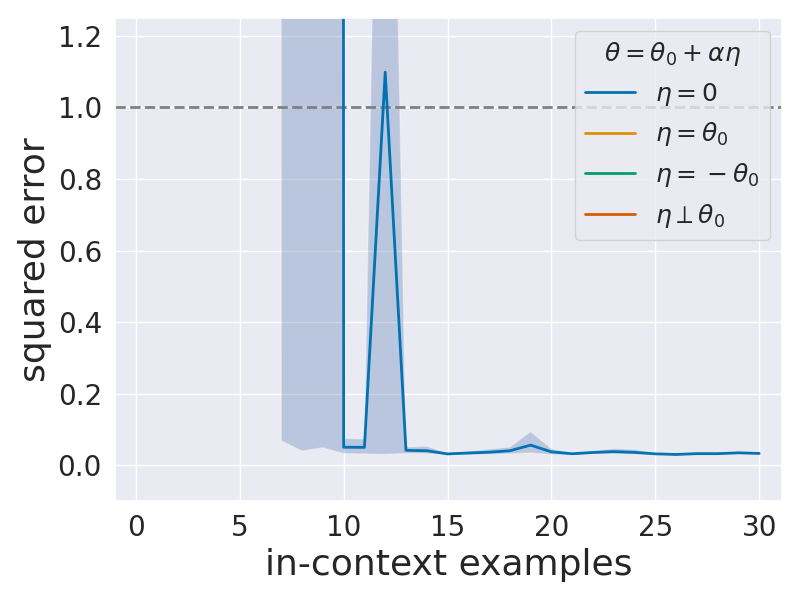}
        \caption{$\sigma=0.01$}
        \label{fig:head1_alpha0}
    \end{subfigure}\hspace{.1in}
    \begin{subfigure}{.42\textwidth}
        \centering
        \includegraphics[width=.9\linewidth]{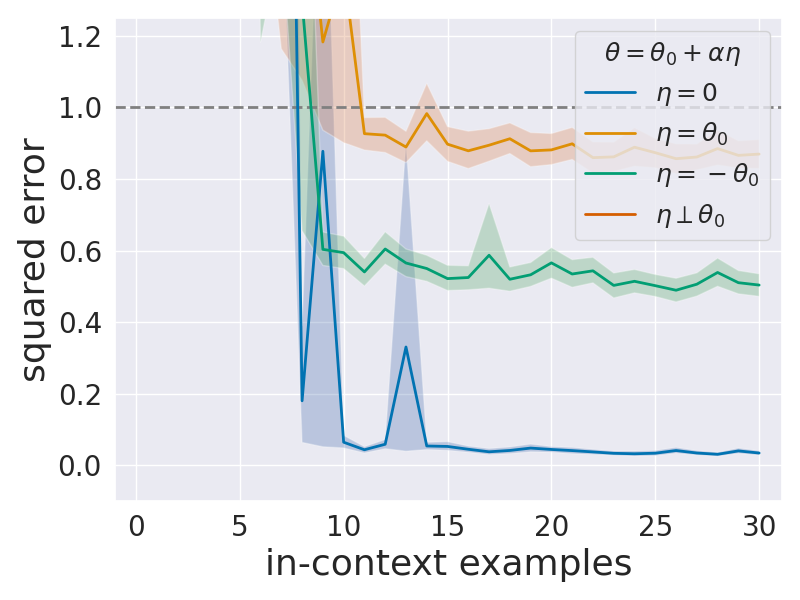}
        \caption{$\sigma=0.05$}
        \label{fig:head1_alpha0.01_random}
    \end{subfigure}
    \begin{subfigure}{.42\textwidth}
        \centering
        \includegraphics[width=.9\linewidth]{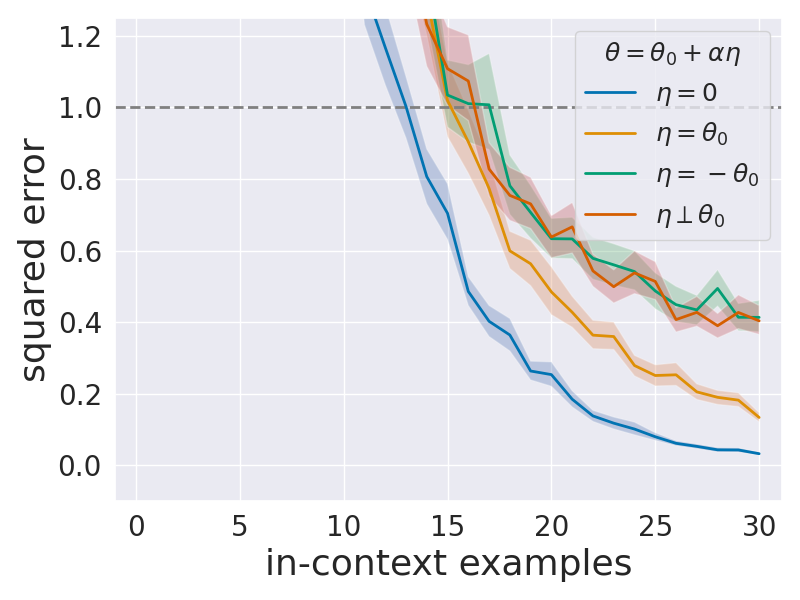}
        \caption{$\sigma=0.1$}
        \label{fig:head1_alpha0.01_parallel}
    \end{subfigure}\hspace{.1in}
    \begin{subfigure}{.42\textwidth}
        \centering
        \includegraphics[width=.9\linewidth]{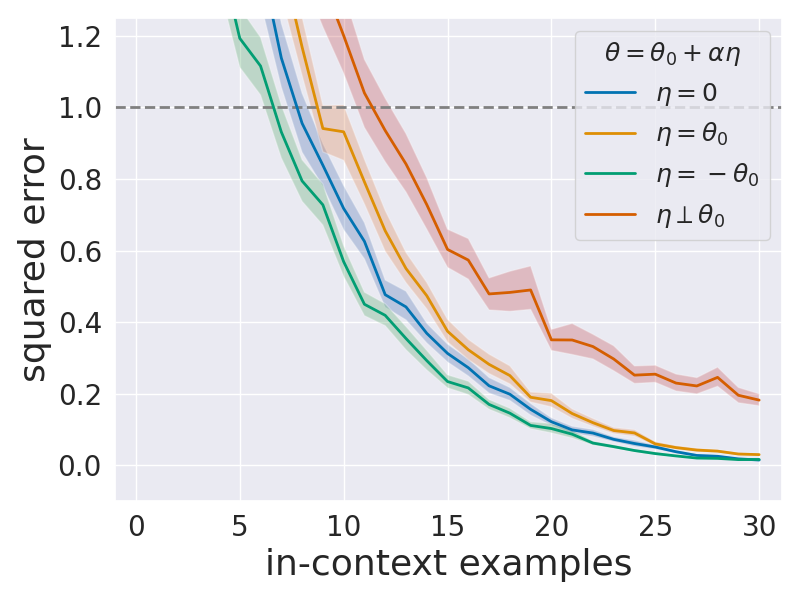}
        \caption{$\sigma=0.5$}
        \label{fig:head1_alpha0.01_reverse_parallel}
    \end{subfigure}
    \caption{ICL performance given prior knowledge. Single-head attention. Different test method of prior knowlegde (random, parallel, etc) in different subfigures. $\alpha=0.1$. Add description}
    \label{fig:append_prior1}
\end{figure}

\begin{figure}[htp]
    \centering
    \begin{subfigure}{.42\textwidth}
        \centering
        \includegraphics[width=.9\linewidth]{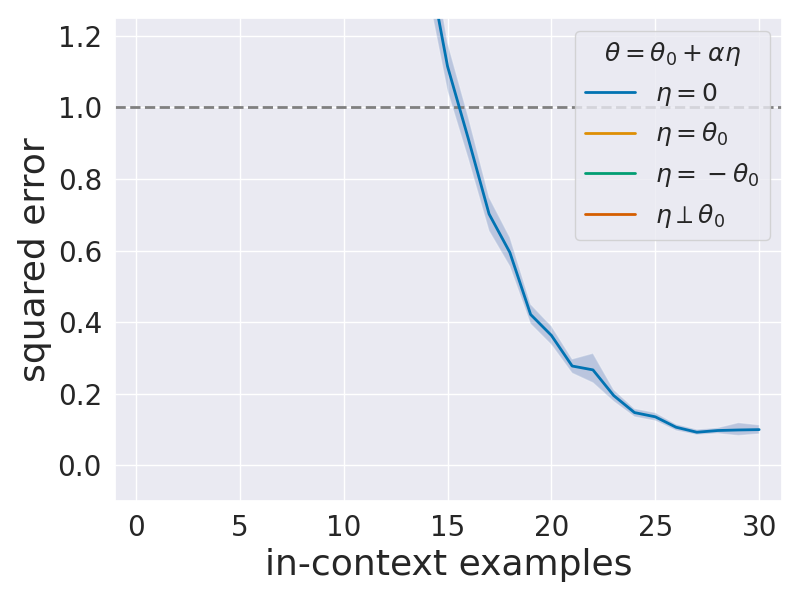}
        \caption{$\sigma=0.01$}
        \label{fig:head16_alpha0}
    \end{subfigure}\hspace{.1in}
    \begin{subfigure}{.42\textwidth}
        \centering
        \includegraphics[width=.9\linewidth]{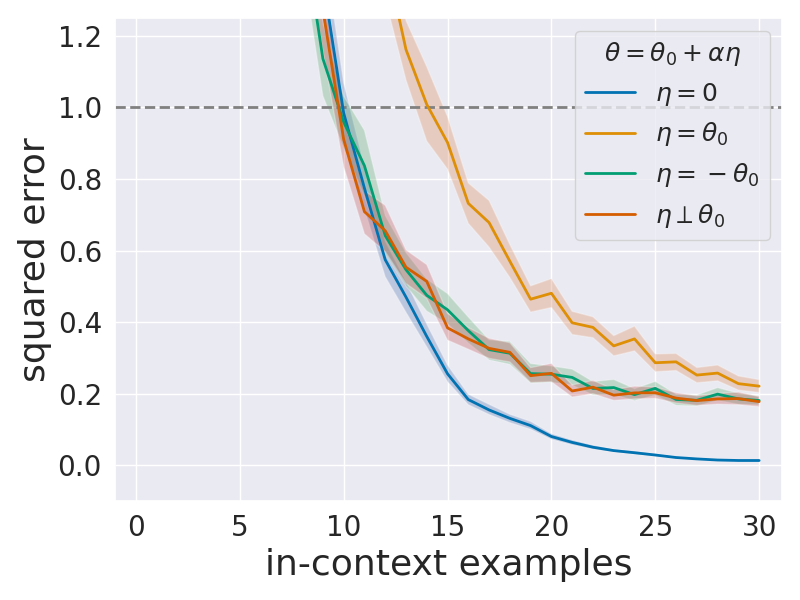}
        \caption{$\sigma=0.05$}
        \label{fig:head16_alpha0.01_random}
    \end{subfigure}
    \begin{subfigure}{.42\textwidth}
        \centering
        \includegraphics[width=.9\linewidth]{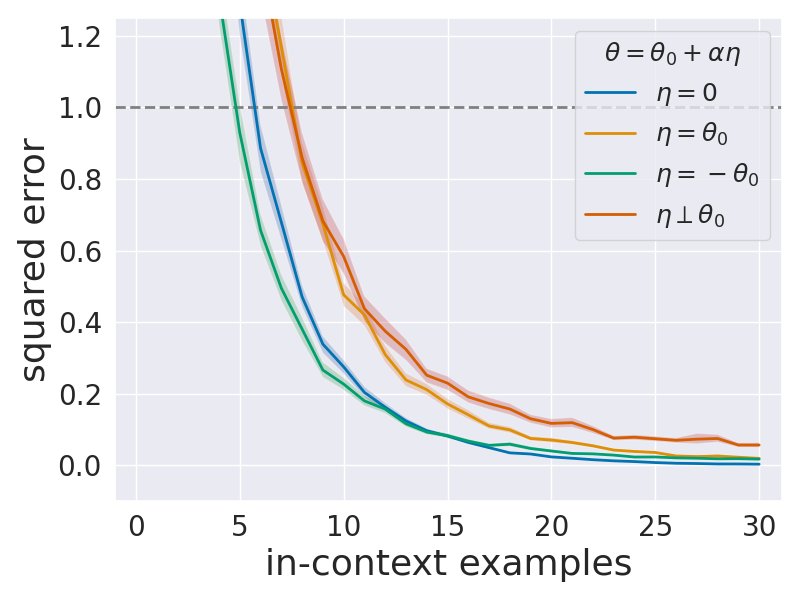}
        \caption{$\sigma=0.1$}
        \label{fig:head16_alpha0.01_parallel}
    \end{subfigure}\hspace{.1in}
    \begin{subfigure}{.42\textwidth}
        \centering
        \includegraphics[width=.9\linewidth]{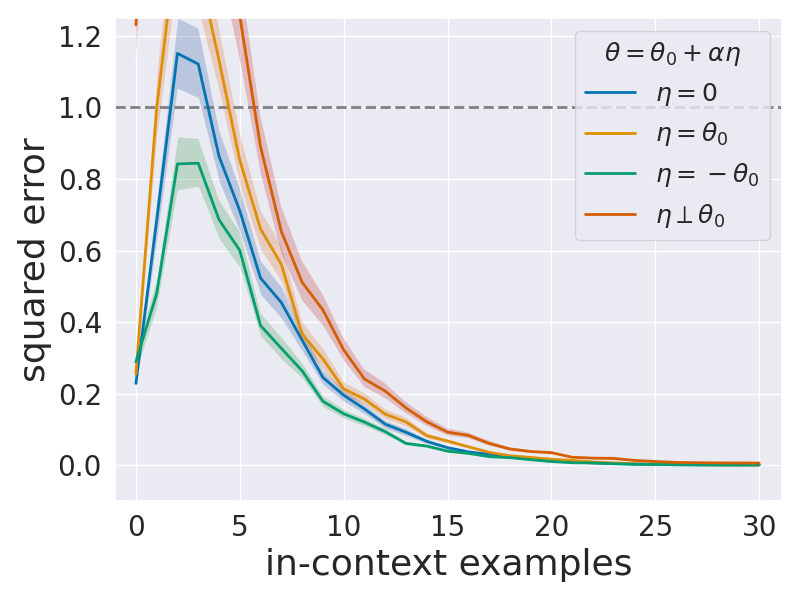}
        \caption{$\sigma=0.5$}
        \label{fig:head16_alpha0.01_reverse_parallel}
    \end{subfigure}
    \caption{$\alpha=0.1$ head=16 Different test method of prior knowlegde (random, parallel ..) in different subfigure, different sigma in different line}
    \label{fig:append_prior2}
\end{figure}

\begin{figure}[htp]
    \centering
    \begin{subfigure}{.42\textwidth}
        \centering
        \includegraphics[width=.9\linewidth]{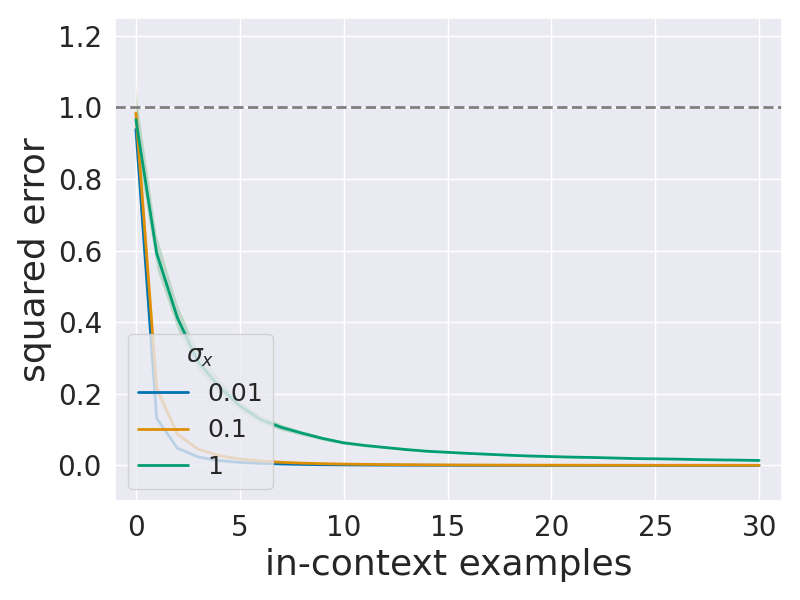}
        \caption{Training and testing datasets follow the same distribution. Single head.}
    \end{subfigure}\hspace{.1in}
    \begin{subfigure}{.42\textwidth}
        \centering
        \includegraphics[width=.9\linewidth]{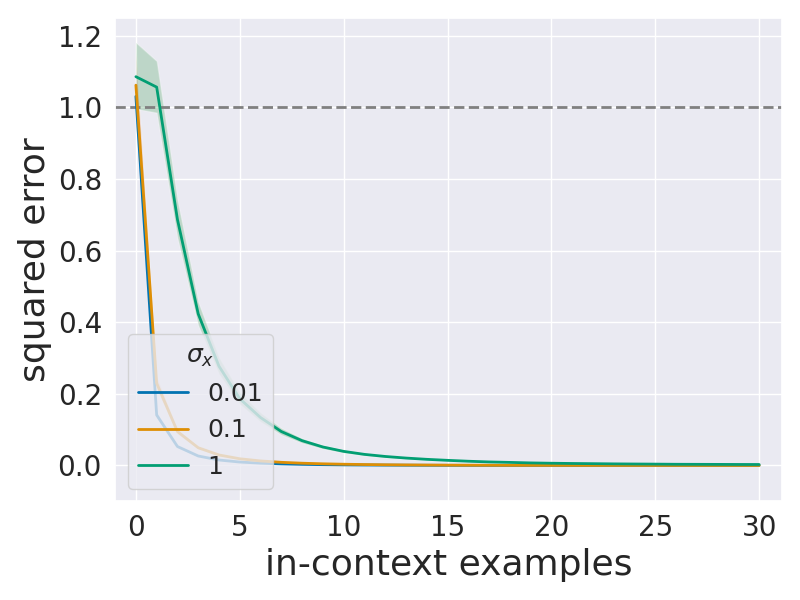}
        \caption{Training and testing datasets follow the same distribution. 16 heads.}
    \end{subfigure}
    \begin{subfigure}{.42\textwidth}
        \centering
        \includegraphics[width=.9\linewidth]{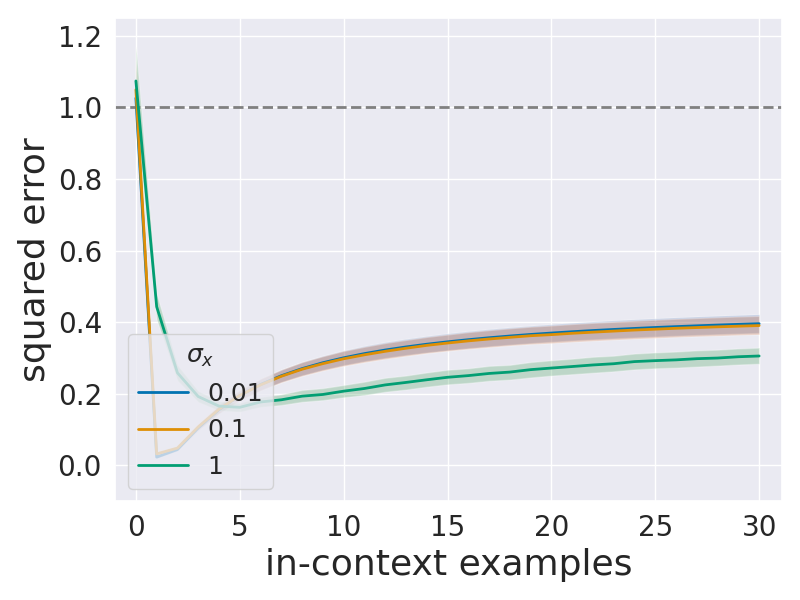}
        \caption{Training and testing datasets follow different distributions. Single head.}
    \end{subfigure}\hspace{.1in}
    \begin{subfigure}{.42\textwidth}
        \centering
        \includegraphics[width=.9\linewidth]{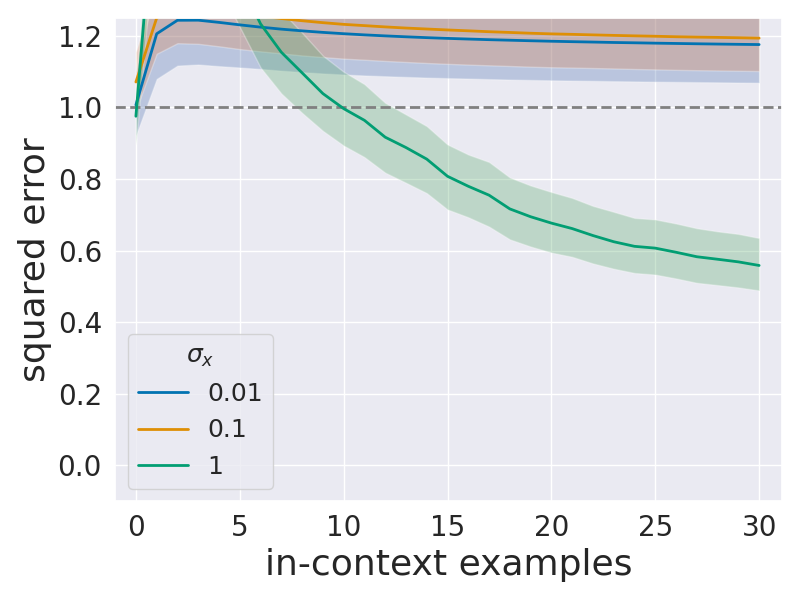}
        \caption{Training and testing datasets follow different distributions. 16 heads.}
    \end{subfigure}
    \caption{ICL Performance with local examples.}
    \label{fig:append_local}
\end{figure}





\end{document}